\documentclass[10pt,twocolumn,letterpaper]{article}

\usepackage[pagenumbers]{cvpr} 

\makeatletter
\ifdefined\IfFormatAtLeastTF
\IfFormatAtLeastTF{2025-06-01}{%
\RemoveFromHook{build/column/before}[lineno]
\AtBeginDocument{%
  \def\@LN@makecol{%
    \@LN@orig@makecol
    \setbox\@outputbox\vbox{%
      \boxmaxdepth\@maxdepth
      \protected@write\@auxout{}{\string\def\string\@LN@column{\if@firstcolumn1\else2\fi}}%
      \box\@outputbox}}}
}{}
\fi
\makeatother

\usepackage[english]{babel}
\usepackage{bbm}

\usepackage{colortbl}
\definecolor{oursrow}{rgb}{0.98,0.93,0.90}
\usepackage{placeins}
\usepackage{titletoc}

\usepackage{algorithm}
\usepackage{algpseudocode}
\usepackage{amsthm}

\newcommand{\fst}[1]{\textbf{#1}}
\newcommand{\snd}[1]{\underline{#1}}

\definecolor{cvprblue}{rgb}{0.21,0.49,0.74}
\usepackage[pagebackref,breaklinks,colorlinks,allcolors=cvprblue]{hyperref}

\def\confName{CVPR}
\def\confYear{2027}

\title{AlignPre: Skill-Aligned Post-Training for Latent-Free Generative Precipitation Nowcasting}

\author{Demin Yu \quad \quad
Kenghong Lin \quad \quad
Xutao Li\thanks{Corresponding author.}\\
Harbin Institute of Technology, Shenzhen\\
{\small\texttt{deminyu98@gmail.com, linkenghong@hit.edu.cn, lixutao@hit.edu.cn}}
}

\begin{document}
\maketitle
\begin{abstract}
Precipitation nowcasting requires accurate short-term predictions under rapidly evolving spatiotemporal dynamics. Generative diffusion models can produce detailed and competitive radar forecasts by modeling conditional distributions. However, most existing approaches rely on separately trained components, such as deterministic branches or latent compression modules. More fundamentally, distribution matching alone fails to explicitly optimize operational forecast skill. We propose AlignPre, a streamlined pixel-space nowcasting framework that bridges this gap via skill-aligned post-training (SAPT). AlignPre employs an end-to-end flow-matching forecaster with a hierarchical spatiotemporal architecture for distribution matching and direct radar-field synthesis, without latent compression. During post-training, SAPT exploits the analytic structure of forecast verification scores to construct a differentiable objective. By backpropagating gradients of this objective from radar fields to the generator, SAPT turns forecast verification into direct training feedback. Across three real-world benchmarks spanning different forecast horizons, AlignPre consistently surpasses state-of-the-art deterministic and generative baselines in forecast skill, demonstrating that aligning generative modeling with domain-specific verification provides an effective paradigm for meteorological forecasting.
Our code and models will be released.
\end{abstract}

\section{Introduction}
\label{sec:intro}

\begin{figure}[!htbp]
\centering
\begin{subfigure}[t]{0.49\linewidth}
\centering
\includegraphics[width=\linewidth]{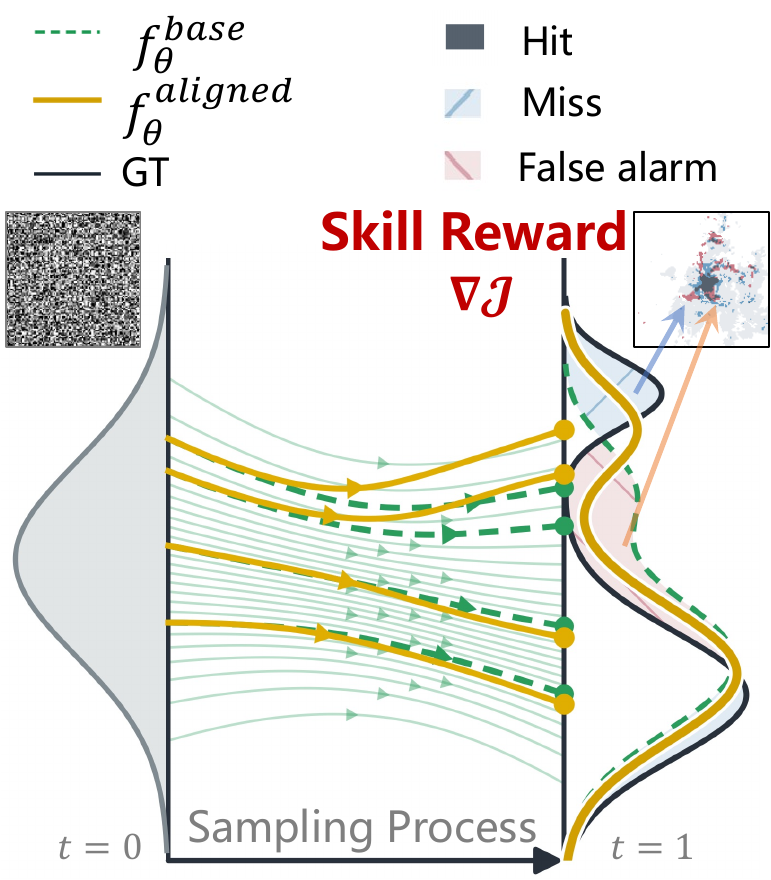}
\caption{}
\label{fig:intro_sampling}  
\end{subfigure}\hfill
\begin{subfigure}[t]{0.49\linewidth}
\centering
\includegraphics[width=\linewidth]{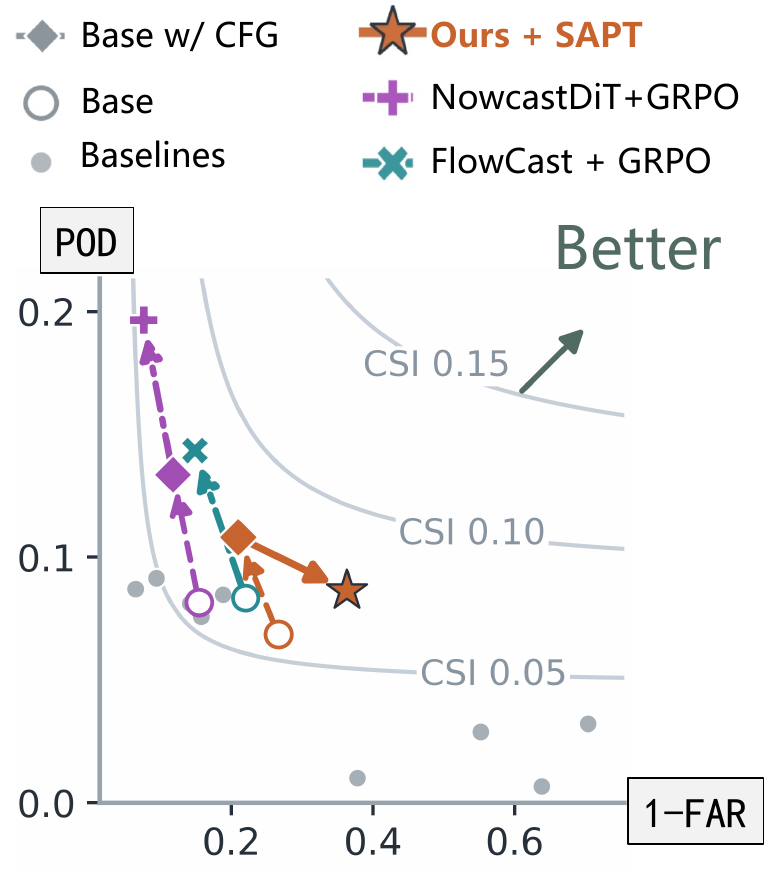}
\caption{}
\label{fig:intro_pod_far}
\end{subfigure}
\vspace{-2ex}
\caption{\textbf{Skill alignment for generative nowcasting.} (a) SAPT backpropagates skill gradients ($\nabla \mathcal{J}$) from predicted fields to steer sampling trajectories toward ground truth (schematic). (b) Precision-recall trade-off on extreme precipitation (VIL $\geq 219$). \textcolor[HTML]{8B959B}{CFG} and \textcolor[HTML]{A24FB5}{GRPO} inflate recall via excessive FAR \textcolor[HTML]{A24FB5}{(upper-left)}, whereas \textcolor[HTML]{C8632E}{SAPT} improves both skill and precision \textcolor[HTML]{C8632E}{(upper-right)}.}
\label{fig:intro_overview}
\vspace{-2ex}
\end{figure}

Radar precipitation nowcasting predicts near-term rainfall location, intensity, and evolution from recent observations to support severe convection and flash-flood warnings, where capturing high-impact events is critical~\cite{ravuri2021dgmr,zhang2023nowcastnet}.
Early deterministic methods trained via error minimization achieve stable average accuracy and low false alarms, yet tend to smooth out fine-scale structures and underestimate extreme precipitation~\cite{shi2015convlstm,gao2022earthformer,ravuri2021dgmr}.
Conditional generative models improve forecast detail by learning future precipitation distributions, but prevailing hybrid designs rely on an auxiliary deterministic branch for structural priors, introducing multi-stage training overhead and tight module coupling~\cite{yu2024diffcast,gong2024cascast}.
Inspired by video Diffusion Transformers, recent pure generative models, such as FlowCast~\cite{ribeiro2026flowcast}, NowcastDiT~\cite{xu2026nowcastdit}, and FREUD~\cite{schusterbauer2026freud}, demonstrate that a standalone generative backbone can achieve competitive performance without deterministic priors.

However, distribution-matching pretraining in pure generative models does not explicitly optimize operational forecast skill.
Operational verification evaluates spatial event agreement across lead times and intensity thresholds using contingency metrics such as the critical success index (CSI) and false alarm ratio (FAR)~\cite{jolliffe2012forecast}.
In contrast, flow-matching and score-based models regress intermediate noisy states to learn conditional distributions, without directly supervising multi-step sampled forecasts on these event metrics~\cite{lipman2023flow,song2021score}.
Post-training bridges this gap: in visual generation, reward and preference signals~\cite{domingoenrich2025adjoint, wallace2024diffusiondpo} are widely used to fine-tune pretrained models for perceptual alignment~\cite{fan2023dpok, clark2024draft}, task objectives~\cite{prabhudesai2024vader, liu2025flowgrpo}, or physical consistency~\cite{li2025pisa}.
Motivated by these advances, recent nowcasting studies adopt group-relative policy optimization~\cite{xu2026nowcastdit} or multi-stage preference learning~\cite{xu2026syncast} to align verification scores, confirming the value of post-training.

Nevertheless, adapting visual generative modeling and post-training to radar nowcasting requires addressing two domain-specific challenges in representation and verification optimization.
First, intense precipitation is sparse, where minor local intensity errors can alter thresholded event decisions; yet latent-space approaches rely on separately pretrained, lossy autoencoders~\cite{rombach2022ldm,schusterbauer2026freud} whose frozen decoders cannot be updated during post-training to calibrate pixel-level intensities.
Second, CSI and FAR are defined over discrete aggregated event counts and involve a miss–false-alarm trade-off: expanding predicted rain areas reduces misses to boost CSI, whereas curbing false alarms for lower FAR favors more conservative coverage.
Existing methods treat these scores as black-box rewards via multi-trajectory reinforcement learning or multi-stage preference optimization~\cite{xu2026nowcastdit,xu2026syncast}, adding sampling or scheduling complexity while collapsing grid-level spatial errors into coarse scalar feedback that can encourage over-forecasting.

To address these issues, we propose AlignPre, a streamlined nowcasting framework integrating end-to-end pixel-space generation with skill-aligned post-training (SAPT).
We first pretrain a conditional flow-matching backbone directly on radar fields to predict clean future sequences via a hierarchical spatiotemporal network, jointly modeling fine-scale local evolution and coarse-scale global context without deterministic branches or separate autoencoders.
Building on this backbone, SAPT leverages the analytic relationship between CSI/FAR improvements and weighted event errors relative to a reference model to form a unified differentiable objective, applying a smooth threshold approximation to backpropagate grid-level gradients directly from predicted radar fields to the generator (Figure~\ref{fig:intro_overview}(a)).
Each stochastic update requires only a single sampled trajectory per condition, while reference regularization constrains drift from the pretrained generator.
Experiments across SEVIR-1h, SEVIR-2h~\cite{veillette2020sevir}, and Guangdong-3h demonstrate that AlignPre achieves leading CSI-M while maintaining low false alarms (FAR-M) and high perceptual quality (LPIPS~\cite{zhang2018lpips}).
As shown in Figure~\ref{fig:intro_overview}(b), whereas guidance and RL baselines boost recall at the cost of severe false alarms, AlignPre uniquely advances toward the upper-right high-CSI, low-FAR regime.

Our main contributions are as follows:
\begin{itemize}
    \item We show for the first time in generative nowcasting that operational verification scores can be formulated into a differentiable skill objective, enabling direct backpropagation of grid-level gradients from predicted radar fields to optimize the generator.
    \item We propose AlignPre, a streamlined generative nowcasting framework that unifies end-to-end pixel-space flow matching with skill-aligned post-training, removing separate deterministic branches and frozen autoencoders.
    \item Through unified evaluation and ablations across three radar benchmarks spanning 1- to 3-hour lead times, we show that AlignPre improves overall and intense-precipitation CSI while effectively controlling FAR and preserving perceptual quality.
\end{itemize}

\section{Related Work}
\label{sec:related}

\noindent\textbf{Generative precipitation nowcasting.}
Generative nowcasting models uncertain storm evolution via conditional sampling~\cite{ravuri2021dgmr,leinonen2023ldcast,gao2023prediff}, where hybrid diffusion and flow methods pair deterministic predictors or physical priors with generative modules to constrain large-scale structure~\cite{yu2024diffcast,gong2024cascast,feng2025perceptual,zhu2026pixelflowcast,nagashima2026pgfmm}.
Recent standalone generators achieve competitive skill without deterministic branches~\cite{ribeiro2026flowcast,xu2026nowcastdit,schusterbauer2026freud}, yet still rely on decoupled latent spaces and distribution-matching objectives that do not directly optimize event verification.
AlignPre advances this standalone paradigm via end-to-end pixel-space generation explicitly aligned with operational event metrics.

\smallskip
\noindent\textbf{Pixel-space generative modeling.}
\label{sec:related_pixel}
Latent generators reduce compute via pretrained autoencoders at the cost of lossy compression and decoding uncertainty~\cite{rombach2022ldm,schusterbauer2026freud}.
Pixel-space models eliminate external tokenizers through resolution-aware schedules~\cite{hoogeboom2023simple,hoogeboom2025sid2}, clean-data ($x$-) prediction~\cite{li2026jit}, and hierarchical Transformers~\cite{crowson2024hourglass,yu2026pixeldit,chen2026dip,vallaeys2026ujit}.
For thresholded radar verification, end-to-end pixel synthesis is vital: whereas a frozen latent decoder passes gradients without updating its reconstruction mapping, a pixel-space generator directly recalibrates output intensities.
We thus pair $x$-prediction with a hierarchical spatiotemporal backbone to jointly learn and align representations and field synthesis.

Concurrent work JWS~\cite{wiese2026jws} also adopts a U-shaped JiT-based nowcaster, but optimizes continuous distributional scores rather than post-training for operational event skill.

\smallskip
\noindent\textbf{Post-training for forecast skill.}
\label{sec:related_posttraining}
Diffusion and flow post-training adapts pretrained samplers via policy gradients~\cite{black2024ddpo,fan2023dpok,liu2025flowgrpo,li2026mixgrpo}, preference optimization~\cite{wallace2024diffusiondpo}, or regularized reward backpropagation~\cite{clark2024draft,prabhudesai2024vader,li2025pisa,domingoenrich2025adjoint}.
In meteorology, proper scores (e.g., CRPS) calibrate ensembles~\cite{lang2024aifscrps,alet2025fgn,jacq2026rmmd}, smooth threshold relaxations supply deterministic event gradients~\cite{choi2026atloss}, and recent generative nowcasters optimize verification metrics via multi-rollout RL or two-stage preference learning~\cite{xu2026nowcastdit,xu2026syncast}.
However, distributional scores overlook discrete CSI--FAR trade-offs, while black-box RL and preference methods collapse grid-level errors into trajectory-level feedback at multi-rollout or multi-stage cost.
SAPT instead exploits the analytic link between CSI/FAR gains and reference-calibrated error counts, pairing a smooth grid-level surrogate with last-step backpropagation~\cite{clark2024draft} and velocity anchoring for alignment.

\section{Preliminaries}
\label{sec:prelim}

\paragraph{Nowcasting and event scores.}
\label{sec:prelim_task}
Given $L_h$ historical radar frames $\mathbf{c}\in\mathbb{R}^{L_h\times H\times W}$, generative nowcasting models the conditional distribution $p_\theta(\mathbf{x}\mid\mathbf{c})$ of $L$ future frames $\mathbf{x}$.
A deterministic sampler maps initial noise $\boldsymbol{\epsilon}$ to a forecast $\hat{\mathbf{x}}_\theta(\mathbf{c},\boldsymbol{\epsilon})$, where varying $\boldsymbol{\epsilon}$ produces ensemble members.

At each forecast lead time $\ell$ and intensity threshold $\tau$, let $y_i=\mathbbm{1}[x_i\ge\tau]$ and $\hat{y}_i=\mathbbm{1}[\hat{x}_i\ge\tau]$ denote binary event masks at grid point $i$, where $\mathbbm{1}[\cdot]$ is the indicator function.
From hits $H=\sum_i\hat{y}_i y_i$, misses $M=\sum_i(1-\hat{y}_i)y_i$, and false alarms $F=\sum_i\hat{y}_i(1-y_i)$, the critical success index (CSI) and false alarm ratio (FAR) are defined as
\begin{equation}
\mathrm{CSI}=\frac{H}{H+M+F},\qquad
\mathrm{FAR}=\frac{F}{H+F}.
\label{eq:csi_far}
\end{equation}
CSI penalizes both misses and false alarms to measure event overlap, whereas FAR measures the false-alarm fraction among predicted events without penalizing misses, inducing a trade-off between expanding coverage to reduce $M$ and restricting predictions to lower FAR~\cite{jolliffe2012forecast}.

\paragraph{Conditional flow matching.}
\label{sec:prelim_fm}

Conditional flow matching~\cite{lipman2023flow,liu2023rectified} transports noise $\boldsymbol{\epsilon}\sim\pi_{\mathbf{c}}$ (independent of $\mathbf{x}$ given $\mathbf{c}$) to $\mathbf{x}$ along $\mathbf{x}_t=t\mathbf{x}+(1-t)\boldsymbol{\epsilon}$ for $t\in[0,1]$ (distinct from lead time $\ell$), training a velocity network $v_\theta(\mathbf{x}_t,t,\mathbf{c})$ via
\begin{equation}
\mathcal{L}_{\mathrm{FM}}(\theta)=\mathbb{E}_{\mathbf{c},\mathbf{x},\boldsymbol{\epsilon},t}\,\big\|v_\theta(\mathbf{x}_t,t,\mathbf{c})-(\mathbf{x}-\boldsymbol{\epsilon})\big\|^2 .
\label{eq:fm}
\end{equation}
Over $0=t_0<\dots<t_N=1$, Euler sampling initializes $\mathbf{x}_{t_0}=\boldsymbol{\epsilon}$ and iterates $\mathbf{x}_{t_{n+1}}=\mathbf{x}_{t_n}+(t_{n+1}-t_n)\,v_\theta(\mathbf{x}_{t_n},t_n,\mathbf{c})$ to yield the forecast $\hat{\mathbf{x}}_\theta(\mathbf{c},\boldsymbol{\epsilon})=\mathbf{x}_{t_N}$.

\begin{figure}[!tb]
\centering
\includegraphics[width=\linewidth]{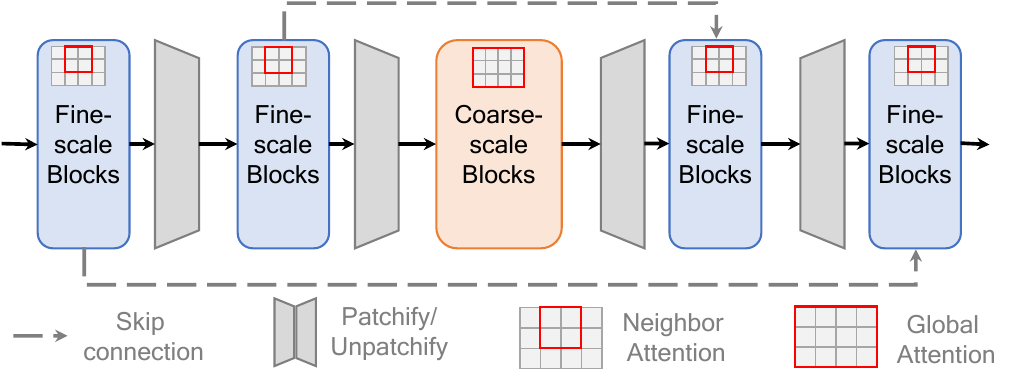}
\vspace{-4ex}
\caption{\textbf{Pixel-space nowcaster backbone.}}
\label{fig:pixel_nowcaster}
\vspace{-3ex}
\end{figure}

\section{Method}
\label{sec:method}

AlignPre comprises an end-to-end pixel-space generative nowcaster (\textbf{AlignPre-Base}, Section~\ref{sec:method_base}) followed by skill-aligned post-training (\textbf{SAPT}, Section~\ref{sec:method_post}).

\subsection{Pixel-space generative nowcaster}
\label{sec:method_base}

AlignPre-Base formulates conditional flow matching directly in radar pixel space, integrating representation compression and field reconstruction into a single generative network without separate autoencoder pretraining~\cite{li2026jit,vallaeys2026ujit}.
Because high-dimensional precipitation sequences concentrate near a structured low-dimensional manifold, we follow JiT~\cite{li2026jit} and adopt the \emph{x-prediction} parameterization, where the network $f_\theta(\mathbf{x}_t,t,\mathbf{c})$ predicts the clean future sequence and induces the velocity field
\begin{equation}
v_\theta(\mathbf{x}_t,t,\mathbf{c})=\frac{f_\theta(\mathbf{x}_t,t,\mathbf{c})-\mathbf{x}_t}{1-t},\qquad t<1.
\label{eq:xpred}
\end{equation}

Pretraining minimizes Equation~\eqref{eq:fm} under a noise-shifted timestep schedule~\cite{hoogeboom2023simple,esser2024scaling} (Appendix~\ref{sec:app_alignpre_implementation}).
Under $N$-step Euler integration to $t_N=1$, substituting Equation~\eqref{eq:xpred} into the final update gives
\begin{equation}
\hat{\mathbf{x}}_\theta(\mathbf{c},\boldsymbol{\epsilon})=\mathbf{x}_{t_N}=f_\theta(\mathbf{x}_{t_{N-1}},t_{N-1},\mathbf{c}).
\label{eq:last_step}
\end{equation}
Thus, the terminal radar forecast is directly the final network output without external decoding, exposing the output mapping to end-to-end skill alignment.

\begin{figure}[t]
\centering
\begin{minipage}[t]{0.495\linewidth}
\centering
\includegraphics[width=\linewidth]{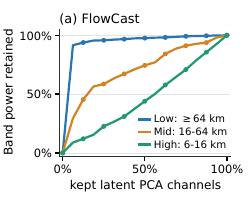}
\end{minipage}\hfill%
\begin{minipage}[t]{0.495\linewidth}
\centering
\includegraphics[width=\linewidth]{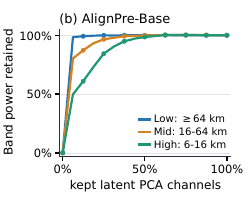}
\end{minipage}
\vspace{-3ex}
\caption{\textbf{Channel-PCA truncation across output scales.} (a) FlowCast latent space; (b) AlignPre-Base bottleneck tokens. Curves show forecast band power relative to each model's full output across wavelengths.}
\label{fig:representation_bands}
\vspace{-3ex}
\end{figure}

\paragraph{Hierarchical spatiotemporal architecture.}
To capture both localized echo evolution and large-scale storm organization at tractable computational cost, we instantiate $f_\theta$ as a U-shaped hierarchical spatiotemporal transformer~\cite{vallaeys2026ujit} (Figure~\ref{fig:pixel_nowcaster}).
High-resolution blocks alternate between intra-frame neighborhood attention and per-location temporal attention~\cite{hassani2023neighborhood,bertasius2021timesformer}; after spatial token merging, coarse-scale bottleneck blocks alternate between intra-frame global attention and temporal attention, followed by upsampling and resolution-matched skip connections to restore the full grid.
Both resolution levels apply 3D rotary position embeddings in physical coordinates to encode spatial displacements and temporal intervals~\cite{su2024roformer}.
Historical observations $\mathbf{c}$ are encoded once, with their key--value (K/V) cache reused across all steps of a sampling trajectory.

\paragraph{Spectral representation analysis.}
To compare multi-scale feature organization under end-to-end pixel-space training versus decoupled latent compression~\cite{rombach2022ldm,schusterbauer2026freud}, Figure~\ref{fig:representation_bands} applies channel-wise PCA truncation to FlowCast's final latents and AlignPre-Base's bottleneck tokens before upsampling. Retaining the top 25\% principal components in FlowCast (Figure~\ref{fig:representation_bands}(a)) preserves 95.7\% of low-frequency ($\ge 64$\,km) power but only 22.6\% of high-frequency (6--16\,km) power, indicating that fine-scale synthesis relies heavily on trailing latent components and a frozen decoder. Conversely, the top 25\% components in AlignPre-Base (Figure~\ref{fig:representation_bands}(b)) retain 84.4\% of high-frequency power alongside coarser bands, confirming that our pixel-space backbone compactly encodes multi-scale structure within a single end-to-end adaptable network.

\subsection{Skill-aligned post-training}
\label{sec:method_post}

\begin{figure}[!tb]
\centering
\includegraphics[width=\linewidth]{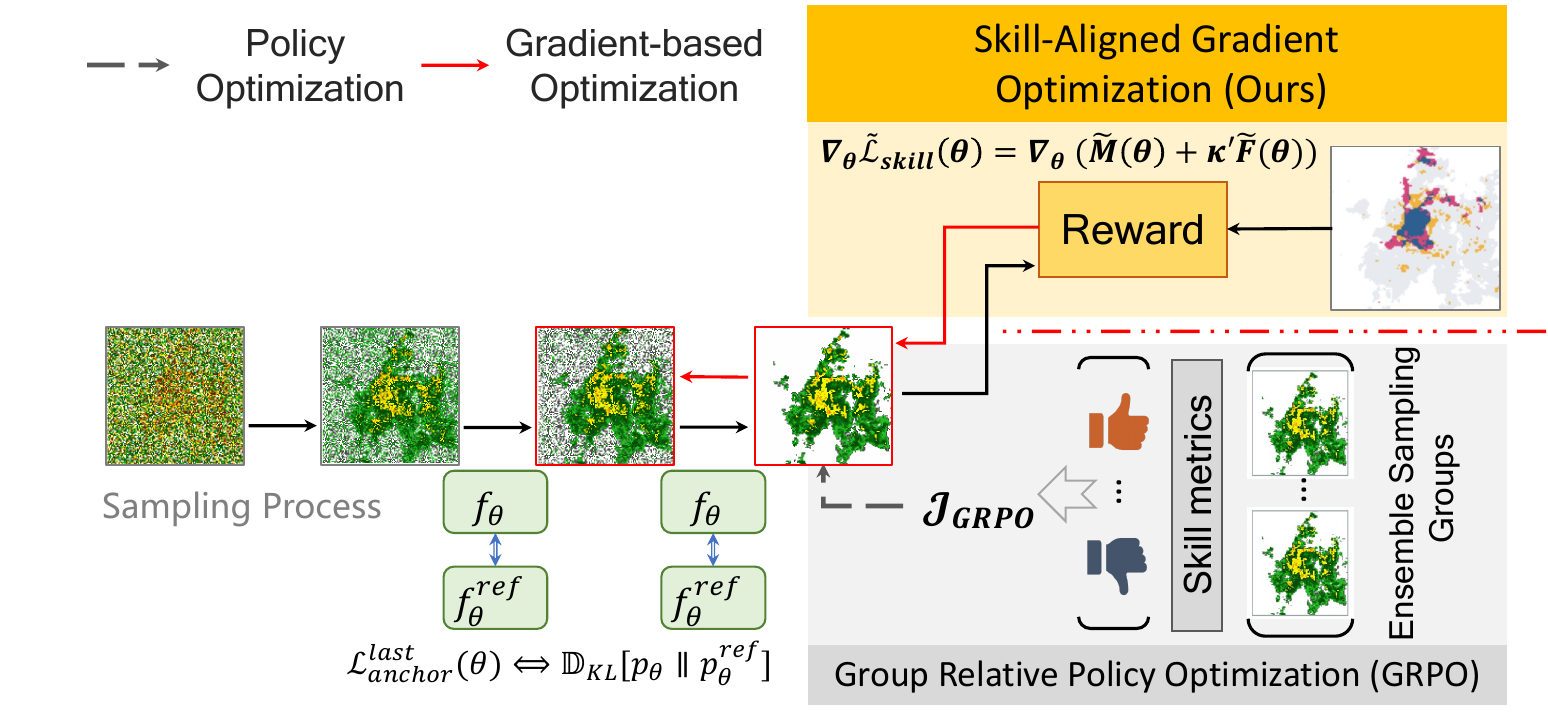}
\vspace{-3ex}
\caption{\textbf{Overview of SAPT and GRPO.} SAPT backpropagates grid-level gradients of a smooth skill objective through the final sampling step (red arrows), whereas GRPO updates the policy via group-relative advantages from scalar rewards (gray arrows). }
\label{fig:posttrain_overview}
\vspace{-3ex}
\end{figure}

To jointly optimize CSI and FAR without multi-rollout policy gradients~\cite{xu2026nowcastdit,liu2025flowgrpo,li2026mixgrpo} or two-stage preference tuning~\cite{xu2026syncast}, \emph{skill-aligned post-training} (SAPT) exploits the analytic structure of contingency scores to form a unified differentiable objective, backpropagating grid-level gradients directly from predicted radar fields~\cite{clark2024draft,li2025pisa} (Figure~\ref{fig:posttrain_overview}).

\paragraph{Optimization target.}
\label{sec:post_target}

For a given lead time and intensity threshold $\tau$, let $H^{\mathrm{ref}},M^{\mathrm{ref}},F^{\mathrm{ref}}$ denote the expected per-sample event counts of a frozen reference model $f^{\mathrm{ref}}_\theta$.
Because the expected positive count $P=H+M$ is fixed by ground truth, CSI and FAR depend on the generator solely through expected misses $M$ and false alarms $F$.
Rearranging Equation~\eqref{eq:csi_far} shows that improving either score over $f^{\mathrm{ref}}_\theta$ is equivalent to reducing a linear weighted error count~\cite{dinkelbach1967nonlinear, koyejo2014consistent, parambath2014fmeasure} (Appendix~\ref{sec:app_linear}):
\begin{equation}
\begin{aligned}
\mathrm{CSI}>\mathrm{CSI}^{\mathrm{ref}}
&\iff M+\kappa_1F<M^{\mathrm{ref}}+\kappa_1F^{\mathrm{ref}},\\
\mathrm{FAR}<\mathrm{FAR}^{\mathrm{ref}}
&\iff M+\kappa_2F<M^{\mathrm{ref}}+\kappa_2F^{\mathrm{ref}},
\end{aligned}
\label{eq:linear}
\end{equation}
where
\begin{equation}
\begin{aligned}
\kappa_1&=\frac{H^{\mathrm{ref}}}{P+F^{\mathrm{ref}}},
\quad 
\kappa_2&=\frac{H^{\mathrm{ref}}}{F^{\mathrm{ref}}}
.
\end{aligned}
\label{eq:kappa}
\end{equation}

Equations~\eqref{eq:linear}--\eqref{eq:kappa} reveal that both improvement criteria share the same linear form $M+\kappa F$, differing only in the false-alarm weight $\kappa$.
Selecting a weight $\kappa$ (calibrated in Section~\ref{sec:post_cost_calibration}) and dropping the $\theta$-independent reference constant $M^{\mathrm{ref}}+\kappa F^{\mathrm{ref}}$ turns score alignment into minimizing the expected weighted error
\begin{equation}
J_\kappa(\theta)=M(\theta)+\kappa F(\theta).
\label{eq:cost}
\end{equation}
Crucially, unlike the dataset-level ratios in Equation~\eqref{eq:csi_far}, $J_\kappa(\theta)$ is linear in $M$ and $F$ and thus decomposes into $J_\kappa(\theta)=\mathbb E[\ell_\kappa(\theta)]$ over per-sample, grid-level penalties (Appendix~\ref{sec:app_target}):
\begin{equation}
\ell_\kappa(\theta)
=\sum_i\big[y_i(1-\hat y_i(\theta))
+\kappa(1-y_i)\hat y_i(\theta)\big].
\label{eq:hard_loss}
\end{equation}

Since binary thresholding $\hat y_i=\mathbbm{1}[\hat x_i\ge\tau]$ has zero gradients almost everywhere, we relax predicted events to $s_i=\sigma((\hat x_i-\tau)/T)$ with temperature $T>0$ and adopt the squared surrogate~\cite{choi2026atloss}:
\begin{equation}
\widetilde\ell_\kappa(\theta)
=\underbrace{\sum_i y_i(1-s_i)^2}_{\widetilde M(\theta)}
+\kappa\underbrace{\sum_i(1-y_i)s_i^2}_{\widetilde F(\theta)}.
\label{eq:soft_loss}
\end{equation}
This surrogate matches Equation~\eqref{eq:hard_loss} on binary predictions while providing dense intensity gradients on both observed event ($y_i=1$) and non-event ($y_i=0$) regions.

\paragraph{Calibration of \texorpdfstring{$\kappa$}{kappa}.}
\label{sec:post_cost_calibration}

\begin{figure}[!tb]
\centering
\includegraphics[width=\linewidth]{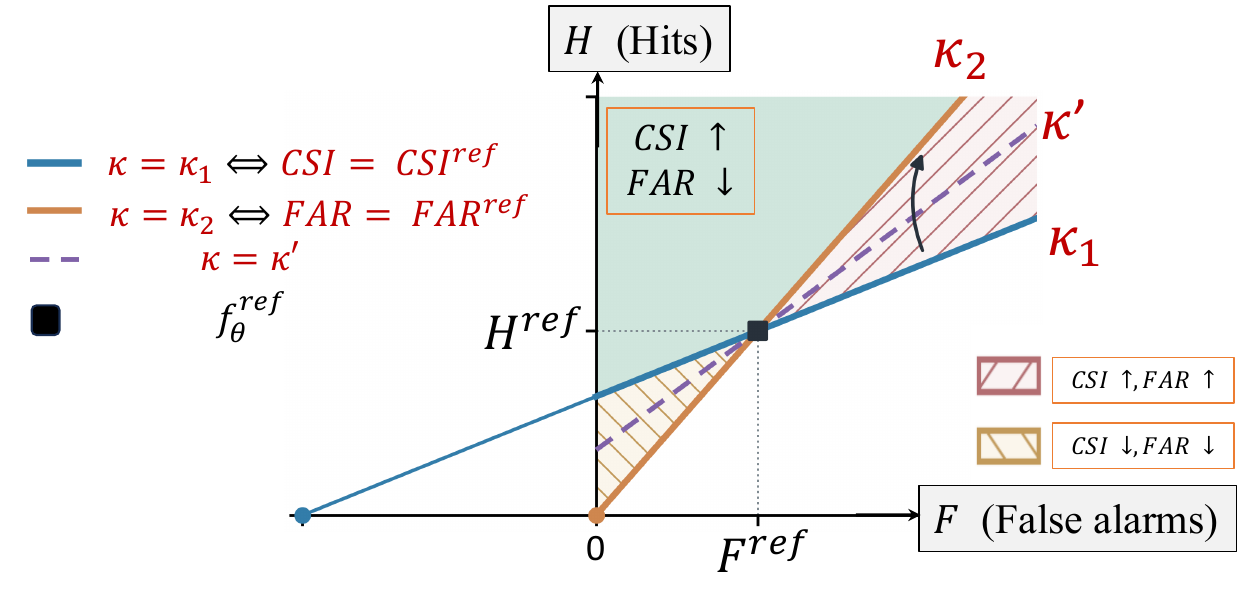}
\vspace{-5ex}
\caption{Geometric calibration of the false-alarm weight in the $(F, H)$ plane.}
\label{fig:hf_plane}
\vspace{-4ex}
\end{figure}

Because $\kappa_2>\kappa_1$ in Equation~\eqref{eq:kappa}, FAR penalizes false alarms more heavily than CSI; optimizing with $\kappa=\kappa_1$ alone thus targets CSI without curbing false-alarm growth.
Taking a convex combination of the two criteria in Equation~\eqref{eq:linear} restricts $\kappa\in[\kappa_1,\kappa_2]$.
Geometrically, increasing $\kappa$ from $\kappa_1$ toward $\kappa_2$ rotates the linear cost contour toward the FAR contour in the $(F,H)$ plane (Figure~\ref{fig:hf_plane}), strengthening false-alarm suppression while ensuring that any cost reduction over $f^{\mathrm{ref}}_\theta$ improves at least one score.

\begin{figure*}[!t]
\centering
\includegraphics[width=\textwidth]{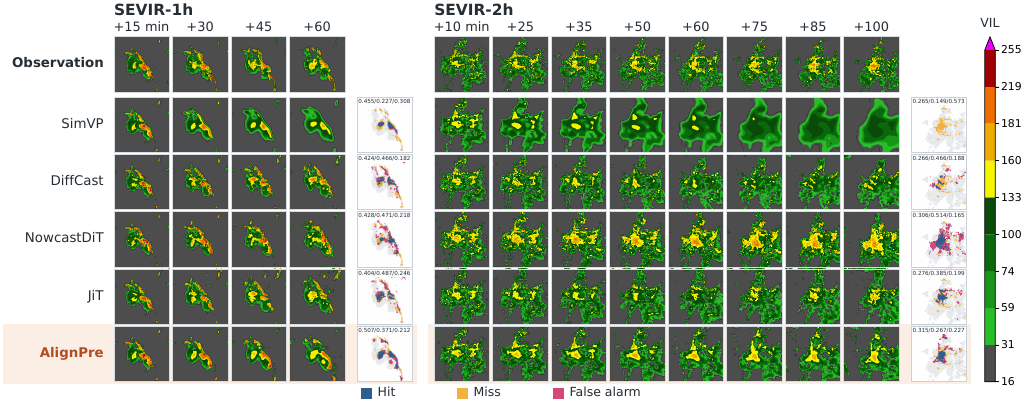}
\vspace{-5ex}
\caption{\textbf{Qualitative forecasts on SEVIR-1h (left) and SEVIR-2h (right).} Final columns mark final-lead hits, misses, and false alarms at VIL $\geq133$. Numbers report case-level CSI-M/FAR-M/LPIPS over all leads.}
\label{fig:cases_sevir}
\end{figure*}

\begin{figure*}[!tp]
\centering
\includegraphics[width=\textwidth]{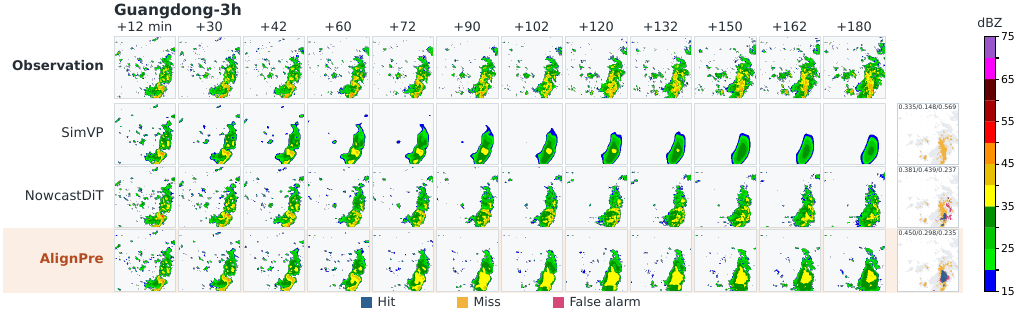}
\vspace{-5ex}
\caption{\textbf{Qualitative forecasts on Guangdong-3h.} The final column marks hits, misses, and false alarms at +180\,min and $\geq35$\,dBZ.}
\label{fig:cases_gd}
\end{figure*}

Across lead times $\ell$ and thresholds $\tau_k$, $\kappa_{1,\ell,k}= H^{\mathrm{ref}}_{\ell,k}/(P_k+F^{\mathrm{ref}}_{\ell,k})$ captures how the miss--false-alarm trade-off varies with lead time, but its magnitude drops sharply at high thresholds where events are rare.
To maintain consistent false-alarm control across thresholds while preserving within-threshold lead-time proportions, we normalize $\kappa_{1,\ell,k}$ by its lead-time peak, scale by a shared target peak $\rho$ (comparable to the unit miss cost when $\rho\approx1$), and clip to $[\kappa_{1,\ell,k},\kappa_{2,\ell,k}]$:
\begin{equation}
\kappa'_{\ell,k}
=\operatorname{clip}_{[\kappa_{1,\ell,k},\,\kappa_{2,\ell,k}]}
\left(
\rho\,\frac{\kappa_{1,\ell,k}}
{\max_{\ell'}\kappa_{1,\ell',k}}
\right).
\label{eq:kappa_fixed}
\end{equation}
Finally, because event counts span orders of magnitude from widespread light rain to sparse severe storms, summing raw errors across lead-time--threshold groups $g=(\ell,k)\in G$ would let low thresholds dominate training.
Since weighted-error changes shift group $g$'s CSI in proportion to $1/(P_g+F_g^{\mathrm{ref}})$ (Appendix~\ref{sec:app_linear}), we normalize each group by this fixed reference denominator so all terms match the scale of their CSI changes:
\begin{equation}
\mathcal L_{\mathrm{skill}}(\theta)
=\frac{1}{|G|}\sum_{g\in G}
\frac{\widetilde M_g(\theta)+\kappa'_g\widetilde F_g(\theta)}
{P_g+F_g^{\mathrm{ref}}}.
\label{eq:skill}
\end{equation}
All reference constants are calibrated once offline on a subset $\mathcal D_{\mathrm{cal}}\subset\mathcal D$ and held fixed during post-training (see Appendix~\ref{sec:app_target} for degenerate cases).

\paragraph{Post-training optimization.}
\label{sec:post_optimization}
\label{sec:post_procedure}

Backpropagating through all $N$ Euler steps incurs substantial memory overhead.
Following~\cite{clark2024draft}, we roll out the first $N-1$ steps without tracking gradients to obtain $\bar{\mathbf x}_*=\operatorname{sg}[\mathbf x_{t_*}]$ at $t_*=t_{N-1}$, and differentiate only through the final step $\hat{\mathbf x}=f_\theta(\bar{\mathbf x}_*,t_*,\mathbf c)$ in Equation~\eqref{eq:last_step}.

Optimizing $\mathcal L_{\mathrm{skill}}$ alone does not require forecasts to retain dependence on initial noise $\boldsymbol\epsilon$, risking ensemble collapse (Appendix~\ref{sec:app_collapse}).
Motivated by KL-regularized fine-tuning~\cite{fan2023dpok,domingoenrich2025adjoint,cover2006elements} (Appendix~\ref{sec:app_anchor}), we penalize the final-step velocity deviation from $f^{\mathrm{ref}}_\theta$ at $\bar{\mathbf x}_*$, which by Equation~\eqref{eq:xpred} yields
\begin{equation}
\mathcal L_{\mathrm{anchor}}^{\mathrm{last}}(\theta)
=\frac{\big\|f_\theta(\bar{\mathbf x}_*,t_*,\mathbf c)-\operatorname{sg}[f^{\mathrm{ref}}_\theta(\bar{\mathbf x}_*,t_*,\mathbf c)]\big\|^2}
{(1-t_*)^2}.
\label{eq:anchor}
\end{equation}
Combining both terms gives the post-training objective
\begin{equation}
\mathcal L(\theta)
=\mathbb E_{(\mathbf c,\mathbf x)\sim\mathcal D,\,\boldsymbol\epsilon}
\left[\mathcal L_{\mathrm{skill}}(\theta)
+\eta\mathcal L_{\mathrm{anchor}}^{\mathrm{last}}(\theta)\right],
\label{eq:total}
\end{equation}
where $\eta>0$ controls the anchor strength.
Each update requires only one sampled trajectory per input condition $\mathbf c$, and last-step gradients update all shared parameters $\theta$ used across subsequent rollouts.
The complete procedure of our algorithm is described in Appendix~\ref{sec:app_alignpre_implementation}.

\section{Experiments}
\label{sec:exp}

\setlength{\textfloatsep}{9pt plus 2pt minus 2pt}
\setlength{\floatsep}{7pt plus 2pt minus 2pt}
\setlength{\intextsep}{8pt plus 2pt minus 2pt}
\setlength{\dbltextfloatsep}{10pt plus 2pt minus 2pt}
\setlength{\dblfloatsep}{8pt plus 2pt minus 2pt}
\setcounter{topnumber}{4}
\setcounter{bottomnumber}{3}
\setcounter{totalnumber}{6}
\setcounter{dbltopnumber}{3}
\renewcommand{\topfraction}{0.94}
\renewcommand{\bottomfraction}{0.90}
\renewcommand{\textfraction}{0.05}
\renewcommand{\dbltopfraction}{0.94}
\renewcommand{\floatpagefraction}{0.80}
\renewcommand{\dblfloatpagefraction}{0.80}

\begin{table*}[!tp]
\centering
\caption{\textbf{Results on SEVIR-1h and SEVIR-2h.} Deterministic results are gray. Among the remaining models, best and second-best values per column are bold and underlined, respectively. Better deterministic values are also bold.}
\label{tab:main_sevir}
\scriptsize
\setlength{\tabcolsep}{1.7pt}
\renewcommand{\arraystretch}{1.08}
\begin{tabular}{l*{7}{c}@{\hspace{8pt}}l*{7}{c}}
\toprule
\multicolumn{8}{c}{SEVIR-1h (13$\to$12 frames)} & \multicolumn{8}{c}{SEVIR-2h (5$\to$20 frames)}\\
\cmidrule(r{4pt}){1-8}\cmidrule(l{4pt}){9-16}
Method & CSI-M$^{\uparrow}$ & CSI$_{219}^{\uparrow}$ & POD-M$^{\uparrow}$ & FAR-M$^{\downarrow}$ & HSS-M$^{\uparrow}$ & LPIPS$^{\downarrow}$ & CRPS$^{\downarrow}$ & Method & CSI-M$^{\uparrow}$ & CSI$_{219}^{\uparrow}$ & POD-M$^{\uparrow}$ & FAR-M$^{\downarrow}$ & HSS-M$^{\uparrow}$ & LPIPS$^{\downarrow}$ & CRPS$^{\downarrow}$ \\
\midrule
PhyDNet~\cite{guen2020phydnet} & \textcolor{gray}{0.3948} & \textcolor{gray}{0.0904} & \textcolor{gray}{0.4632} & \textcolor{gray}{\fst{0.3054}} & \textcolor{gray}{0.4915} & \textcolor{gray}{0.3464} & \textcolor{gray}{6.616} & PhyDNet~\cite{guen2020phydnet} & \textcolor{gray}{0.2552} & \textcolor{gray}{0.0063} & \textcolor{gray}{0.3122} & \textcolor{gray}{\fst{0.3574}} & \textcolor{gray}{0.3125} & \textcolor{gray}{0.3816} & \textcolor{gray}{8.270} \\
SimVP~\cite{gao2022simvp} & \textcolor{gray}{0.4027} & \textcolor{gray}{0.0876} & \textcolor{gray}{0.4708} & \textcolor{gray}{\fst{0.2698}} & \textcolor{gray}{0.4996} & \textcolor{gray}{0.3503} & \textcolor{gray}{6.324} & SimVP~\cite{gao2022simvp} & \textcolor{gray}{0.2673} & \textcolor{gray}{0.0240} & \textcolor{gray}{0.3270} & \textcolor{gray}{\fst{0.2633}} & \textcolor{gray}{0.3283} & \textcolor{gray}{0.3830} & \textcolor{gray}{8.138} \\
Earthformer~\cite{gao2022earthformer} & \textcolor{gray}{0.4290} & \textcolor{gray}{0.1338} & \textcolor{gray}{0.5113} & \textcolor{gray}{\fst{0.3330}} & \textcolor{gray}{0.5407} & \textcolor{gray}{0.3564} & \textcolor{gray}{6.373} & Earthformer~\cite{gao2022earthformer} & \textcolor{gray}{0.2424} & \textcolor{gray}{0.0089} & \textcolor{gray}{0.3031} & \textcolor{gray}{\fst{0.2786}} & \textcolor{gray}{0.2966} & \textcolor{gray}{0.4138} & \textcolor{gray}{9.417} \\
NowcastNet~\cite{zhang2023nowcastnet} & \textcolor{gray}{0.4203} & \textcolor{gray}{0.1407} & \textcolor{gray}{0.5360} & \textcolor{gray}{0.4148} & \textcolor{gray}{0.5317} & \textcolor{gray}{\fst{0.1917}} & \textcolor{gray}{7.312} & NowcastNet~\cite{zhang2023nowcastnet} & \textcolor{gray}{0.2994} & \textcolor{gray}{0.0592} & \textcolor{gray}{\fst{0.4370}} & \textcolor{gray}{0.6236} & \textcolor{gray}{0.3816} & \textcolor{gray}{0.1688} & \textcolor{gray}{9.620} \\
\addlinespace[2pt]
DiffCast~\cite{yu2024diffcast} & 0.3795 & 0.1197 & 0.4903 & 0.4812 & 0.4878 & \snd{0.2259} & 5.112 & DiffCast~\cite{yu2024diffcast} & 0.2713 & 0.0481 & 0.3455 & 0.5995 & 0.3490 & 0.1950 & 6.001 \\
CasCast~\cite{gong2024cascast} & 0.4168 & 0.1556 & \snd{0.5732} & 0.4716 & 0.5340 & 0.2563 & 5.653 & CasCast~\cite{gong2024cascast} & 0.2634 & 0.0326 & 0.3696 & 0.6464 & 0.3404 & 0.1858 & 6.704 \\
\addlinespace[2pt]
FlowCast~\cite{ribeiro2026flowcast} & \snd{0.4367} & \snd{0.1772} & 0.5403 & \snd{0.3941} & \snd{0.5562} & 0.2329 & \fst{4.892} & FlowCast~\cite{ribeiro2026flowcast} & 0.2902 & 0.0601 & 0.3600 & 0.5598 & 0.3734 & \snd{0.1581} & \fst{5.919} \\
NowcastDiT~\cite{xu2026nowcastdit} & 0.3948 & 0.1338 & 0.5580 & 0.5047 & 0.5077 & 0.2516 & 5.831 & NowcastDiT~\cite{xu2026nowcastdit} & \snd{0.2999} & 0.0572 & \fst{0.5025} & 0.6465 & \snd{0.3854} & \fst{0.1549} & 7.102 \\
\addlinespace[2pt]
JiT~\cite{li2026jit} & 0.3473 & 0.1052 & 0.4717 & 0.5074 & 0.4485 & 0.3184 & 6.616 & JiT~\cite{li2026jit} & 0.2817 & 0.0553 & 0.3636 & 0.5852 & 0.3628 & 0.2002 & 6.523 \\
\addlinespace[2pt]
PreDiff~\cite{gao2023prediff} & 0.3653 & 0.1102 & 0.5071 & 0.5187 & 0.4734 & 0.2832 & 5.873 & DuoCast~\cite{wen2026duocast} & 0.2818 & \snd{0.0619} & 0.3556 & 0.5816 & 0.3594 & 0.2208 & 6.656 \\
FREUD~\cite{schusterbauer2026freud} & 0.3646 & 0.1093 & 0.4575 & 0.4807 & 0.4684 & \fst{0.2256} & 5.249 & AlphaPre~\cite{lin2025alphapre} & \textcolor{gray}{0.2765} & \textcolor{gray}{0.0293} & \textcolor{gray}{0.3252} & \textcolor{gray}{\fst{0.3484}} & \textcolor{gray}{0.3414} & \textcolor{gray}{0.3372} & \textcolor{gray}{7.580} \\
\midrule
AlignPre-Base & 0.4129 & 0.1482 & 0.5104 & 0.4140 & 0.5281 & 0.2411 & \snd{4.906} & AlignPre-Base & 0.2925 & 0.0549 & 0.3570 & \snd{0.5407} & 0.3738 & 0.1888 & \snd{5.935} \\
\rowcolor{oursrow}AlignPre & \fst{0.4605} & \fst{0.2041} & \fst{0.5747} & \fst{0.3755} & \fst{0.5836} & 0.2344 & 5.355 & AlignPre & \fst{0.3169} & \fst{0.0728} & \snd{0.3890} & \fst{0.4984} & \fst{0.4076} & 0.1886 & 6.285 \\
\bottomrule
\end{tabular}
\end{table*}

\subsection{Experimental setup}
\label{sec:exp_setup}

\paragraph{Benchmarks.}
We evaluate AlignPre on three radar benchmarks spanning 1- to 3-hour lead times. 
SEVIR~\cite{veillette2020sevir} provides 5\,min vertically integrated liquid (VIL) sequences evaluated at thresholds $\{16,74,133,160,181,219\}$. Following standard protocols, \emph{SEVIR-1h} predicts $13\to12$ frames (60\,min) at $384\times384$~\cite{gao2022earthformer,gong2024cascast} and \emph{SEVIR-2h} predicts $5\to20$ frames (100\,min, about 2 hours) at $128\times128$~\cite{yu2024diffcast,lin2025alphapre,xu2026nowcastdit}.
\emph{Guangdong-3h} uses 6\,min composite radar reflectivity over Guangdong, China, predicting $10\to30$ frames (3\,h) at $128\times128$ across $\{20,30,35,40\}$\,dBZ. More details are in Appendix~\ref{sec:app_alignpre_implementation}.

\paragraph{Baselines.}
We compare four model families:
(i) deterministic predictors (PhyDNet~\cite{guen2020phydnet}, SimVP~\cite{gao2022simvp}, Earthformer~\cite{gao2022earthformer}, NowcastNet~\cite{zhang2023nowcastnet}, AlphaPre~\cite{lin2025alphapre});
(ii) hybrid generators (DiffCast~\cite{yu2024diffcast}, CasCast~\cite{gong2024cascast}, DuoCast~\cite{wen2026duocast});
(iii) latent generators (PreDiff~\cite{gao2023prediff}, FlowCast~\cite{ribeiro2026flowcast}, FREUD~\cite{schusterbauer2026freud}, NowcastDiT~\cite{xu2026nowcastdit}); and
(iv) JiT~\cite{li2026jit}, adapted as a single-scale pixel-space DiT.

\paragraph{Metrics.}
For operational verification, we aggregate test-set contingency counts at each lead time and threshold to compute CSI, probability of detection (POD), FAR, and Heidke skill score (HSS)~\cite{jolliffe2012forecast}, reporting macro-averages over both dimensions (suffix -M) and top-threshold skill (CSI$_{219}$ and CSI$_{40}$).
We also report LPIPS~\cite{zhang2018lpips} for perceptual quality, and CRPS~\cite{gneiting2007proper} for ensemble calibration.

\paragraph{Implementation details.}
We pretrain AlignPre-Base for 100 epochs using the AdamW~\cite{loshchilov2019adamw} optimizer with a learning rate of $1\times 10^{-4}$, followed by full-parameter SAPT for one epoch with a learning rate of $1\times 10^{-5}$.
For flow-matching sampling, both SAPT rollouts and inference use $N=10$ Euler steps with classifier-free guidance~\cite{ho2022cfg} at $w=1.6$ .
We set default post-training hyperparameters in Equations~\eqref{eq:kappa_fixed}--\eqref{eq:anchor} correspondingly across benchmarks.
All experiments are conducted on A6000 GPUs.

\begin{table}[!tb]
\centering
\caption{\textbf{Results on Guangdong-3h.} Conventions follow Table~\ref{tab:main_sevir}.}
\label{tab:main_gd}
\scriptsize
\setlength{\tabcolsep}{1.4pt}
\renewcommand{\arraystretch}{1.05}
\begin{tabular}{l*{7}{c}}
\toprule
Method & CSI-M$^{\uparrow}$ & CSI$_{40}^{\uparrow}$ & POD-M$^{\uparrow}$ & FAR-M$^{\downarrow}$ & HSS-M$^{\uparrow}$ & LPIPS$^{\downarrow}$ & CRPS$^{\downarrow}$ \\
\midrule
PhyDNet~\cite{guen2020phydnet} & \textcolor{gray}{0.2297} & \textcolor{gray}{0.0643} & \textcolor{gray}{0.2589} & \textcolor{gray}{\fst{0.3443}} & \textcolor{gray}{0.3135} & \textcolor{gray}{0.5502} & \textcolor{gray}{4.000} \\
SimVP~\cite{gao2022simvp} & \textcolor{gray}{0.2390} & \textcolor{gray}{0.0691} & \textcolor{gray}{0.2719} & \textcolor{gray}{\fst{0.3213}} & \textcolor{gray}{0.3250} & \textcolor{gray}{0.5537} & \textcolor{gray}{4.119} \\
Earthformer~\cite{gao2022earthformer} & \textcolor{gray}{0.1622} & \textcolor{gray}{0.0260} & \textcolor{gray}{0.1803} & \textcolor{gray}{\fst{0.2899}} & \textcolor{gray}{0.2278} & \textcolor{gray}{0.6091} & \textcolor{gray}{4.202} \\
NowcastNet~\cite{zhang2023nowcastnet} & \textcolor{gray}{0.2586} & \textcolor{gray}{0.1183} & \textcolor{gray}{\fst{0.4913}} & \textcolor{gray}{0.6715} & \textcolor{gray}{0.3396} & \textcolor{gray}{0.3522} & \textcolor{gray}{6.899} \\
\addlinespace[2pt]
DiffCast~\cite{yu2024diffcast} & 0.2550 & 0.1104 & 0.3579 & 0.5942 & 0.3489 & 0.2537 & 2.735 \\
CasCast~\cite{gong2024cascast} & 0.2419 & 0.0954 & 0.3885 & 0.6545 & 0.3292 & 0.2846 & 3.279 \\
\addlinespace[2pt]
FlowCast~\cite{ribeiro2026flowcast} & 0.2808 & 0.1192 & 0.3664 & \snd{0.5191} & 0.3849 & \snd{0.2380} & \snd{2.667} \\
NowcastDiT~\cite{xu2026nowcastdit} & \snd{0.3003} & \snd{0.1419} & \fst{0.4540} & 0.5763 & \snd{0.4078} & \fst{0.2379} & 2.836 \\
\addlinespace[2pt]
JiT~\cite{li2026jit} & 0.2599 & 0.1164 & 0.3836 & 0.6084 & 0.3526 & 0.2652 & 2.835 \\
\midrule
AlignPre-Base & 0.2778 & 0.1225 & 0.3577 & 0.5192 & 0.3792 & 0.2656 & \fst{2.661} \\
\rowcolor{oursrow}AlignPre & \fst{0.3277} & \fst{0.1787} & \snd{0.4411} & \fst{0.4926} & \fst{0.4438} & 0.2661 & 2.763 \\
\bottomrule
\end{tabular}
\end{table}

\raggedbottom
\subsection{Main results}
\label{sec:exp_main}

\begin{figure}[!tb]
\centering
\captionsetup{skip=2pt}
\includegraphics[width=\linewidth]{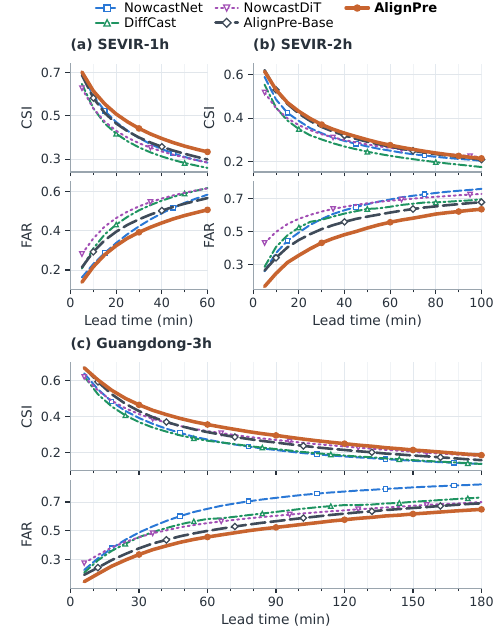}
\caption{\textbf{Lead-time CSI and FAR.} (a) SEVIR-1h, (b) SEVIR-2h, and (c) Guangdong-3h.}
\label{fig:leadtime}
\vspace{-2ex}
\end{figure}

\paragraph{Quantitative comparisons.}
Across all three 1--3\,h benchmarks, AlignPre achieves state-of-the-art overall and extreme-precipitation CSI and HSS with the lowest FAR among probabilistic generators (Tables~\ref{tab:main_sevir} and~\ref{tab:main_gd}).
Deterministic predictors (gray) keep FAR low via mean regression but blur fine structures and miss intense cores; among generators, hybrid, latent, and pixel (JiT) models sharpen textures at the cost of inflated FAR, especially under black-box RL.
Without deterministic priors or external autoencoders, AlignPre-Base already outperforms most hybrid models with low FAR and top-tier CRPS.
Adding classifier-free guidance and SAPT further boosts mean and top-threshold CSI and drives FAR below all generative baselines at negligible LPIPS cost; the mild CRPS rise reflects sharpening discrete event boundaries over continuous spread.
Across lead times under matched guidance (Figure~\ref{fig:leadtime}), AlignPre suppresses FAR throughout the rollout and sustains clear CSI gains at long horizons (Appendix~\ref{sec:app_exp_main}).

\paragraph{Qualitative comparisons.}
On SEVIR convective events (Figure~\ref{fig:cases_sevir}), SimVP severely blurs rainbands, while DiffCast and GRPO-tuned NowcastDiT hallucinate scattered intense cells that cause extensive perimeter false alarms; in contrast, AlignPre synthesizes cohesive, accurately positioned cores with realistic internal structure. On Guangdong-3h (Figure~\ref{fig:cases_gd}), baseline generators prematurely dissipate or displace the leading squall line, whereas AlignPre maintains organized high-reflectivity bands through $+180$\,min with balanced hit--false-alarm coverage.

\subsection{Comparison of post-training methods}
\label{sec:exp_posttrain}

\begin{figure}[!htbp]
\centering
\renewcommand{\thesubfigure}{\Alph{subfigure}}
\captionsetup{skip=2pt}
\captionsetup[subfigure]{skip=1pt,justification=centering}
\begin{subfigure}{\linewidth}
\centering
\includegraphics[width=\linewidth]{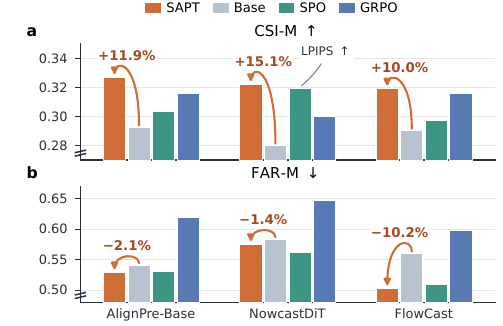}
\caption{Post-training scores.}
\label{fig:posttrain_bars}
\end{subfigure}
\par\vspace{2pt}
\begin{subfigure}{\linewidth}
\centering
\includegraphics[width=\linewidth]{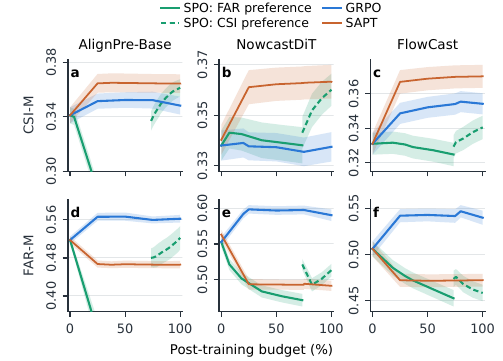}
\caption{Validation progress.}
\label{fig:posttrain_curves}
\end{subfigure}
\caption{\textbf{Post-training comparison on SEVIR-2h.} (A) Arrows give relative changes. (B) Validation progress of post-training.}
\label{fig:posttrain_comparison}
\end{figure}

\paragraph{Post-training across generators.}
\label{sec:exp_posttrain_comparison}
We compare SAPT, GRPO~\cite{xu2026nowcastdit,li2026mixgrpo}, and two-stage SPO~\cite{xu2026syncast} on AlignPre-Base, NowcastDiT, and FlowCast from identical checkpoints in test skill (Figure~\ref{fig:posttrain_bars}) and validation progress (Figure~\ref{fig:posttrain_curves}). Under matched guidance, SAPT converges rapidly and improves CSI-M and FAR-M across pixel and latent bases, peaking on AlignPre-Base as end-to-end pixel synthesis avoids frozen latent-decoder bottlenecks. Conversely, standard-reward GRPO boosts CSI-M by over-forecasting, inflating FAR-M; even with our weighted-error reward, trajectory-level policy gradients degrade LPIPS more than grid-level SAPT (Appendix~\ref{sec:app_exp_posttrain}). Meanwhile, FAR-then-CSI SPO improves both scores, resuming CSI tuning from the best rather than final FAR checkpoint (disconnected Stage-2 curves in Figure~\ref{fig:posttrain_curves}), yet severe LPIPS degradation. Full protocols and fixed-$\kappa'$ repeats are in Appendix~\ref{sec:app_exp_posttrain}.

\begin{figure}[!htbp]
\centering
\captionsetup{skip=2pt}
\captionsetup[subfigure]{skip=1pt,justification=centering}
\begin{subfigure}{\linewidth}
\centering
\includegraphics[width=\linewidth]{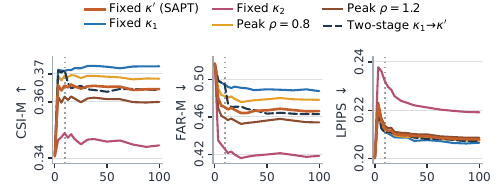}
\caption{False-alarm cost strategies.}
\label{fig:kappa_curves}
\end{subfigure}
\par\vspace{1pt}
\begin{subfigure}{\linewidth}
\centering
\includegraphics[width=\linewidth]{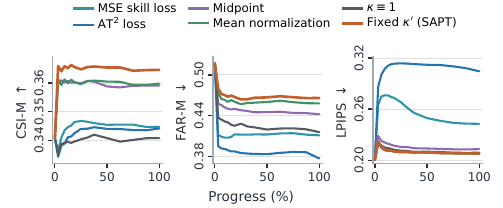}
\caption{Skill loss and normalization.}
\label{fig:skill_loss_curves}
\end{subfigure}
\caption{\textbf{Skill-loss design on SEVIR-2h.}}
\label{fig:skill_loss_design}
\end{figure}

\paragraph{Skill-loss design.}
\label{sec:exp_objective_design}
Figure~\ref{fig:kappa_curves} and Figure~\ref{fig:skill_loss_curves} evaluate cost calibration and group normalization, respectively. Fixed $\kappa_1$ favors CSI-M with elevated FAR-M, whereas fixed $\kappa_2$ over-penalizes coverage, lowering CSI-M and worsening LPIPS. Our calibrated $\kappa'$ balances both scores—matching a two-stage $\kappa_1 \to \kappa'$ curriculum without scheduling overhead—while varying $\rho$ provides smooth CSI--FAR control. Alternative objectives ($\kappa \equiv 1$, MSE, and AT$^2$~\cite{choi2026atloss}) lower CSI-M and inflate LPIPS, while reference-denominator normalization outperforms mean and midpoint scaling by balancing gradients across events (Appendix~\ref{sec:app_kappa_control}).

\subsection{Analysis and discussion}
\label{sec:exp_ablation}

\begin{figure}[!tb]
\centering
\includegraphics[width=\linewidth]{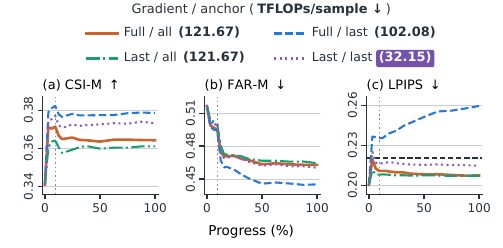}
\vspace{-4ex}
\caption{\textbf{Gradient and anchor coverage curves.} Legend: gradient/anchor scope, with denoiser \textbf{TFLOPs/sample}.}
\label{fig:gradient_anchor_curves}
\end{figure}

\paragraph{Gradient depth and anchor coverage.}
Figure~\ref{fig:gradient_anchor_curves} tracks validation progress and denoiser cost across gradient/anchor scopes.
Restricting both to the final step (Last/last, 32.15\,TFLOPs/sample) cuts compute by 73.6\% over Full/all (121.67), converging stably to higher CSI-M and slightly lower FAR-M at a modest LPIPS offset.
Retaining all-step anchoring (Last/all, 121.67) saves no compute and caps CSI-M, whereas unanchored early gradients (Full/last, 102.08) trigger continual LPIPS drift despite event-score gains.
These dynamics support our matched final-step design under $x$-prediction (Appendix~\ref{sec:app_gradient_anchor}).

\begin{figure}[!t]
\centering
\includegraphics[width=\linewidth]{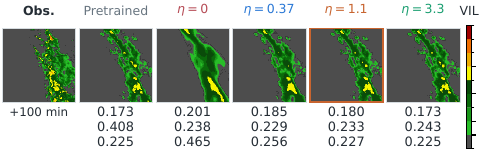}
\vspace{-2ex}
\caption{\textbf{Anchor strength at +100\,min.} Scores below each forecast are frame-level CSI-M/FAR-M/LPIPS (top to bottom). }
\label{fig:anchor_case}
\end{figure}

\begin{figure}[!t]
\centering
\captionsetup{skip=2pt}
\includegraphics[width=\linewidth]{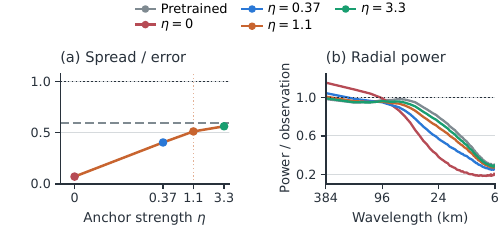}
\vspace{-1ex}
\caption{\textbf{Anchor diagnostics.} (a) Corrected spread/ensemble-mean RMSE. (b) Radial power relative to observations.}
\label{fig:ablation_anchor}
\end{figure}

\paragraph{Anchor strength.}
Velocity anchoring prevents reward hacking via mode collapse. Without it ($\eta=0$), forecasts gain CSI/FAR yet blur into oversmoothed fields with poor LPIPS (Figure~\ref{fig:anchor_case}), as ensemble calibration (spread-to-error, Figure~\ref{fig:ablation_anchor}(a)) and sub-$96$~km spectral power (Figure~\ref{fig:ablation_anchor}(b)) plummet. Increasing $\eta$ restores dispersion at stable mean RMSE and recovers fine-scale power; our default $\eta=1.1$ preserves pretrained diversity, spectrum, and LPIPS while securing clear CSI/FAR gains (Appendix~\ref{sec:app_ablation_anchor}).

\begin{figure}[!t]
\centering
\captionsetup{skip=2pt}
\includegraphics[width=\linewidth]{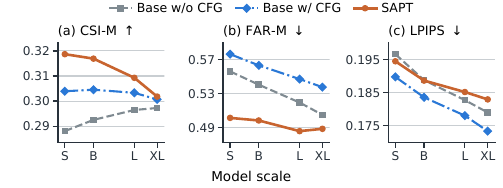}
\caption{\textbf{Model scale.} The parameter axis is logarithmic.}
\label{fig:ablation_scaling}
\vspace{-2ex}
\end{figure}

\paragraph{Model scale.}
Scaling base models monotonically improves CSI-M, FAR-M, and LPIPS, whereas classifier-free guidance alone yields diminishing CSI-M returns at larger capacities while inflating FAR-M (Figure~\ref{fig:ablation_scaling}). Across all scales, SAPT consistently boosts CSI-M and lowers FAR-M over guided bases, enabling compact models to surpass larger pretrained counterparts. Although CSI-M gains narrow as the largest model approaches predictability limits, SAPT maintains substantial false-alarm reductions across capacities (Appendix~\ref{sec:app_ablation_scaling}).

\section{Conclusion}
\label{sec:conclusion}
We introduced AlignPre, an end-to-end pixel-space generative nowcaster coupled with skill-aligned post-training (SAPT). Removing deterministic priors and frozen autoencoders, AlignPre-Base models multi-scale storm evolution directly in radar space via clean-data flow matching. SAPT further translates operational contingency scores into a reference-calibrated differentiable objective, backpropagating grid-level gradients through the final sampling step with velocity anchoring for efficient, single-rollout alignment. Across three 1- to 3-hour benchmarks, AlignPre consistently improves overall and extreme-precipitation CSI while curbing false alarms and preserving perceptual quality, offering a principled post-training paradigm for operational meteorological forecasting.


{
    \small
    \bibliographystyle{ieeenat_fullname}
    \bibliography{main}

\begin{thebibliography}{59}
\providecommand{\natexlab}[1]{#1}
\providecommand{\url}[1]{\texttt{#1}}
\expandafter\ifx\csname urlstyle\endcsname\relax
  \providecommand{\doi}[1]{doi: #1}\else
  \providecommand{\doi}{doi: \begingroup \urlstyle{rm}\Url}\fi

\bibitem[Alet et~al.(2025)Alet, Price, El-Kadi, Masters, Markou, Andersson, Stott, Lam, Willson, Sanchez-Gonzalez, and Battaglia]{alet2025fgn}
Ferran Alet, Ilan Price, Andrew El-Kadi, Dominic Masters, Stratis Markou, Tom~R. Andersson, Jacklynn Stott, Remi Lam, Matthew Willson, Alvaro Sanchez-Gonzalez, and Peter Battaglia.
\newblock Skillful joint probabilistic weather forecasting from marginals.
\newblock \emph{arXiv preprint arXiv:2506.10772}, 2025.

\bibitem[Bertasius et~al.(2021)Bertasius, Wang, and Torresani]{bertasius2021timesformer}
Gedas Bertasius, Heng Wang, and Lorenzo Torresani.
\newblock Is space-time attention all you need for video understanding?
\newblock In \emph{Proceedings of the 38th International Conference on Machine Learning}, pages 813--824. PMLR, 2021.

\bibitem[Black et~al.(2024)Black, Janner, Du, Kostrikov, and Levine]{black2024ddpo}
Kevin Black, Michael Janner, Yilun Du, Ilya Kostrikov, and Sergey Levine.
\newblock Training diffusion models with reinforcement learning.
\newblock In \emph{International Conference on Learning Representations}, pages 4965--4987, 2024.

\bibitem[Chen et~al.(2026)Chen, Zhu, Chen, Zhang, Hu, Zhao, Wang, Yang, and Tai]{chen2026dip}
Zhennan Chen, Junwei Zhu, Xu Chen, Jiangning Zhang, Xiaobin Hu, Hanzhen Zhao, Chengjie Wang, Jian Yang, and Ying Tai.
\newblock {DiP}: Taming diffusion models in pixel space.
\newblock In \emph{Proceedings of the IEEE/CVF Conference on Computer Vision and Pattern Recognition}, pages 36136--36146, 2026.

\bibitem[Choi et~al.(2026)Choi, Kim, Kim, and Lee]{choi2026atloss}
Jaeho Choi, Hyeri Kim, Kwang-Ho Kim, and Jaesung Lee.
\newblock Advanced torrential loss function for precipitation forecasting.
\newblock \emph{Physical Review Letters}, 136\penalty0 (2):\penalty0 024201, 2026.

\bibitem[Clark et~al.(2024)Clark, Vicol, Swersky, and Fleet]{clark2024draft}
Kevin Clark, Paul Vicol, Kevin Swersky, and David~J. Fleet.
\newblock Directly fine-tuning diffusion models on differentiable rewards.
\newblock In \emph{International Conference on Learning Representations}, 2024.

\bibitem[Cover and Thomas(2006)]{cover2006elements}
Thomas~M. Cover and Joy~A. Thomas.
\newblock \emph{Elements of Information Theory}.
\newblock Wiley, 2nd edition, 2006.

\bibitem[Crowson et~al.(2024)Crowson, Baumann, Birch, Abraham, Kaplan, and Shippole]{crowson2024hourglass}
Katherine Crowson, Stefan~Andreas Baumann, Alex Birch, Tanishq~Mathew Abraham, Daniel~Z. Kaplan, and Enrico Shippole.
\newblock Scalable high-resolution pixel-space image synthesis with hourglass diffusion transformers.
\newblock In \emph{Proceedings of the 41st International Conference on Machine Learning}, pages 9550--9575. PMLR, 2024.

\bibitem[Dinkelbach(1967)]{dinkelbach1967nonlinear}
Werner Dinkelbach.
\newblock On nonlinear fractional programming.
\newblock \emph{Management Science}, 13\penalty0 (7):\penalty0 492--498, 1967.

\bibitem[Domingo-Enrich et~al.(2025)Domingo-Enrich, Drozdzal, Karrer, and Chen]{domingoenrich2025adjoint}
Carles Domingo-Enrich, Michal Drozdzal, Brian Karrer, and Ricky T.~Q. Chen.
\newblock Adjoint matching: Fine-tuning flow and diffusion generative models with memoryless stochastic optimal control.
\newblock In \emph{International Conference on Learning Representations}, pages 53791--53846, 2025.

\bibitem[Esser et~al.(2024)Esser, Kulal, Blattmann, Entezari, M{\"u}ller, Saini, Levi, Lorenz, Sauer, Boesel, Podell, Dockhorn, English, and Rombach]{esser2024scaling}
Patrick Esser, Sumith Kulal, Andreas Blattmann, Rahim Entezari, Jonas M{\"u}ller, Harry Saini, Yam Levi, Dominik Lorenz, Axel Sauer, Frederic Boesel, Dustin Podell, Tim Dockhorn, Zion English, and Robin Rombach.
\newblock Scaling rectified flow transformers for high-resolution image synthesis.
\newblock In \emph{Proceedings of the 41st International Conference on Machine Learning}, pages 12606--12633. PMLR, 2024.

\bibitem[Fan et~al.(2023)Fan, Watkins, Du, Liu, Ryu, Boutilier, Abbeel, Ghavamzadeh, Lee, and Lee]{fan2023dpok}
Ying Fan, Olivia Watkins, Yuqing Du, Hao Liu, Moonkyung Ryu, Craig Boutilier, Pieter Abbeel, Mohammad Ghavamzadeh, Kangwook Lee, and Kimin Lee.
\newblock {DPOK}: Reinforcement learning for fine-tuning text-to-image diffusion models.
\newblock In \emph{Advances in Neural Information Processing Systems}, pages 79858--79885, 2023.

\bibitem[Feng et~al.(2025)Feng, Li, Wu, Lin, Yu, Ye, and Wang]{feng2025perceptual}
Wenzhi Feng, Xutao Li, Zhe Wu, Kenghong Lin, Demin Yu, Yunming Ye, and Yaowei Wang.
\newblock Perceptually constrained precipitation nowcasting model.
\newblock In \emph{Proceedings of the 42nd International Conference on Machine Learning}, pages 16777--16795. PMLR, 2025.

\bibitem[Gao et~al.(2022{\natexlab{a}})Gao, Shi, Wang, Zhu, Wang, Li, and Yeung]{gao2022earthformer}
Zhihan Gao, Xingjian Shi, Hao Wang, Yi Zhu, Yuyang Wang, Mu Li, and Dit-Yan Yeung.
\newblock {Earthformer}: Exploring space-time transformers for earth system forecasting.
\newblock In \emph{Advances in Neural Information Processing Systems}, pages 25390--25403, 2022{\natexlab{a}}.

\bibitem[Gao et~al.(2022{\natexlab{b}})Gao, Tan, Wu, and Li]{gao2022simvp}
Zhangyang Gao, Cheng Tan, Lirong Wu, and Stan~Z. Li.
\newblock {SimVP}: Simpler yet better video prediction.
\newblock In \emph{Proceedings of the IEEE/CVF Conference on Computer Vision and Pattern Recognition}, pages 3170--3180, 2022{\natexlab{b}}.

\bibitem[Gao et~al.(2023)Gao, Shi, Han, Wang, Jin, Maddix, Zhu, Li, and Wang]{gao2023prediff}
Zhihan Gao, Xingjian Shi, Boran Han, Hao Wang, Xiaoyong Jin, Danielle Maddix, Yi Zhu, Mu Li, and Yuyang Wang.
\newblock {PreDiff}: Precipitation nowcasting with latent diffusion models.
\newblock In \emph{Advances in Neural Information Processing Systems}, pages 78621--78656, 2023.

\bibitem[Gneiting and Raftery(2007)]{gneiting2007proper}
Tilmann Gneiting and Adrian~E. Raftery.
\newblock Strictly proper scoring rules, prediction, and estimation.
\newblock \emph{Journal of the American Statistical Association}, 102\penalty0 (477):\penalty0 359--378, 2007.

\bibitem[Gong et~al.(2024)Gong, Bai, Ye, Xu, Liu, Dai, Yang, and Ouyang]{gong2024cascast}
Junchao Gong, Lei Bai, Peng Ye, Wanghan Xu, Na Liu, Jianhua Dai, Xiaokang Yang, and Wanli Ouyang.
\newblock {CasCast}: Skillful high-resolution precipitation nowcasting via cascaded modelling.
\newblock In \emph{Proceedings of the 41st International Conference on Machine Learning}, pages 15809--15822. PMLR, 2024.

\bibitem[Hassani et~al.(2023)Hassani, Walton, Li, Li, and Shi]{hassani2023neighborhood}
Ali Hassani, Steven Walton, Jiachen Li, Shen Li, and Humphrey Shi.
\newblock Neighborhood attention transformer.
\newblock In \emph{Proceedings of the IEEE/CVF Conference on Computer Vision and Pattern Recognition}, pages 6185--6194, 2023.

\bibitem[Ho and Salimans(2022)]{ho2022cfg}
Jonathan Ho and Tim Salimans.
\newblock Classifier-free diffusion guidance.
\newblock \emph{arXiv preprint arXiv:2207.12598}, 2022.

\bibitem[Hoogeboom et~al.(2023)Hoogeboom, Heek, and Salimans]{hoogeboom2023simple}
Emiel Hoogeboom, Jonathan Heek, and Tim Salimans.
\newblock {simple diffusion}: End-to-end diffusion for high resolution images.
\newblock In \emph{Proceedings of the 40th International Conference on Machine Learning}, pages 13213--13232. PMLR, 2023.

\bibitem[Hoogeboom et~al.(2025)Hoogeboom, Mensink, Heek, Lamerigts, Gao, and Salimans]{hoogeboom2025sid2}
Emiel Hoogeboom, Thomas Mensink, Jonathan Heek, Kay Lamerigts, Ruiqi Gao, and Tim Salimans.
\newblock Simpler diffusion: 1.5 {FID} on {ImageNet512} with pixel-space diffusion.
\newblock In \emph{Proceedings of the IEEE/CVF Conference on Computer Vision and Pattern Recognition}, pages 18062--18071, 2025.

\bibitem[Jacq et~al.(2026)Jacq, Couairon, Berthet, De~Bortoli, Doucet, and Elie]{jacq2026rmmd}
Alexis Jacq, Guillaume Couairon, Quentin Berthet, Valentin De~Bortoli, Arnaud Doucet, and Romuald Elie.
\newblock Diffusion fine-tuning with rewarded moment matching distillation.
\newblock In \emph{Advances in Neural Information Processing Systems}, 2026.
\newblock To appear.

\bibitem[Jolliffe and Stephenson(2012)]{jolliffe2012forecast}
Ian~T. Jolliffe and David~B. Stephenson, editors.
\newblock \emph{Forecast Verification: A Practitioner's Guide in Atmospheric Science}.
\newblock John Wiley \& Sons, 2nd edition, 2012.

\bibitem[Koyejo et~al.(2014)Koyejo, Natarajan, Ravikumar, and Dhillon]{koyejo2014consistent}
Oluwasanmi Koyejo, Nagarajan Natarajan, Pradeep Ravikumar, and Inderjit~S. Dhillon.
\newblock Consistent binary classification with generalized performance metrics.
\newblock In \emph{Advances in Neural Information Processing Systems}, 2014.

\bibitem[Lang et~al.(2026)Lang, Alexe, Clare, Roberts, Adewoyin, Ben~Bouall{\`e}gue, Chantry, Dramsch, Dueben, Hahner, Maciel, Prieto-Nemesio, O'Brien, Pinault, Polster, Raoult, Tietsche, and Leutbecher]{lang2024aifscrps}
Simon Lang, Mihai Alexe, Mariana C.~A. Clare, Christopher Roberts, Rilwan Adewoyin, Zied Ben~Bouall{\`e}gue, Matthew Chantry, Jesper Dramsch, Peter~D. Dueben, Sara Hahner, Pedro Maciel, Ana Prieto-Nemesio, Cathal O'Brien, Florian Pinault, Jan Polster, Baudouin Raoult, Steffen Tietsche, and Martin Leutbecher.
\newblock {AIFS-CRPS}: Ensemble forecasting using a model trained with a loss function based on the continuous ranked probability score.
\newblock \emph{npj Artificial Intelligence}, 2:\penalty0 18, 2026.

\bibitem[Le~Guen and Thome(2020)]{guen2020phydnet}
Vincent Le~Guen and Nicolas Thome.
\newblock Disentangling physical dynamics from unknown factors for unsupervised video prediction.
\newblock In \emph{Proceedings of the IEEE/CVF Conference on Computer Vision and Pattern Recognition}, pages 11474--11484, 2020.

\bibitem[Leinonen et~al.(2023)Leinonen, Hamann, Nerini, Germann, and Franch]{leinonen2023ldcast}
Jussi Leinonen, Ulrich Hamann, Daniele Nerini, Urs Germann, and Gabriele Franch.
\newblock Latent diffusion models for generative precipitation nowcasting with accurate uncertainty quantification.
\newblock \emph{arXiv preprint arXiv:2304.12891}, 2023.

\bibitem[Li et~al.(2025)Li, Michel, Pan, Liu, Roberts, and Xie]{li2025pisa}
Chenyu Li, Oscar Michel, Xichen Pan, Sainan Liu, Mike Roberts, and Saining Xie.
\newblock {PISA} experiments: Exploring physics post-training for video diffusion models by watching stuff drop.
\newblock In \emph{Proceedings of the 42nd International Conference on Machine Learning}, pages 35685--35709. PMLR, 2025.

\bibitem[Li et~al.(2026)Li, Cui, Huang, Kong, Zeng, Ma, Fan, Yang, Zhong, and Bo]{li2026mixgrpo}
Junzhe Li, Yutao Cui, Tao Huang, Weijie Kong, Chuxuan Zeng, Yinping Ma, Chun Fan, Miles Yang, Zhao Zhong, and Liefeng Bo.
\newblock {MixGRPO}: Unlocking flow-based {GRPO} efficiency with mixed {ODE-SDE}.
\newblock In \emph{Computer Vision -- ECCV 2026}, pages 231--248. Springer, 2026.

\bibitem[Li and He(2026)]{li2026jit}
Tianhong Li and Kaiming He.
\newblock Back to basics: Let denoising generative models denoise.
\newblock In \emph{Proceedings of the IEEE/CVF Conference on Computer Vision and Pattern Recognition}, pages 36115--36125, 2026.

\bibitem[Lin et~al.(2025)Lin, Zhang, Yu, Feng, Chen, Gao, Li, and Ye]{lin2025alphapre}
Kenghong Lin, Baoquan Zhang, Demin Yu, Wenzhi Feng, Shidong Chen, Feifan Gao, Xutao Li, and Yunming Ye.
\newblock {AlphaPre}: Amplitude-phase disentanglement model for precipitation nowcasting.
\newblock In \emph{Proceedings of the IEEE/CVF Conference on Computer Vision and Pattern Recognition}, pages 17841--17850, 2025.

\bibitem[Lipman et~al.(2023)Lipman, Chen, Ben-Hamu, Nickel, and Le]{lipman2023flow}
Yaron Lipman, Ricky T.~Q. Chen, Heli Ben-Hamu, Maximilian Nickel, and Matthew Le.
\newblock Flow matching for generative modeling.
\newblock In \emph{International Conference on Learning Representations}, 2023.

\bibitem[Liu et~al.(2025)Liu, Liu, Liang, Li, Liu, Wang, Wan, Zhang, and Ouyang]{liu2025flowgrpo}
Jie Liu, Gongye Liu, Jiajun Liang, Yangguang Li, Jiaheng Liu, Xintao Wang, Pengfei Wan, Di Zhang, and Wanli Ouyang.
\newblock {Flow-GRPO}: Training flow matching models via online {RL}.
\newblock In \emph{Advances in Neural Information Processing Systems}, pages 40783--40818, 2025.

\bibitem[Liu et~al.(2023)Liu, Gong, and Liu]{liu2023rectified}
Xingchao Liu, Chengyue Gong, and Qiang Liu.
\newblock Flow straight and fast: Learning to generate and transfer data with rectified flow.
\newblock In \emph{International Conference on Learning Representations}, 2023.

\bibitem[Loshchilov and Hutter(2019)]{loshchilov2019adamw}
Ilya Loshchilov and Frank Hutter.
\newblock Decoupled weight decay regularization.
\newblock In \emph{International Conference on Learning Representations}, 2019.

\bibitem[Nagashima et~al.(2026)Nagashima, Bannai, Misaizu, Maeda, Ogawa, and Haseyama]{nagashima2026pgfmm}
Shunya Nagashima, Takumi Bannai, Makoto Misaizu, Keisuke Maeda, Takahiro Ogawa, and Miki Haseyama.
\newblock Physics-guided flow-map matching for precipitation nowcasting.
\newblock In \emph{Proceedings of the Asian Conference on Computer Vision}, 2026.
\newblock To appear.

\bibitem[Parambath et~al.(2014)Parambath, Usunier, and Grandvalet]{parambath2014fmeasure}
Shameem A.~Puthiya Parambath, Nicolas Usunier, and Yves Grandvalet.
\newblock Optimizing {F}-measures by cost-sensitive classification.
\newblock In \emph{Advances in Neural Information Processing Systems}, 2014.

\bibitem[Perrone~Ribeiro and Faganeli~Pucer(2026)]{ribeiro2026flowcast}
Bernardo Perrone~Ribeiro and Jana Faganeli~Pucer.
\newblock {FlowCast}: Advancing precipitation nowcasting with conditional flow matching.
\newblock In \emph{International Conference on Learning Representations}, pages 156863--156883, 2026.

\bibitem[Prabhudesai et~al.(2024)Prabhudesai, Mendonca, Qin, Fragkiadaki, and Pathak]{prabhudesai2024vader}
Mihir Prabhudesai, Russell Mendonca, Zheyang Qin, Katerina Fragkiadaki, and Deepak Pathak.
\newblock Video diffusion alignment via reward gradients.
\newblock \emph{arXiv preprint arXiv:2407.08737}, 2024.

\bibitem[Ravuri et~al.(2021)Ravuri, Lenc, Willson, Kangin, Lam, Mirowski, Fitzsimons, Athanassiadou, Kashem, Madge, Prudden, Mandhane, Clark, Brock, Simonyan, Hadsell, Robinson, Clancy, Arribas, and Mohamed]{ravuri2021dgmr}
Suman Ravuri, Karel Lenc, Matthew Willson, Dmitry Kangin, Remi Lam, Piotr Mirowski, Megan Fitzsimons, Maria Athanassiadou, Sheleem Kashem, Sam Madge, Rachel Prudden, Amol Mandhane, Aidan Clark, Andrew Brock, Karen Simonyan, Raia Hadsell, Niall Robinson, Ellen Clancy, Alberto Arribas, and Shakir Mohamed.
\newblock Skilful precipitation nowcasting using deep generative models of radar.
\newblock \emph{Nature}, 597\penalty0 (7878):\penalty0 672--677, 2021.

\bibitem[Roebber(2009)]{roebber2009performance}
Paul~J. Roebber.
\newblock Visualizing multiple measures of forecast quality.
\newblock \emph{Weather and Forecasting}, 24\penalty0 (2):\penalty0 601--608, 2009.

\bibitem[Rombach et~al.(2022)Rombach, Blattmann, Lorenz, Esser, and Ommer]{rombach2022ldm}
Robin Rombach, Andreas Blattmann, Dominik Lorenz, Patrick Esser, and Bj{\"o}rn Ommer.
\newblock High-resolution image synthesis with latent diffusion models.
\newblock In \emph{Proceedings of the IEEE/CVF Conference on Computer Vision and Pattern Recognition}, pages 10684--10695, 2022.

\bibitem[Schusterbauer et~al.(2026)Schusterbauer, Wiese, Stracke, Phan, and Ommer]{schusterbauer2026freud}
Johannes Schusterbauer, Jannik Wiese, Nick Stracke, Timy Phan, and Bj{\"o}rn Ommer.
\newblock Probabilistic precipitation nowcasting with rectified flow transformers.
\newblock In \emph{Proceedings of the IEEE/CVF Conference on Computer Vision and Pattern Recognition}, pages 25742--25756, 2026.

\bibitem[Shi et~al.(2015)Shi, Chen, Wang, Yeung, Wong, and Woo]{shi2015convlstm}
Xingjian Shi, Zhourong Chen, Hao Wang, Dit-Yan Yeung, Wai-kin Wong, and Wang-chun Woo.
\newblock Convolutional {LSTM} network: A machine learning approach for precipitation nowcasting.
\newblock In \emph{Advances in Neural Information Processing Systems}, 2015.

\bibitem[Song et~al.(2021)Song, Sohl-Dickstein, Kingma, Kumar, Ermon, and Poole]{song2021score}
Yang Song, Jascha Sohl-Dickstein, Diederik~P. Kingma, Abhishek Kumar, Stefano Ermon, and Ben Poole.
\newblock Score-based generative modeling through stochastic differential equations.
\newblock In \emph{International Conference on Learning Representations}, 2021.

\bibitem[Su et~al.(2024)Su, Ahmed, Lu, Pan, Bo, and Liu]{su2024roformer}
Jianlin Su, Murtadha Ahmed, Yu Lu, Shengfeng Pan, Wen Bo, and Yunfeng Liu.
\newblock {RoFormer}: Enhanced transformer with rotary position embedding.
\newblock \emph{Neurocomputing}, 568:\penalty0 127063, 2024.

\bibitem[Vallaeys et~al.(2026)Vallaeys, Verbeek, and Cord]{vallaeys2026ujit}
Th{\'e}ophane Vallaeys, Jakob Verbeek, and Matthieu Cord.
\newblock Pixel-space video generation with {U-JiT}.
\newblock \emph{arXiv preprint arXiv:2611.04764}, 2026.

\bibitem[Veillette et~al.(2020)Veillette, Samsi, and Mattioli]{veillette2020sevir}
Mark Veillette, Siddharth Samsi, and Chris Mattioli.
\newblock {SEVIR}: A storm event imagery dataset for deep learning applications in radar and satellite meteorology.
\newblock In \emph{Advances in Neural Information Processing Systems}, pages 22009--22019, 2020.

\bibitem[Wallace et~al.(2024)Wallace, Dang, Rafailov, Zhou, Lou, Purushwalkam, Ermon, Xiong, Joty, and Naik]{wallace2024diffusiondpo}
Bram Wallace, Meihua Dang, Rafael Rafailov, Linqi Zhou, Aaron Lou, Senthil Purushwalkam, Stefano Ermon, Caiming Xiong, Shafiq Joty, and Nikhil Naik.
\newblock Diffusion model alignment using direct preference optimization.
\newblock In \emph{Proceedings of the IEEE/CVF Conference on Computer Vision and Pattern Recognition}, pages 8228--8238, 2024.

\bibitem[Wen et~al.(2026)Wen, He, Filippi, Zhao, Zhang, Bishop, Wang, and Hu]{wen2026duocast}
Penghui Wen, Mengwei He, Patrick Filippi, Na Zhao, Feng Zhang, Thomas~Francis Bishop, Zhiyong Wang, and Kun Hu.
\newblock {DuoCast}: Duo-probabilistic diffusion for precipitation nowcasting.
\newblock In \emph{Proceedings of the AAAI Conference on Artificial Intelligence}, pages 39442--39450, 2026.

\bibitem[Wiese et~al.(2026)Wiese, Schusterbauer, Martorella, and Ommer]{wiese2026jws}
Jannik Wiese, Johannes Schusterbauer, Tommaso Martorella, and Bj{\"o}rn Ommer.
\newblock Just weather scoring: Efficient end-to-end nowcasting with distributional diffusion.
\newblock \emph{arXiv preprint arXiv:2610.12189}, 2026.

\bibitem[Xu et~al.(2026{\natexlab{a}})Xu, Guo, Zhang, Zhong, Wang, and Long]{xu2026nowcastdit}
Haoran Xu, Xingzhuo Guo, Yuchen Zhang, Jincheng Zhong, Jianmin Wang, and Mingsheng Long.
\newblock {NowcastDiT}: Diffusion transformers are effective precipitation nowcasters.
\newblock \emph{arXiv preprint arXiv:2609.37038}, 2026{\natexlab{a}}.

\bibitem[Xu et~al.(2026{\natexlab{b}})Xu, Gong, Zhang, Fei, Bai, and Ouyang]{xu2026syncast}
Kaiyi Xu, Junchao Gong, Wenlong Zhang, Ben Fei, Lei Bai, and Wanli Ouyang.
\newblock {SynCast}: Synergizing contradictions in precipitation nowcasting via diffusion sequential preference optimization.
\newblock \emph{IEEE Transactions on Circuits and Systems for Video Technology}, 36\penalty0 (6):\penalty0 8169--8180, 2026{\natexlab{b}}.

\bibitem[Yu et~al.(2024)Yu, Li, Ye, Zhang, Luo, Dai, Wang, and Chen]{yu2024diffcast}
Demin Yu, Xutao Li, Yunming Ye, Baoquan Zhang, Chuyao Luo, Kuai Dai, Rui Wang, and Xunlai Chen.
\newblock {DiffCast}: A unified framework via residual diffusion for precipitation nowcasting.
\newblock In \emph{Proceedings of the IEEE/CVF Conference on Computer Vision and Pattern Recognition}, pages 27758--27767, 2024.

\bibitem[Yu et~al.(2026)Yu, Xiong, Nie, Sheng, Liu, and Luo]{yu2026pixeldit}
Yongsheng Yu, Wei Xiong, Weili Nie, Yichen Sheng, Shiqiu Liu, and Jiebo Luo.
\newblock {PixelDiT}: Pixel diffusion transformers for image generation.
\newblock In \emph{Proceedings of the IEEE/CVF Conference on Computer Vision and Pattern Recognition}, pages 14273--14282, 2026.

\bibitem[Zhang et~al.(2018)Zhang, Isola, Efros, Shechtman, and Wang]{zhang2018lpips}
Richard Zhang, Phillip Isola, Alexei~A. Efros, Eli Shechtman, and Oliver Wang.
\newblock The unreasonable effectiveness of deep features as a perceptual metric.
\newblock In \emph{Proceedings of the IEEE Conference on Computer Vision and Pattern Recognition}, pages 586--595, 2018.

\bibitem[Zhang et~al.(2023)Zhang, Long, Chen, Xing, Jin, Jordan, and Wang]{zhang2023nowcastnet}
Yuchen Zhang, Mingsheng Long, Kaiyuan Chen, Lanxiang Xing, Ronghua Jin, Michael~I. Jordan, and Jianmin Wang.
\newblock Skilful nowcasting of extreme precipitation with {NowcastNet}.
\newblock \emph{Nature}, 619\penalty0 (7970):\penalty0 526--532, 2023.

\bibitem[Zhu et~al.(2026)Zhu, Shi, Feng, and Niu]{zhu2026pixelflowcast}
Yufeng Zhu, Chunlei Shi, Yongchao Feng, and Dan Niu.
\newblock {PixelFlowCast}: Latent-free precipitation nowcasting via pixel mean flows.
\newblock \emph{arXiv preprint arXiv:2605.10046}, 2026.

\end{thebibliography}
}

\clearpage
\appendix
\setcounter{page}{1}
\renewcommand{\thepage}{S\arabic{page}}
\maketitlesupplementary

\renewcommand{\floatpagefraction}{.72}
\renewcommand{\dblfloatpagefraction}{.76}
\setlength{\textfloatsep}{12pt plus 2pt minus 2pt}
\setlength{\dbltextfloatsep}{12pt plus 2pt minus 2pt}

\renewcommand{\cvprsect}[1]{%
  \cvprsection[#1]{\texorpdfstring{\hskip -1em.~}{}#1}}
\renewcommand{\cvprsubsect}[1]{%
  \cvprsubsection[#1]{\texorpdfstring{\hskip -1em.~}{}#1}}

\startcontents[appendix]
\begingroup
\small
\pdfbookmark[1]{Appendix contents}{appendix.contents}
\section*{Appendix contents}
\printcontents[appendix]{}{1}[2]{}
\endgroup
\medskip

\section{Theoretical Analysis of Skill-Aligned Post-Training}
\label{sec:app_theory}

This appendix details the logical progression and mathematical derivations underlying skill-aligned post-training (SAPT) in Section~\ref{sec:method_post}.
Appendix~\ref{sec:app_linear} derives the exact linear score-improvement criteria relative to a reference generator and the reference-denominator scaling across lead-time--threshold groups.
Appendix~\ref{sec:app_target} shows how linearity yields an unbiased single-rollout grid-level estimator and specifies the boundary rules for degenerate calibration groups.
Appendix~\ref{sec:app_surrogate} analyzes the pointwise consistency and asymmetric gradient structure of the squared-sigmoid surrogate.
Appendices~\ref{sec:app_range} and~\ref{sec:app_calibration} examine the inherent CSI--FAR trade-off across $[\kappa_1, \kappa_2]$ in the $(F,H)$ plane and explain the design of peak-normalized calibration as a practical compromise across precipitation regimes.
Finally, Appendices~\ref{sec:app_collapse} and~\ref{sec:app_anchor} show why skill-only tuning induces ensemble collapse and spatial blurring, and establish the path-space KL motivation and scope matching of velocity anchoring.

\subsection{Linearization of Score Improvements and Group Scaling}
\label{sec:app_linear}

Fix a forecast lead time $\ell$ and an intensity threshold $\tau_k$, suppressing the group subscript $g=(\ell,k)$ where clear from context.
Let $(H,M,F)$ and $(H^{\mathrm{ref}},M^{\mathrm{ref}},F^{\mathrm{ref}})$ denote the expected per-sample contingency counts of a candidate generator $f_\theta$ and the frozen reference model $f_\theta^{\mathrm{ref}}$, respectively, evaluated on the same data distribution.
(Because dataset-aggregated counts over $N_{\mathrm{eval}}$ sequences equal $N_{\mathrm{eval}}$ times the empirical per-sample means, all algebraic identities below apply identically to dataset-aggregated evaluation metrics.)
Assume a positive ground-truth event count $P = H + M = H^{\mathrm{ref}} + M^{\mathrm{ref}} > 0$, and define the error increments relative to $f_\theta^{\mathrm{ref}}$:
\begin{equation}
\Delta M = M - M^{\mathrm{ref}}, \qquad
\Delta F = F - F^{\mathrm{ref}},
\label{eq:app_delta}
\end{equation}
which immediately implies $H - H^{\mathrm{ref}} = -\Delta M$.
Since $F, F^{\mathrm{ref}} \ge 0$, the CSI denominators $P + F$ and $P + F^{\mathrm{ref}}$ are strictly positive.
For FAR, we assume $F^{\mathrm{ref}} > 0$ (ensuring a finite weight $\kappa_2$) and $H + F > 0$ (ensuring well-defined candidate FAR); strict ordering of the two weights additionally uses $H^{\mathrm{ref}} > 0$.

\paragraph{Exact improvement criteria.}
Subtracting the reference CSI in Equation~\eqref{eq:csi_far} and substituting $H - H^{\mathrm{ref}} = -\Delta M$ yields
\begin{equation}
\begin{aligned}
&\mathrm{CSI} - \mathrm{CSI}^{\mathrm{ref}} \\
&\quad = \frac{H(P + F^{\mathrm{ref}}) - H^{\mathrm{ref}}(P + F)}{(P + F)(P + F^{\mathrm{ref}})} \\
&\quad = \frac{(H - H^{\mathrm{ref}})(P + F^{\mathrm{ref}}) - H^{\mathrm{ref}}\Delta F}{(P + F)(P + F^{\mathrm{ref}})} \\
&\quad = -\frac{\Delta M + \kappa_1 \Delta F}{P + F},
\end{aligned}
\label{eq:app_csi}
\end{equation}
where $\kappa_1 = H^{\mathrm{ref}} / (P + F^{\mathrm{ref}})$.
Because $P + F > 0$, the sign of $\mathrm{CSI} - \mathrm{CSI}^{\mathrm{ref}}$ is governed solely by the numerator:
\begin{equation}
\begin{aligned}
\mathrm{CSI} > \mathrm{CSI}^{\mathrm{ref}}
&\iff \Delta M + \kappa_1 \Delta F < 0 \\
&\iff M + \kappa_1 F < M^{\mathrm{ref}} + \kappa_1 F^{\mathrm{ref}}.
\end{aligned}
\label{eq:app_csi_criterion}
\end{equation}
Similarly, subtracting the reference FAR in Equation~\eqref{eq:csi_far} gives
\begin{equation}
\begin{aligned}
&\mathrm{FAR} - \mathrm{FAR}^{\mathrm{ref}} \\
&\quad = \frac{F(H^{\mathrm{ref}} + F^{\mathrm{ref}}) - F^{\mathrm{ref}}(H + F)}{(H + F)(H^{\mathrm{ref}} + F^{\mathrm{ref}})} \\
&\quad = \frac{F^{\mathrm{ref}}\Delta M + H^{\mathrm{ref}}\Delta F}{(H + F)(H^{\mathrm{ref}} + F^{\mathrm{ref}})} \\
&\quad = \frac{F^{\mathrm{ref}}(\Delta M + \kappa_2 \Delta F)}{(H + F)(H^{\mathrm{ref}} + F^{\mathrm{ref}})},
\end{aligned}
\label{eq:app_far}
\end{equation}
where $\kappa_2 = H^{\mathrm{ref}} / F^{\mathrm{ref}}$.
Since $F^{\mathrm{ref}} > 0$ and $(H + F)(H^{\mathrm{ref}} + F^{\mathrm{ref}}) > 0$,
\begin{equation}
\begin{aligned}
\mathrm{FAR} < \mathrm{FAR}^{\mathrm{ref}}
&\iff \Delta M + \kappa_2 \Delta F < 0 \\
&\iff M + \kappa_2 F < M^{\mathrm{ref}} + \kappa_2 F^{\mathrm{ref}}.
\end{aligned}
\label{eq:app_far_criterion}
\end{equation}
Equations~\eqref{eq:app_csi_criterion} and~\eqref{eq:app_far_criterion} establish Equations~\eqref{eq:linear}--\eqref{eq:kappa} as exact algebraic equivalences without local Taylor approximation~\cite{dinkelbach1967nonlinear,koyejo2014consistent,parambath2014fmeasure}.
Crucially, although CSI and FAR are nonlinear fractional scores with distinct denominators, comparing a candidate model against $f_\theta^{\mathrm{ref}}$ reduces both improvement conditions to the same linear weighted-error form $M + \kappa F$, differing only in the false-alarm weight $\kappa$.

\paragraph{Groupwise normalization by the reference denominator.}
Restoring the group index $g=(\ell,k)\in G$, Equation~\eqref{eq:app_csi} can be factored relative to the fixed reference denominator $D_g = P_g + F_g^{\mathrm{ref}}$ as
\begin{equation}
\begin{aligned}
&\mathrm{CSI}_g - \mathrm{CSI}_g^{\mathrm{ref}} \\
&\quad = -\frac{\Delta M_g + \kappa_{1,g}\Delta F_g}{P_g + F_g^{\mathrm{ref}}}
         \cdot \frac{P_g + F_g^{\mathrm{ref}}}{P_g + F_g} \\
&\quad = -\frac{\Delta M_g + \kappa_{1,g}\Delta F_g}{P_g + F_g^{\mathrm{ref}}}
         \left(1 + \mathcal{O}\!\left(\frac{|\Delta F_g|}{P_g + F_g^{\mathrm{ref}}}\right)\right).
\end{aligned}
\label{eq:app_group_scale}
\end{equation}
In radar precipitation nowcasting, $D_g$ spans several orders of magnitude from widespread stratiform rain ($\tau = 16$\,VIL) to localized convective cores ($\tau = 219$\,VIL).
Summing unnormalized error counts across $G$ would weight group gradients by raw event prevalence rather than verification skill.
As Equation~\eqref{eq:app_group_scale} shows, around $f_\theta^{\mathrm{ref}}$ ($|\Delta F_g| \ll P_g + F_g^{\mathrm{ref}}$), a given reduction in weighted error shifts $\mathrm{CSI}_g$ in direct proportion to $(P_g + F_g^{\mathrm{ref}})^{-1}$.
Dividing each group's calibrated error by $P_g + F_g^{\mathrm{ref}}$ in Equation~\eqref{eq:skill} renders every group term dimensionless and aligns its baseline gradient scale with macro-averaged CSI.

\subsection{Unbiased Grid-Level Objective and Degenerate Groups}
\label{sec:app_target}

Dropping the $\theta$-independent reference constant $M^{\mathrm{ref}} + \kappa F^{\mathrm{ref}}$ reduces score alignment to minimizing $J_\kappa(\theta) = M(\theta) + \kappa F(\theta)$ in Equation~\eqref{eq:cost}.
For $(\mathbf{c}, \mathbf{x}) \sim \mathcal{D}$ and $\boldsymbol{\epsilon} \sim \pi_{\mathbf{c}}$, let $\ell_\kappa(\theta; \mathbf{c}, \mathbf{x}, \boldsymbol{\epsilon})$ be the per-sample grid-level penalty in Equation~\eqref{eq:hard_loss}.
By linearity of expectation—requiring no spatial or temporal independence across grid entries $i$—$J_\kappa(\theta)$ decomposes into per-sample expectations, so a mini-batch of $B$ samples $\{(\mathbf{c}_b, \mathbf{x}_b, \boldsymbol{\epsilon}_b)\}_{b=1}^B$ with a single rollout per condition yields an exact unbiased estimator ($\mathbb{E}[\widehat{J}_\kappa(\theta)] = J_\kappa(\theta)$) for any $B \ge 1$:
\begin{equation}
\begin{aligned}
J_\kappa(\theta)
&= \mathbb{E}_{(\mathbf{c},\mathbf{x})\sim\mathcal{D},\,\boldsymbol{\epsilon}\sim\pi_{\mathbf{c}}}
   \big[\ell_\kappa(\theta; \mathbf{c}, \mathbf{x}, \boldsymbol{\epsilon})\big], \\
\widehat{J}_\kappa(\theta)
&= \frac{1}{B}\sum_{b=1}^B \ell_\kappa(\theta; \mathbf{c}_b, \mathbf{x}_b, \boldsymbol{\epsilon}_b).
\end{aligned}
\label{eq:app_risk}
\end{equation}
In contrast, directly evaluating nonlinear plug-in ratios such as $\widehat{H}_B / (\widehat{H}_B + \widehat{M}_B + \widehat{F}_B)$ on mini-batches introduces severe ratio bias ($\mathbb{E}[\widehat{H}_B / \widehat{D}_B] \neq \mathbb{E}[H] / \mathbb{E}[D]$), especially at high thresholds where batch-level event counts fluctuate near zero.

\paragraph{Degenerate calibration groups.}
Offline calibration estimates per-sample expected counts $P_g, H_g^{\mathrm{ref}}, F_g^{\mathrm{ref}}$ on $\mathcal{D}_{\mathrm{cal}} \subset \mathcal{D}$.
To safeguard against empty or near-empty extreme-precipitation groups in Equation~\eqref{eq:skill}, the active set $G$ retains only groups satisfying $D_g = P_g + F_g^{\mathrm{ref}} > 10^{-6}$ counts per calibration sample (setting $\mathcal{L}_{\mathrm{skill}}(\theta) = 0$ if $G = \varnothing$).
When computing reference weights in Equations~\eqref{eq:kappa} and~\eqref{eq:kappa_fixed}, we set $\kappa_{1,g} = 0$ if $D_g = 0$, remove the upper clip bound ($\kappa_{2,g} = +\infty$) if $F_g^{\mathrm{ref}} = 0$, and set $\kappa'_{\ell,k} = 0$ across all $\ell$ if $\max_{\ell'} \kappa_{1,\ell',k} = 0$ at threshold $\tau_k$.

\subsection{Smooth Squared-Sigmoid Surrogate and Gradient Structure}
\label{sec:app_surrogate}

Because the hard indicator $\hat{y}_i = \mathbbm{1}[\hat{x}_i \ge \tau]$ has zero derivative everywhere except at $\hat{x}_i = \tau$, Equation~\eqref{eq:soft_loss} relaxes $\hat{y}_i$ to $s_i = \sigma((\hat{x}_i - \tau)/T)$ with $\sigma(u) = (1 + e^{-u})^{-1}$ and temperature $T > 0$, forming the squared surrogate~\cite{choi2026atloss}:
\begin{equation}
\widetilde{\ell}_\kappa(\theta)
= \sum_i \big[y_i(1 - s_i)^2 + \kappa(1 - y_i)s_i^2\big].
\label{eq:app_soft_def}
\end{equation}
On binary values $s_i \in \{0, 1\}$, $(1 - s_i)^2 = 1 - s_i$ and $s_i^2 = s_i$, so $\widetilde{\ell}_\kappa$ coincides identically with $\ell_\kappa$.
More generally, for any forecast field $\hat{\mathbf{x}}$ with no grid entry lying exactly on the threshold boundary ($\hat{x}_i \neq \tau$ for all $i$), $\lim_{T \to 0^+} s_i = \mathbbm{1}[\hat{x}_i > \tau] = \hat{y}_i$, ensuring pointwise consistency:
\begin{equation}
\lim_{T \to 0^+} \widetilde{\ell}_\kappa(\theta) = \ell_\kappa(\theta).
\label{eq:app_zero_temp}
\end{equation}

\paragraph{Asymmetric margin-focused gradients.}
Differentiating Equation~\eqref{eq:app_soft_def} with respect to the predicted pixel intensity $\hat{x}_i$ using $\partial s_i / \partial \hat{x}_i = s_i(1 - s_i)/T$ and $y_i \in \{0, 1\}$ yields
\begin{equation}
\begin{aligned}
\frac{\partial \widetilde{\ell}_\kappa}{\partial \hat{x}_i}
&= \frac{2}{T}s_i(1 - s_i)\big[\kappa(1 - y_i)s_i - y_i(1 - s_i)\big] \\
&= \begin{cases}
   -\dfrac{2}{T}s_i(1 - s_i)^2, & y_i = 1, \\[8pt]
   +\dfrac{2\kappa}{T}s_i^2(1 - s_i), & y_i = 0.
   \end{cases}
\end{aligned}
\label{eq:app_surrogate_grad}
\end{equation}
Equation~\eqref{eq:app_surrogate_grad} explains why the squared formulation is well suited for pixel-space radar fields:
\begin{enumerate}
\item \textbf{Directional event calibration:} On observed precipitation pixels ($y_i = 1$), $\partial \widetilde{\ell}_\kappa / \partial \hat{x}_i < 0$, so gradient descent pushes $\hat{x}_i$ upward across $\tau$ to eliminate misses; on non-event pixels ($y_i = 0$), $\partial \widetilde{\ell}_\kappa / \partial \hat{x}_i > 0$, pushing $\hat{x}_i$ downward below $\tau$ with gain $\kappa$ to suppress false alarms.
\item \textbf{Error-side concentration and texture preservation:} Whereas a linear sigmoid relaxation $y_i(1 - s_i) + \kappa(1 - y_i)s_i$ produces symmetric bell-shaped gradients peaking at $\hat{x}_i = \tau$, the extra $(1 - s_i)$ factor for $y_i = 1$ shifts the peak gradient magnitude onto the missed-event side at $s_i = 1/3$ ($\hat{x}_i = \tau - T\ln 2$) while damping gradients as $e^{-2(\hat{x}_i - \tau)/T}$ once $\hat{x}_i > \tau$ (hits). Symmetrically, for $y_i = 0$, the extra $s_i$ factor shifts the peak gradient onto the false-alarm side at $s_i = 2/3$ ($\hat{x}_i = \tau + T\ln 2$) and damps updates as $e^{-2(\tau - \hat{x}_i)/T}$ on clear-air pixels ($\hat{x}_i < \tau$). Concentrating updates on misclassified margins prevents runaway intensity inflation inside storm cores and preserves realistic internal echo structure.
\end{enumerate}

\subsection{CSI--FAR Weight Trade-Off and \texorpdfstring{$(F,H)$}{(F,H)}-Plane Geometry}
\label{sec:app_range}

Under the non-degeneracy conditions of Appendix~\ref{sec:app_linear} ($P > 0, H^{\mathrm{ref}} > 0, F^{\mathrm{ref}} > 0$), comparing the two endpoint weights in Equation~\eqref{eq:kappa} gives
\begin{equation}
\begin{aligned}
\kappa_2 - \kappa_1
&= \frac{H^{\mathrm{ref}}}{F^{\mathrm{ref}}}
   - \frac{H^{\mathrm{ref}}}{P + F^{\mathrm{ref}}} \\
&= \frac{H^{\mathrm{ref}} P}{F^{\mathrm{ref}}(P + F^{\mathrm{ref}})} > 0.
\end{aligned}
\label{eq:app_order}
\end{equation}
Equation~\eqref{eq:app_order} reveals the inherent trade-off between CSI and FAR: relative to a unit miss penalty, reducing FAR requires a strictly heavier false-alarm weight ($\kappa_2$) than improving CSI ($\kappa_1$).
Consequently, no single weight $\kappa$ can optimize both metrics to their standalone limits simultaneously.
As verified empirically in Figure~10a (Section~5.3), optimizing with $\kappa = \kappa_1$ yields the highest CSI by tolerating wider rain coverage and higher FAR, whereas optimizing with $\kappa = \kappa_2$ achieves the lowest FAR via conservative predictions that sacrifice CSI.

\paragraph{Convex combination and the trade-off interval \texorpdfstring{$[\kappa_1, \kappa_2]$}{[kappa1, kappa2]}.}
To balance the two objectives within a single linear loss, consider any normalized non-negative combination ($a, b \ge 0$, $a + b > 0$) of the endpoint costs $J_{\kappa_1}(\theta)$ and $J_{\kappa_2}(\theta)$:
\begin{equation}
\begin{aligned}
\frac{a J_{\kappa_1}(\theta) + b J_{\kappa_2}(\theta)}{a + b}
&= M(\theta) + \big[(1 - \alpha)\kappa_1 + \alpha \kappa_2\big] F(\theta) \\
&= J_\kappa(\theta),
\end{aligned}
\label{eq:app_nonnegative}
\end{equation}
where $\alpha = b / (a + b) \in [0, 1]$ and $\kappa = (1 - \alpha)\kappa_1 + \alpha \kappa_2 \in [\kappa_1, \kappa_2]$.
Conversely, every $\kappa \in [\kappa_1, \kappa_2]$ corresponds to a unique trade-off coefficient $\alpha = (\kappa - \kappa_1)/(\kappa_2 - \kappa_1)$.
Defining the reference-centered cost change
\begin{equation}
\begin{aligned}
w(\kappa)
&= J_\kappa(\theta) - J_\kappa^{\mathrm{ref}}
 = \Delta M + \kappa \Delta F \\
&= (1 - \alpha)w(\kappa_1) + \alpha w(\kappa_2),
\end{aligned}
\label{eq:app_w}
\end{equation}
Equations~\eqref{eq:app_csi_criterion}, \eqref{eq:app_far_criterion}, and~\eqref{eq:app_w} clarify why restricting $\kappa$ to $[\kappa_1, \kappa_2]$ defines the meaningful trade-off range:
\begin{itemize}
\item \emph{At least one metric improves:} If neither score improves over $f_\theta^{\mathrm{ref}}$, then $w(\kappa_1) \ge 0$ and $w(\kappa_2) \ge 0$, implying $w(\kappa) \ge 0$ for all $\alpha \in [0, 1]$. Hence, for any $\kappa \in [\kappa_1, \kappa_2]$, achieving a strict cost reduction $w(\kappa) < 0$ guarantees $\mathrm{CSI} > \mathrm{CSI}^{\mathrm{ref}}$ or $\mathrm{FAR} < \mathrm{FAR}^{\mathrm{ref}}$.
\item \emph{All joint improvements are preserved:} Whenever a candidate forecast improves both metrics ($\mathrm{CSI} > \mathrm{CSI}^{\mathrm{ref}}$ and $\mathrm{FAR} < \mathrm{FAR}^{\mathrm{ref}}$), both $w(\kappa_1) < 0$ and $w(\kappa_2) < 0$ hold, so $w(\kappa) < 0$ holds for every $\kappa \in [\kappa_1, \kappa_2]$, while varying $\alpha$ adjusts the relative preference between CSI gains and FAR reduction.
\end{itemize}

\paragraph{Geometric interpretation in the \texorpdfstring{$(F,H)$}{(F,H)} plane.}
Substituting $\Delta M = -(H - H^{\mathrm{ref}})$ and $\Delta F = F - F^{\mathrm{ref}}$ into Equation~\eqref{eq:app_w} expresses $w(\kappa) < 0$ as the open half-plane above the line of slope $\kappa$ passing through $(F^{\mathrm{ref}}, H^{\mathrm{ref}})$ (Figure~\ref{fig:hf_plane}):
\begin{equation}
w(\kappa) < 0
\iff H - H^{\mathrm{ref}} > \kappa(F - F^{\mathrm{ref}}).
\label{eq:app_hf_halfplane}
\end{equation}
Evaluating the boundary line $H = H^{\mathrm{ref}} + \kappa(F - F^{\mathrm{ref}})$ at the two endpoints $\kappa_1$ and $\kappa_2$ gives
\begin{equation}
\begin{aligned}
\kappa = \kappa_1:\\
H &= H^{\mathrm{ref}} + \frac{H^{\mathrm{ref}}}{P + F^{\mathrm{ref}}}(F - F^{\mathrm{ref}}) \\
   &= \frac{H^{\mathrm{ref}}}{P + F^{\mathrm{ref}}}(P + F), \\
\kappa = \kappa_2:\\
H &= H^{\mathrm{ref}} + \frac{H^{\mathrm{ref}}}{F^{\mathrm{ref}}}(F - F^{\mathrm{ref}}) \\
   &= \frac{H^{\mathrm{ref}}}{F^{\mathrm{ref}}} F.
\end{aligned}
\label{eq:app_boundary_endpoints}
\end{equation}
By Equations~\eqref{eq:app_csi} and~\eqref{eq:app_far}, the $\kappa_1$ boundary ($w(\kappa_1) = 0$) intersects $H = 0$ at $F = -P$ and coincides with the exact iso-CSI contour $\mathrm{CSI} = \mathrm{CSI}^{\mathrm{ref}}$, whereas the $\kappa_2$ boundary ($w(\kappa_2) = 0$) passes through the origin $(0, 0)$ and coincides with the exact iso-FAR ray $\mathrm{FAR} = \mathrm{FAR}^{\mathrm{ref}}$.
Together, these two lines partition the feasible strip $F \ge 0, H \in [0, P]$ around $(F^{\mathrm{ref}}, H^{\mathrm{ref}})$ into four sectors (Figure~\ref{fig:hf_plane}):
\begin{enumerate}
\item \textbf{Upper-left joint-improvement cone} ($w(\kappa_1) < 0$ and $w(\kappa_2) < 0$): both $\mathrm{CSI}\uparrow$ and $\mathrm{FAR}\downarrow$;
\item \textbf{Upper-right over-forecasting wedge} ($F > F^{\mathrm{ref}}$, $\kappa_1(F - F^{\mathrm{ref}}) < H - H^{\mathrm{ref}} < \kappa_2(F - F^{\mathrm{ref}})$): expanding rain coverage yields $\mathrm{CSI}\uparrow$ at the cost of $\mathrm{FAR}\uparrow$;
\item \textbf{Lower-left under-forecasting wedge} ($F < F^{\mathrm{ref}}$, $\kappa_2(F - F^{\mathrm{ref}}) < H - H^{\mathrm{ref}} < \kappa_1(F - F^{\mathrm{ref}})$): shrinking rain coverage yields $\mathrm{FAR}\downarrow$ at the cost of $\mathrm{CSI}\downarrow$;
\item \textbf{Lower-right joint-degradation cone} ($w(\kappa_1) > 0$ and $w(\kappa_2) > 0$): both $\mathrm{CSI}\downarrow$ and $\mathrm{FAR}\uparrow$.
\end{enumerate}
Increasing $\kappa$ from $\kappa_1$ to an interior trade-off weight $\kappa' \in (\kappa_1, \kappa_2)$ rotates the active half-plane boundary counter-clockwise from the iso-CSI contour toward the iso-FAR ray—retaining the entire joint-improvement cone while pruning excessive false-alarm growth in the over-forecasting wedge to achieve a balanced CSI--FAR compromise.

\subsection{Peak-Normalized Weight Calibration}
\label{sec:app_calibration}

Applying this trade-off across all groups $g=(\ell,k)$ must account for severe class imbalance across precipitation intensities.
Within a given threshold $\tau_k$, $\kappa_{1,\ell,k} = H_{\ell,k}^{\mathrm{ref}} / (P_k + F_{\ell,k}^{\mathrm{ref}})$ captures how the hit--false-alarm balance decays with lead time $\ell$.
At high thresholds, however, convective cores are sparse ($H_{\ell,k}^{\mathrm{ref}} \ll P_k + F_{\ell,k}^{\mathrm{ref}}$), causing $\kappa_{1,\ell,k}$ to drop toward zero and leaving extreme-precipitation false alarms under-penalized.

Peak-normalized calibration in Equation~\eqref{eq:kappa_fixed} addresses this as a practical cross-threshold compromise rather than a globally optimal weight, fulfilling three design goals:
\begin{enumerate}
\item \textbf{Rare-event false-alarm uplift:} Dividing $\kappa_{1,\ell,k}$ by its within-threshold peak $\max_{\ell'} \kappa_{1,\ell',k}$ and scaling by $\rho > 0$ raises the peak false-alarm penalty at rare, intense thresholds to a consistent target scale ($\approx \rho$, comparable to unit miss cost), while preserving exact lead-time proportions $\kappa'_{\ell,k} / \kappa'_{\ell',k} = \kappa_{1,\ell,k} / \kappa_{1,\ell',k}$ whenever the upper clip is inactive.
\item \textbf{Guaranteed trade-off bounds:} Clipping to $[\kappa_{1,\ell,k}, \kappa_{2,\ell,k}]$ ensures that every group stays strictly within the valid trade-off interval of Appendix~\ref{sec:app_range}, preventing both uncurbed false-alarm growth and overly conservative rain suppression.
\item \textbf{Single-knob CSI--FAR control:} Adjusting $\rho$ smoothly shifts the compromise between higher CSI (smaller $\rho$) and lower FAR (larger $\rho$), allowing single-stage tuning with fixed $\kappa'_{\ell,k}$ to match a two-stage $\kappa_1 \to \kappa'$ curriculum without scheduling overhead (Section~5.3, Figure~10a).
\end{enumerate}

\subsection{Ensemble Collapse under Skill-Only Optimization}
\label{sec:app_collapse}

We now show why optimizing the soft skill objective $\mathcal{L}_{\mathrm{skill}}(\theta)$ in Equation~\eqref{eq:skill} without regularization ($\eta = 0$) causes forecast distributions to lose ensemble spread and fine-scale structure.
Physical radar intensities lie in a bounded interval $\mathcal{X} = [x_{\min}, x_{\max}]$ (e.g., $[0, 255]$\,VIL), so candidate forecasts take values in the compact domain $\mathcal{X}^d$ ($d = L H W$).
Fix a historical condition $\mathbf{c}$, write $\ell_{\mathrm{skill}}(\hat{\mathbf{x}}, \mathbf{x})$ for the single-sample evaluation of Equation~\eqref{eq:skill}, and define the continuous conditional skill risk and its minimum on $\mathcal{X}^d$:
\begin{equation}
\phi_{\mathbf{c}}(\hat{\mathbf{x}})
= \mathbb{E}_{\mathbf{x}\mid\mathbf{c}}\big[\ell_{\mathrm{skill}}(\hat{\mathbf{x}}, \mathbf{x})\big],
\qquad
m_{\mathbf{c}}
= \min_{\hat{\mathbf{x}}\in\mathcal{X}^d} \phi_{\mathbf{c}}(\hat{\mathbf{x}}).
\label{eq:app_conditional_risk}
\end{equation}
Because initial noise $\boldsymbol{\epsilon} \sim \pi_{\mathbf{c}}$ is conditionally independent of $\mathbf{x}$ given $\mathbf{c}$, denoting by $p_\theta(\cdot \mid \mathbf{c})$ the forecast distribution of $\hat{\mathbf{x}}_\theta(\mathbf{c}, \boldsymbol{\epsilon})$, the law of iterated expectations gives
\begin{equation}
\begin{aligned}
&\mathbb{E}_{\boldsymbol{\epsilon},\,\mathbf{x}\mid\mathbf{c}}
 \big[\ell_{\mathrm{skill}}(\hat{\mathbf{x}}_\theta(\mathbf{c}, \boldsymbol{\epsilon}), \mathbf{x})\big] \\
&\quad = \mathbb{E}_{\hat{\mathbf{x}}\sim p_\theta(\cdot\mid\mathbf{c})}
   \big[\phi_{\mathbf{c}}(\hat{\mathbf{x}})\big]
 \ge m_{\mathbf{c}}.
\end{aligned}
\label{eq:app_collapse}
\end{equation}
Unlike strictly proper distributional scores such as CRPS—which include a pairwise ensemble-spread repulsion term—$\phi_{\mathbf{c}}(\hat{\mathbf{x}})$ evaluates forecasts marginally: a deterministic Dirac mass $p = \delta_{\hat{\mathbf{x}}^\star(\mathbf{c})}$ at any $\hat{\mathbf{x}}^\star(\mathbf{c}) \in \operatorname{arg\,min}_{\hat{\mathbf{x}}\in\mathcal{X}^d} \phi_{\mathbf{c}}(\hat{\mathbf{x}})$ already attains the global lower bound $m_{\mathbf{c}}$, so stochasticity offers no reduction in skill risk.

To see what $\hat{\mathbf{x}}^\star(\mathbf{c})$ looks like spatially, let $\ell_i$ denote the lead time of grid entry $i$ and $q_{i,k}(\mathbf{c}) = \Pr(x_i \ge \tau_k \mid \mathbf{c})$ be the true conditional exceedance probability.
By linearity of expectation, $\phi_{\mathbf{c}}(\hat{\mathbf{x}})$ decouples across pixels:
\begin{equation}
\begin{aligned}
\phi_{\mathbf{c}}(\hat{\mathbf{x}})
&= \frac{1}{|G|}\sum_{i=1}^d \psi_i(\hat{x}_i), \\
\psi_i(z)
&= \sum_{k:(\ell_i,k)\in G}
   \big[a_{i,k}(1 - s_k(z))^2 + b_{i,k}s_k(z)^2\big],
\end{aligned}
\label{eq:app_phi}
\end{equation}
where $s_k(z) = \sigma((z - \tau_k)/T)$, $a_{i,k} = q_{i,k}(\mathbf{c}) / D_{\ell_i,k}$, and $b_{i,k} = \kappa'_{\ell_i,k}(1 - q_{i,k}(\mathbf{c})) / D_{\ell_i,k}$.
In the illustrative case of a single active threshold $\tau$ with weight $\kappa' > 0$ and $q_i(\mathbf{c}) \in (0, 1)$, setting the derivative of $q_i(\mathbf{c})(1 - s)^2 + \kappa'(1 - q_i(\mathbf{c}))s^2$ with respect to $s \in (0, 1)$ to zero yields the unique closed-form minimizer on $\mathcal{X}$:
\begin{equation}
\begin{aligned}
s^\star_i(\mathbf{c})
&= \frac{q_i(\mathbf{c})}{q_i(\mathbf{c}) + \kappa'\big(1 - q_i(\mathbf{c})\big)}, \\
\hat{x}_i^\star(\mathbf{c})
&= \operatorname{clip}_{[x_{\min},\,x_{\max}]}\!\left(
   \tau + T \ln \frac{q_i(\mathbf{c})}{\kappa'\big(1 - q_i(\mathbf{c})\big)}
   \right),
\end{aligned}
\label{eq:app_closed_form}
\end{equation}
with boundary probabilities $q_i(\mathbf{c}) = 0$ and $q_i(\mathbf{c}) = 1$ mapping directly to $x_{\min}$ and $x_{\max}$ (and multi-threshold sums $\psi_i(z)$ similarly attaining deterministic pixelwise minima on $\mathcal{X}$).
Equations~\eqref{eq:app_collapse}--\eqref{eq:app_closed_form} directly explain the $\eta = 0$ behavior in Section~5.4 (Figures~12 and~13): (i) because $\hat{\mathbf{x}}^\star(\mathbf{c})$ depends solely on $\mathbf{c}$, unanchored tuning drives the generator to ignore $\boldsymbol{\epsilon}$, collapsing ensemble spread; and (ii) at longer lead times where convective cell positions are uncertain, $q_i(\mathbf{c})$ is spatially diffuse, so mapping $q_i(\mathbf{c})$ deterministically into pixel intensities produces oversmoothed rainbands with severe loss of sub-96\,km spectral power.

\subsection{Path-Space KL Interpretation and Scope Matching of Velocity Anchoring}
\label{sec:app_anchor}

To prevent ensemble collapse while preserving fine-scale generative textures, SAPT anchors the fine-tuned velocity field toward $f_\theta^{\mathrm{ref}}$~\cite{fan2023dpok,domingoenrich2025adjoint}.
Fix condition $\mathbf{c}$, identify any $L \times H \times W$ sequence with a vector in $\mathbb{R}^d$ ($d = L H W$), write $\|\mathbf{u}\|^2 = d^{-1}\|\mathbf{u}\|_2^2$ for the entrywise mean squared norm used in the main text, and let $h_n = t_{n+1} - t_n > 0$ denote the Euler step sizes on $0 = t_0 < \dots < t_N = 1$.
Under the $x$-prediction parameterization of Equation~\eqref{eq:xpred}, the linear state term $-\mathbf{z}/(1 - t_n)$ cancels at any $t_n < 1$, giving the velocity residual
\begin{equation}
\begin{aligned}
\Delta v_n(\mathbf{z})
&= v_\theta(\mathbf{z}, t_n, \mathbf{c})
   - v_\theta^{\mathrm{ref}}(\mathbf{z}, t_n, \mathbf{c}) \\
&= \frac{f_\theta(\mathbf{z}, t_n, \mathbf{c})
         - f_\theta^{\mathrm{ref}}(\mathbf{z}, t_n, \mathbf{c})}
        {1 - t_n}.
\end{aligned}
\label{eq:app_dv}
\end{equation}
Under classifier-free guidance at scale $w$, both $f_\theta$ and $f_\theta^{\mathrm{ref}}$ denote the guided clean-field prediction $f(\mathbf{z}, t_n, \varnothing) + w(f(\mathbf{z}, t_n, \mathbf{c}) - f(\mathbf{z}, t_n, \varnothing))$; by linearity, the guided velocities satisfy Equation~\eqref{eq:app_dv} identically.

\paragraph{Path-space KL upper bound along stochasticized rollouts.}
Because deterministic Euler updates define singular Dirac transition kernels, consider an auxiliary stochastic Euler chain with perturbation scale $\sigma > 0$:
\begin{equation}
\begin{aligned}
\widetilde{\mathbf{x}}_{n+1}
&= \widetilde{\mathbf{x}}_n
   + h_n\big[v(\widetilde{\mathbf{x}}_n, t_n, \mathbf{c}) + \sigma\boldsymbol{\xi}_n\big], \\
\boldsymbol{\xi}_n &\stackrel{\mathrm{i.i.d.}}{\sim} \mathcal{N}(\mathbf{0}, \mathbf{I}_d),
\qquad \widetilde{\mathbf{x}}_0 \sim \pi_{\mathbf{c}}.
\end{aligned}
\label{eq:app_kernel}
\end{equation}
For $v = v_\theta$ and $v = v_\theta^{\mathrm{ref}}$, Equation~\eqref{eq:app_kernel} induces Gaussian transition kernels $\widetilde{K}_{\theta,n}^\sigma(\cdot \mid \mathbf{z})$ and $\widetilde{K}_{\mathrm{ref},n}^\sigma(\cdot \mid \mathbf{z})$ with means $\mathbf{z} + h_n v(\mathbf{z}, t_n, \mathbf{c})$ and shared covariance $\sigma^2 h_n^2 \mathbf{I}_d$.
Let $\widetilde{\mathbb{P}}_\theta^\sigma, \widetilde{\mathbb{P}}_{\mathrm{ref}}^\sigma$ denote the joint path distributions and $\widetilde{p}_\theta^\sigma, \widetilde{p}_{\mathrm{ref}}^\sigma$ their terminal marginals at $t_N = 1$.
By the data-processing inequality, KL chain rule~\cite{cover2006elements}, and equal-covariance Gaussian KL formula,
\begin{equation}
\begin{aligned}
&\mathrm{KL}(\widetilde{p}_\theta^\sigma \| \widetilde{p}_{\mathrm{ref}}^\sigma)
 \le \mathrm{KL}(\widetilde{\mathbb{P}}_\theta^\sigma \| \widetilde{\mathbb{P}}_{\mathrm{ref}}^\sigma) \\
&\quad = \sum_{n=0}^{N-1} \mathbb{E}_{\widetilde{\mathbb{P}}_\theta^\sigma}\!\left[
  \mathrm{KL}\big(\widetilde{K}_{\theta,n}^\sigma(\cdot \mid \widetilde{\mathbf{x}}_n) \,\big\|\,
  \widetilde{K}_{\mathrm{ref},n}^\sigma(\cdot \mid \widetilde{\mathbf{x}}_n)\big)\right] \\
&\quad = \frac{d}{2\sigma^2}\sum_{n=0}^{N-1}
  \mathbb{E}_{\widetilde{\mathbb{P}}_\theta^\sigma}
  \big\|\Delta v_n(\widetilde{\mathbf{x}}_n)\big\|^2,
\end{aligned}
\label{eq:app_chain}
\end{equation}
where $h_n^2$ cancels between the squared mean shift $\|h_n\Delta v_n\|_2^2$ and covariance $\sigma^2 h_n^2 \mathbf{I}_d$, yielding uniform weights in velocity space and $(1 - t_n)^{-2}$ weights in $x$-prediction space (evaluated along deterministic rollouts $\mathbf{x}_n^\theta$ as $\sigma \to 0^+$).

\paragraph{Final-step velocity anchor and scope matching.}
Under SAPT's truncated backpropagation~\cite{clark2024draft}, the first $N - 1$ Euler steps are rolled out without gradient tracking to obtain $\bar{\mathbf{x}}_* = \operatorname{sg}[\mathbf{x}_{t_*}]$ at $t_* = t_{N-1}$, differentiating solely through the final step $\hat{\mathbf{x}} = f_\theta(\bar{\mathbf{x}}_*, t_*, \mathbf{c})$ in Equation~\eqref{eq:last_step}.
Isolating the final transition ($n = N - 1$) conditioned on $\bar{\mathbf{x}}_*$ in Equation~\eqref{eq:app_chain} directly recovers Equation~\eqref{eq:anchor}:
\begin{equation}
\begin{aligned}
&\frac{2\sigma^2}{d}\,\mathrm{KL}\!\left(
  \widetilde{K}_{\theta,N-1}^\sigma(\cdot \mid \bar{\mathbf{x}}_*) \,\big\|\,
  \widetilde{K}_{\mathrm{ref},N-1}^\sigma(\cdot \mid \bar{\mathbf{x}}_*)
 \right) \\
&\quad = \big\|\Delta v_{N-1}(\bar{\mathbf{x}}_*)\big\|^2 \\
&\quad = \frac{\big\|f_\theta(\bar{\mathbf{x}}_*, t_*, \mathbf{c})
         - \operatorname{sg}[f_\theta^{\mathrm{ref}}(\bar{\mathbf{x}}_*, t_*, \mathbf{c})]\big\|^2}
        {(1 - t_*)^2} \\
&\quad = \mathcal{L}_{\mathrm{anchor}}^{\mathrm{last}}(\theta).
\end{aligned}
\label{eq:app_anchor_last_kl}
\end{equation}
Equations~\eqref{eq:app_chain} and~\eqref{eq:app_anchor_last_kl} clarify why matching gradient horizon and anchor coverage is essential in Section~5.4 (Figure~11):
\begin{itemize}
\item \textbf{Matched final-step scope (\texttt{Last/last}, default):} Both $\nabla_\theta \mathcal{L}_{\mathrm{skill}}$ and $\nabla_\theta \mathcal{L}_{\mathrm{anchor}}^{\mathrm{last}}$ act on the same state--time pair $(\bar{\mathbf{x}}_*, t_*)$, directly regularizing the terminal map $f_\theta(\bar{\mathbf{x}}_*, t_*, \mathbf{c})$ while avoiding backward passes through steps $0, \dots, N - 2$.
\item \textbf{Mismatched scopes (\texttt{Full/last} vs.\ \texttt{Last/all}):} Backpropagating $\mathcal{L}_{\mathrm{skill}}$ through all $N$ steps while anchoring only $n = N - 1$ (\texttt{Full/last}) leaves intermediate transitions $n < N - 1$ unpenalized in Equation~\eqref{eq:app_chain}, allowing $\mathbf{x}_{t_*}$ to drift off the pretrained manifold and degrading LPIPS. Conversely, truncating skill gradients to $n = N - 1$ while penalizing detached velocity residuals at all steps (\texttt{Last/all}) constrains shared parameters $\theta$ at $N - 1$ unrewarded timesteps, forfeiting compute savings and limiting CSI gains.
\end{itemize}

\paragraph{Output-space coupling and cross-benchmark scaling of \texorpdfstring{$\eta$}{eta}.}
By Equation~\eqref{eq:last_step}, for any shared noise draw $\boldsymbol{\epsilon} \sim \pi_{\mathbf{c}}$ producing $\bar{\mathbf{x}}_*$, the terminal forecast deviation from the reference final-step prediction $\hat{\mathbf{x}}_*^{\mathrm{ref}} = f_\theta^{\mathrm{ref}}(\bar{\mathbf{x}}_*, t_*, \mathbf{c})$ satisfies the exact identity
\begin{equation}
\big\|\hat{\mathbf{x}}_\theta(\mathbf{c}, \boldsymbol{\epsilon})
 - \hat{\mathbf{x}}_*^{\mathrm{ref}}\big\|^2
= (1 - t_*)^2\,\mathcal{L}_{\mathrm{anchor}}^{\mathrm{last}}(\theta),
\label{eq:app_last_disp}
\end{equation}
penalizing any collapse of noise-induced sample variations across $\bar{\mathbf{x}}_*$.
Because resolution-dependent timestep shifts adjust the sampling grid $\{t_n\}_{n=0}^{N-1}$ across benchmarks, scaling $\eta$ inversely with $(1 - t_*)^{-2}$ (or $\sum_{n=0}^{N-1}(1 - t_n)^{-2}$ under all-step anchoring) maintains a consistent effective output-space regularization strength across datasets (see Appendix~\ref{sec:app_alignpre_implementation}).
\section{Pixel-Space Architecture \& Implementation Details}
\label{sec:app_alignpre_implementation}

\subsection{Hierarchical Spatiotemporal Backbone (AlignPre-Base)}
\label{sec:app_alignpre_backbone}

Table~\ref{tab:app_alignpre_sizes} summarizes the spatiotemporal input geometry across the three radar benchmarks and the architectural specifications of the four model scales (\textrm{S}, \textrm{B}, \textrm{L}, and \textrm{XL}).

\begin{table*}[tp]
\centering
\small
\setlength{\tabcolsep}{5pt}
\renewcommand{\arraystretch}{1.06}
\caption{\textbf{AlignPre-Base architecture and benchmark geometry.} Main benchmark comparisons adopt the \textrm{B} model across all datasets. Parameter counts are reported for the SEVIR-2h network, including all embedding layers and the linear pixel readout.}
\label{tab:app_alignpre_sizes}
\begin{tabular*}{\textwidth}{@{\extracolsep{\fill}}lccc@{}}
\toprule
\textbf{Benchmark geometry (\textrm{B} model)} & SEVIR-1h & SEVIR-2h & Guangdong-3h \\
\midrule
Single-channel radar grid ($H\times W$) & $384\times384$ & $128\times128$ & $128\times128$ \\
Physical grid spacing $\Delta_x$ / frame interval $\Delta_t$ & $1.0\,\mathrm{km}\;/\;5\,\mathrm{min}$ & $3.0\,\mathrm{km}\;/\;5\,\mathrm{min}$ & $2.8125\,\mathrm{km}\;/\;6\,\mathrm{min}$ \\
Sequence horizon ($L_h \to L$ frames) & $13 \to 12$ ($60\,\mathrm{min}$) & $5 \to 20$ ($100\,\mathrm{min}$) & $10 \to 30$ ($180\,\mathrm{min}$) \\
Patch side in pixels (fine $P$ / bottleneck $2P$) & $8\;/\;16$ & $4\;/\;8$ & $4\;/\;8$ \\
Token grid per frame (fine / bottleneck) & $48^2\;/\;24^2$ & $32^2\;/\;16^2$ & $32^2\;/\;16^2$ \\
\bottomrule
\end{tabular*}

\vspace{5pt}
\begin{tabular*}{\textwidth}{@{\extracolsep{\fill}}lcccc@{}}
\toprule
\textbf{Model scale configuration (SEVIR-2h)} & \textrm{S} & \textrm{B} & \textrm{L} & \textrm{XL} \\
\midrule
Parameters (M) & $51.2$ & $115.0$ & $363.7$ & $692.6$ \\
Blocks (fine down / coarse bottleneck / fine up) & $2\;/\;8\;/\;2$ & $2\;/\;8\;/\;2$ & $2\;/\;16\;/\;2$ & $2\;/\;20\;/\;2$ \\
Feature width (fine / bottleneck) & $256\;/\;512$ & $384\;/\;768$ & $512\;/\;1024$ & $640\;/\;1280$ \\
Attention heads (fine / bottleneck, head dim $64$) & $4\;/\;8$ & $6\;/\;12$ & $8\;/\;16$ & $10\;/\;20$ \\
\midrule
Fine-scale attention & \multicolumn{4}{c}{$5\times5$ intra-frame neighborhood attention followed by per-location temporal attention} \\
Coarse bottleneck attention & \multicolumn{4}{c}{Alternating full intra-frame global attention and per-location temporal attention} \\
Feed-forward network & \multicolumn{4}{c}{SwiGLU with hidden dimension $3\times$ the block feature width} \\
\bottomrule
\end{tabular*}
\end{table*}
\paragraph{Token hierarchy and skip fusion.}
AlignPre-Base instantiates $f_\theta(x_t,t,c)$ as a two-level, U-shaped hierarchical spatiotemporal transformer~\cite{vallaeys2026ujit} operating directly on radar frames (Figure~\ref{fig:pixel_nowcaster}). A shared linear patchify layer embeds non-overlapping $P\times P$ pixel patches from each frame into fine-scale tokens, adding learnable binary role embeddings to distinguish clean historical observations $c$ from noisy future inputs $x_t$. Both streams traverse the fine-scale downsampling path and coarse bottleneck. Each fine-scale block sequentially applies $5\times5$ intra-frame neighborhood attention, per-location temporal attention~\cite{hassani2023neighborhood,bertasius2021timesformer}, and a SwiGLU feed-forward network (FFN) to capture localized convective evolution. Spatial token merging then groups each $2\times2$ window of fine-scale tokens and linearly projects them into a single coarse-scale bottleneck token. Beginning with intra-frame global attention, consecutive bottleneck blocks alternate single global spatial attention and per-location temporal attention layers—each paired with its own SwiGLU FFN—to model large-scale storm organization. Historical tokens terminate after the bottleneck; along the future-only upsampling path, a linear projection splits each future bottleneck token back into $2\times2$ fine-scale tokens $h_{\mathrm{up}}$, which are fused with the corresponding future skip features $h_{\mathrm{skip}}$ from the downsampling path via affine interpolation:
\begin{equation}
h_{\mathrm{fused}} = h_{\mathrm{skip}} + \alpha\big(h_{\mathrm{up}} - h_{\mathrm{skip}}\big),
\label{eq:app_alignpre_skip}
\end{equation}
where $\alpha$ is a learnable scalar initialized to $0.5$. After the fine-scale up-path blocks, an adaptive RMSNorm and a zero-initialized linear readout project each token back to a $P\times P$ patch and unpatchify them into the predicted clean future sequence.

\paragraph{Time-lead conditioning and normalization.}
Two scalar embedders encode the continuous flow-matching time as $1000t$ and the physical lead time $l$ in minutes, respectively. Each embedder maps its scalar input through a $256$-dimensional sinusoidal encoding followed by a two-layer SiLU MLP matching the bottleneck width, and the resulting vectors are summed per frame. Within every transformer block, zero-initialized linear projections map this frame-level conditioning vector to adaptive shift, scale, and residual gating parameters for each RMS-normalized attention and SwiGLU sublayer. Prior to rotary position encoding~\cite{su2024roformer}, queries and keys undergo separate per-head RMSNorm~\cite{esser2024scaling} to stabilize attention logits across scales.

\paragraph{Physical-coordinate 3D RoPE.}
Let $a\in\{0,1\}$ index the fine ($a=0$) and coarse bottleneck ($a=1$) resolution levels with spatial token grid $G_a\times G_a$, and let $\Delta_x$ denote the physical pixel spacing in kilometers (Table~\ref{tab:app_alignpre_sizes}). Taking the latest historical observation at $l=0\,\mathrm{min}$ (so earlier observations have $l<0$ and future forecast frames have $l>0$), a token centered at spatial grid index $(i,j)\in\{0,\ldots,G_a-1\}^2$ and time offset $l$ is assigned physical coordinates
\begin{equation}
p_a(l,i,j) = \big(l,\,(i+\tfrac{1}{2})2^a P\Delta_x,\,(j+\tfrac{1}{2})2^a P\Delta_x\big).
\label{eq:app_alignpre_rope_position}
\end{equation}
Each $64$-dimensional attention head partitions its channels into $(16,24,24)$ dimensions—i.e., $(8,12,12)$ feature pairs—for the temporal, row, and column axes, respectively. Along each axis, wavelengths $\lambda_m$ are spaced geometrically across the closed range and each feature pair is rotated by $2\pi p/\lambda_m$, where $p$ is the corresponding coordinate in Equation~\eqref{eq:app_alignpre_rope_position}. 

\paragraph{One-pass history K/V caching and classifier-free guidance.}
Along the downsampling path and bottleneck, historical frames $c$ are encoded with their flow-matching time fixed at $t=1$. Asymmetric masking in temporal attention ensures that historical queries attend solely to historical keys and values, whereas future queries attend to concatenated historical and future keys and values. Because the historical branch is independent of $x_t$ and $t$, its temporal keys and values are computed once per condition $c$ and cached across all $N$ Euler steps as well as across ensemble members. For the \textrm{B} model, this cache stores two fine-scale and four bottleneck key--value (K/V) pairs; at level $a$ with $h_a$ attention heads, each cached key or value tensor has shape $G_a^2\times h_a\times L_h\times64$. The upsampling path processes future tokens exclusively via future-only temporal attention.
During pretraining, we drop all historical keys and values in temporal attention independently per sequence with probability $p_{\mathrm{drop}}=0.25$ to support classifier-free guidance (CFG)~\cite{ho2022cfg}. Because all spatial, merging, and conditioning operations act strictly within each frame, omitting the historical cache and evaluating future tokens alone at inference implements the exact unconditional pass $f_\theta(x_t,t,\emptyset)$. Along a single rollout initialized from $\epsilon\sim\mathcal{N}(0,I)$, the guided clean prediction at shared state $x_t$ and scale $w$ is
\begin{equation}
f_\theta^{(w)}(x_t,t,c) = \hat{x}_{\emptyset} + w\big(\hat{x}_c - \hat{x}_{\emptyset}\big),
\label{eq:app_alignpre_cfg}
\end{equation}
where $\hat{x}_c = f_\theta(x_t,t,c)$ and $\hat{x}_{\emptyset} = f_\theta(x_t,t,\emptyset)$ are evaluated at the identical state $x_t$. During the differentiable final step of SAPT, both branches are stacked into a single batch with historical keys masked on the unconditional slice, whereas unguided evaluation ($w=1.0$) executes only the conditional pass.

\subsection{Pretraining, SAPT Algorithm \& Hyperparameter Summary}
\label{sec:app_alignpre_training}

Table~\ref{tab:app_alignpre_hyperparameters} lists the pretraining, post-training, and sampling hyperparameters across all three benchmarks, and Algorithm~\ref{alg:app_alignpre_sapt} details the complete optimization procedure of skill-aligned post-training (SAPT).

\begin{table*}[tp]
\centering
\small
\setlength{\tabcolsep}{7pt}
\renewcommand{\arraystretch}{1.06}
\caption{\textbf{Hyperparameters for AlignPre pretraining, SAPT post-training, and inference across benchmarks.}}
\label{tab:app_alignpre_hyperparameters}
\begin{tabular*}{\textwidth}{@{\extracolsep{\fill}}lccc@{}}
\toprule
\textbf{Hyperparameter} & SEVIR-1h & SEVIR-2h & Guangdong-3h \\
\midrule
\multicolumn{4}{@{}l}{\textit{Flow-matching pretraining (AlignPre-Base)}} \\
Backbone scale, patch size (fine / bottleneck) & \textrm{B}, $8\;/\;16$ & \textrm{B}, $4\;/\;8$ & \textrm{B}, $4\;/\;8$ \\
Noise-shifted schedule mode \& shift factor $s$ & Decaying, $s=7.5$ & Constant, $s=2.5$ & Constant, $s\approx3.16$ \\
Velocity denominator floor $\epsilon_t$ & \multicolumn{3}{c}{$0.05$} \\
Initial noise prior $\pi_c$ \& CFG dropout $p_{\mathrm{drop}}$ & \multicolumn{3}{c}{Standard Gaussian $\mathcal{N}(0,I)$,\quad $p_{\mathrm{drop}}=0.25$} \\
Optimizer & \multicolumn{3}{c}{AdamW, $(\beta_1,\beta_2)=(0.9,0.95)$, $\epsilon=10^{-8}$, weight decay $0$} \\
Learning rate \& schedule & \multicolumn{3}{c}{$1\times10^{-4}$, linear warmup over $5{,}000$ updates, then constant} \\
Training budget \& global batch size (seq. / update) & $100$ epochs, $56$ & $100$ epochs, $80$ & $100$ epochs, $72$ \\
EMA decay, gradient clip norm, precision & \multicolumn{3}{c}{$0.9999$,\quad $1.0$,\quad $\mathrm{bf16}$} \\
\midrule
\multicolumn{4}{@{}l}{\textit{Skill-aligned post-training (SAPT)}} \\
Trainable scope \& optimizer & \multicolumn{3}{c}{Full-parameter fine-tuning,\quad AdamW ($(\beta_1,\beta_2)=(0.9,0.95)$, weight decay $0$)} \\
Learning rate \& linear warmup (updates, then const.) & $1\times10^{-5}$ ($20$) & $1\times10^{-5}$ ($13$) & $1\times10^{-5}$ ($10$) \\
Training budget \& global batch size (cond. / update) & $1$ epoch, $8$ ($1$ rollout) & $1$ epoch, $12$ ($1$ rollout) & $1$ epoch, $16$ ($1$ rollout) \\
EMA decay, gradient clip norm, precision & \multicolumn{3}{c}{$0.999$,\quad $1.0$,\quad $\mathrm{bf16}$} \\
Offline calibration subset $|\mathcal{D}_{\mathrm{cal}}|$ & \multicolumn{3}{c}{$512$ evenly spaced training sequences (without augmentation)} \\
Sigmoid relaxation temperature $T$ & $10\,\mathrm{VIL}$ & $10\,\mathrm{VIL}$ & $3\,\mathrm{dBZ}$ \\
Target false-alarm peak scale $\rho$ & \multicolumn{3}{c}{$0.99$} \\
Last-step velocity anchor weight $\eta$ & $0.47$ & $1.1$ & $1.4$ \\
Differentiable \& anchored step scope & \multicolumn{3}{c}{Final Euler step ($t_* = t_{N-1}$)} \\
\midrule
\multicolumn{4}{@{}l}{\textit{Sampling \& inference}} \\
ODE solver \& integration steps $N$ & \multicolumn{3}{c}{Deterministic Euler solver,\quad $N=10$ steps} \\
Sampling timestep grid $\{t_n\}_{n=0}^N$ & \multicolumn{3}{c}{Shifted grid $t_n = S_s(n/N)$ matched to pretraining} \\
Classifier-free guidance scale $w$ & \multicolumn{3}{c}{AlignPre-Base (default): $w=1.0$;\quad SAPT rollout \& AlignPre: $w=1.6$} \\
Ensemble members $K$ (deterministic / CRPS) & \multicolumn{3}{c}{$K=1$ (member $0$) / $K=4$} \\
\bottomrule
\end{tabular*}
\end{table*}

\paragraph{Flow-matching pretraining and noise-shifted schedule.}
\label{sec:app_shifted_schedule}
Because higher-dimensional spatiotemporal sequences exhibit greater pixel redundancy and require stronger noise corruption to prevent trivial local interpolation~\cite{hoogeboom2023simple,esser2024scaling}, we scale the timestep shift factor $s$ with the total sequence volume $(L_h+L)HW$. Specifically, for each training sequence we draw $u\sim\mathcal{U}(0,1)$ and map it to flow time $t=S_s(u)\in[0,1)$ via
\begin{equation}
\begin{aligned}
s &= \sqrt{\frac{(L_h+L)HW}{256^2}},\\
t &= S_s(u) = \frac{r(u)u}{1-u+r(u)u},\\
r(u) &=
\begin{cases}
s^{-1}, & \text{constant shift},\\
s^{u-1}, & \text{decaying shift}.
\end{cases}
\end{aligned}
\label{eq:app_alignpre_time_shift}
\end{equation}
This formula yields $s=7.5$ on SEVIR-1h, $s=2.5$ on SEVIR-2h, and $s=\sqrt{10}\approx3.16$ on Guangdong-3h (Table~\ref{tab:app_alignpre_hyperparameters}). Given independent standard Gaussian noise $\epsilon\sim\mathcal{N}(0,I)$, we construct the linear interpolant $x_t = tx + (1-t)\epsilon$. Substituting the clean-data ($x$-) prediction parameterization into the conditional flow-matching loss with a numerical stability floor $\epsilon_t=0.05$ on $(1-t)$ gives the pretraining objective~\cite{li2026jit}:
\begin{equation}
\mathcal{L}_{\mathrm{pre}}(\theta) = \mathbb{E}_{c,x,\epsilon,t}\!\left[
\frac{\|f_\theta(x_t,t,c) - x\|^2}{\max(1-t,\,\epsilon_t)^2}
\right],
\label{eq:app_alignpre_pretrain}
\end{equation}
where $\|\cdot\|^2$ averages squared errors across all future frames and spatial grid points. Pretraining runs for $100$ epochs using AdamW~\cite{loshchilov2019adamw} with sequence-consistent random rotations (multiples of $90^\circ$) and horizontal/vertical flips, selecting EMA checkpoints by unguided ($w=1.0$) validation CSI-M.

\paragraph{Offline reference calibration and SAPT optimization.}
Prior to post-training, we freeze the pretrained EMA checkpoint as the reference generator $f_\theta^{\mathrm{ref}}$ and run inference once on a fixed calibration subset $\mathcal{D}_{\mathrm{cal}}\subset\mathcal{D}$ evenly spaced training sequences without augmentation, using the target post-training sampler. For each lead-time and threshold pair $g=(l,k)$, we accumulate binary contingency counts over $\mathcal{D}_{\mathrm{cal}}$ and divide by $|\mathcal{D}_{\mathrm{cal}}|$ to obtain per-sample expected reference hits $H_g^{\mathrm{ref}}$, false alarms $F_g^{\mathrm{ref}}$, and ground-truth positives $P_g$. Retaining active groups $G=\{g : D_g > 10^{-6}\}$ with reference denominator $D_g = P_g + F_g^{\mathrm{ref}}$ (Appendix~\ref{sec:app_target}), we compute the endpoint weights $\kappa_{1,g}=H_g^{\mathrm{ref}}/D_g$ and $\kappa_{2,g}=H_g^{\mathrm{ref}}/F_g^{\mathrm{ref}}$, determine the calibrated false-alarm weight $\kappa'_g$ via Equation~\eqref{eq:kappa_fixed}.
SAPT runs for one epoch following Algorithm~\ref{alg:app_alignpre_sapt}. In each update, every condition $c$ samples a single rollout: the first $N-1$ guided Euler steps execute without gradient tracking to obtain the detached state $\bar{x}_* = \operatorname{sg}[x_{t_*}]$ at $t_*=t_{N-1}$. At the final step, the trainable guided model predicts $\hat{x} = f_\theta^{(w)}(\bar{x}_*,t_*,c)$—backpropagating grid-level skill gradients from Equation~\eqref{eq:skill} through all shared network parameters including the historical encoding. The frozen reference model evaluates $\operatorname{sg}[f_\theta^{\mathrm{ref},(w)}(\bar{x}_*,t_*,c)]$ at the identical state $\bar{x}_*$ to compute the velocity anchor in Equation~\eqref{eq:anchor}. 

\begin{algorithm}[!htbp]
\caption{Skill-aligned post-training (SAPT)}
\label{alg:app_alignpre_sapt}
\small
\begin{algorithmic}[1]
\Require Training dataset $\mathcal{D}$, calibration subset $\mathcal{D}_{\mathrm{cal}}\subset\mathcal{D}$, frozen reference $f_\theta^{\mathrm{ref}}$ (fixed weights $\theta^{\mathrm{ref}}$)
\Require Shifted time grid $\{t_n\}_{n=0}^N$, guidance scale $w$, temperature $T$, peak scale $\rho$, anchor weight $\eta$, update budget $U$
\State Initialize trainable parameters $\theta \gets \theta^{\mathrm{ref}}$ from pretrained EMA.
\State Evaluate $f_\theta^{\mathrm{ref},(w)}$ on $\mathcal{D}_{\mathrm{cal}}$ to obtain per-sample counts $H_g^{\mathrm{ref}}, F_g^{\mathrm{ref}}, P_g$ for each pair $g=(l,k)$.
\State Set $D_g = P_g + F_g^{\mathrm{ref}}$, retain active groups $G=\{g : D_g > 10^{-6}\}$, and fix calibrated weights $\kappa'_g$ via Equation~\eqref{eq:kappa_fixed}.
\For{$u = 1, \ldots, U$}
  \State Sample mini-batch $\mathcal{B} = \{(c, x)\} \subset \mathcal{D}$.
  \State For each $(c, x) \in \mathcal{B}$, draw $\epsilon \sim \mathcal{N}(0, I)$ and initialize $x_{t_0} \gets \epsilon$.
  \For{$n = 0, \ldots, N-2$} \Comment{Rollout without gradient tracking}
    \State $\hat{x}^{(n)} \gets f_\theta^{(w)}(x_{t_n}, t_n, c)$
    \State $v_n \gets (\hat{x}^{(n)} - x_{t_n}) / (1 - t_n)$
    \State $x_{t_{n+1}} \gets x_{t_{n}} + (t_{n+1} - t_n) v_n$
  \EndFor
  \State Set $t_* \gets t_{N-1}$ and detach state $\bar{x}_* \gets \operatorname{sg}[x_{t_*}]$.
  \State Predict clean forecast $\hat{x} \gets f_\theta^{(w)}(\bar{x}_*, t_*, c)$. \Comment{Differentiable final step}
  \State Evaluate reference prediction $\hat{x}^{\mathrm{ref}} \gets \operatorname{sg}\big[f_\theta^{\mathrm{ref},(w)}(\bar{x}_*, t_*, c)\big]$.
  \State Compute $\mathcal{L}_{\mathrm{skill}}(\theta)$ on $(\hat{x}, x)$ using Equations~\eqref{eq:soft_loss} and~\eqref{eq:skill}.
  \State Compute $\mathcal{L}_{\mathrm{anchor}}^{\mathrm{last}}(\theta) = \|\hat{x} - \hat{x}^{\mathrm{ref}}\|^2 / (1 - t_*)^2$ via Equation~\eqref{eq:anchor}.
  \State Update $\theta$ using $\nabla_\theta \Big(\frac{1}{|\mathcal{B}|}\sum_{\mathcal{B}}\big[\mathcal{L}_{\mathrm{skill}}(\theta) + \eta\mathcal{L}_{\mathrm{anchor}}^{\mathrm{last}}(\theta)\big]\Big)$ and update EMA of $\theta$.
\EndFor
\Ensure Aligned generator parameters $\theta$ (EMA)
\end{algorithmic}
\end{algorithm}
\subsection{Benchmark Splits, Baseline Configurations \& Evaluation Metrics}
\label{sec:app_alignpre_evaluation}

\paragraph{Benchmark datasets and preprocessing.}
SEVIR~\cite{veillette2020sevir} comprises storm events recorded across the contiguous United States, with each $4\,\mathrm{h}$ event spanning $384\,\mathrm{km}\times384\,\mathrm{km}$ ($49$ frames at $5\,\mathrm{min}$ intervals). We use the vertically integrated liquid (VIL) radar channel normalized from $[0,255]$ to $[0,1]$ by dividing by $255$. Following standard high-resolution protocols~\cite{gao2022earthformer,gong2024cascast}, \textbf{SEVIR-1h} retains the native $384\times384$ ($1\,\mathrm{km}$) grid, predicts $13\to12$ frames ($60\,\mathrm{min}$), and adopts the chronological split of Earthformer~\cite{gao2022earthformer}, providing $35{,}718 / 9{,}060 / 12{,}159$ train/validation/test sequences. Following longer-horizon protocols~\cite{yu2024diffcast,lin2025alphapre,xu2026nowcastdit}, \textbf{SEVIR-2h} bilinearly downsamples frames without antialiasing to $128\times128$ ($3\,\mathrm{km}$) and predicts $5\to20$ frames ($100\,\mathrm{min}$); $25$-frame sequences are extracted from each $49$-frame event with a stride of $5$ frames and split chronologically at 1 January 2019 UTC and 1 October 2019 UTC into $59{,}530 / 28{,}145 / 7{,}220$ train/validation/test sequences.
\textbf{Guangdong-3h} consists of $6\,\mathrm{min}$ composite radar reflectivity mosaics over Guangdong, China, cropped to $360\,\mathrm{km}\times360\,\mathrm{km}$ $40$-frame sequences. Reflectivity values are normalized by $75$, bilinearly resized without antialiasing to $128\times128$ ($2.8125\,\mathrm{km}$ spacing). The task predicts $10\to30$ frames ($180\,\mathrm{min}$), using $70{,}000$ sequences from 2016--2022 for training (with a fixed $512$-sequence validation subset from the training set) and $1{,}500$ sequences from 2023 for testing.

\begin{table*}[tp]
\centering
\small
\setlength{\tabcolsep}{5pt}
\renewcommand{\arraystretch}{1.06}
\caption{\textbf{Summary of baseline configurations across the four model families.}}
\label{tab:app_baseline_configs}
\begin{tabular}{@{}llc@{}}
\toprule
\textbf{Method} & \textbf{Architecture} & \textbf{Generation Space} \\
\midrule
\multicolumn{3}{@{}l}{\textit{(i) Deterministic predictors}} \\
PhyDNet~\cite{guen2020phydnet} & Recurrent PDE-constrained + residual CNN & --- \\
SimVP~\cite{gao2022simvp} & Spatiotemporal Inception-CNN translator & --- \\
Earthformer~\cite{gao2022earthformer} & Hierarchical cuboid space-time transformer & --- \\
NowcastNet~\cite{zhang2023nowcastnet} & Motion-intensity deterministic evolution network & --- \\
AlphaPre~\cite{lin2025alphapre} & Fourier amplitude-phase disentangled forecaster & --- \\
\addlinespace[3pt]
\multicolumn{3}{@{}l}{\textit{(ii) Hybrid generators (deterministic prior + generative module)}} \\
DiffCast~\cite{yu2024diffcast} & PhyDNet predictor + residual diffusion & Pixel \\
CasCast~\cite{gong2024cascast} & Deterministic predictor + cascaded DiT & Latent \\
DuoCast~\cite{wen2026duocast} & SimVP predictor + two-stage diffusion & Pixel + latent \\
\addlinespace[3pt]
\multicolumn{3}{@{}l}{\textit{(iii) Latent-space standalone generators}} \\
PreDiff~\cite{gao2023prediff} & Spatiotemporal UNet diffusion & Latent \\
FlowCast~\cite{ribeiro2026flowcast} & Conditional flow-matching DiT & Latent \\
FREUD~\cite{schusterbauer2026freud} & Rectified-flow DiT + flow decoder & Latent \\
NowcastDiT~\cite{xu2026nowcastdit} & Spatiotemporal DiT + GRPO & Latent \\
\addlinespace[3pt]
\multicolumn{3}{@{}l}{\textit{(iv) Single-scale pixel-space standalone generator}} \\
JiT~\cite{li2026jit} & Single-scale spatiotemporal $x$-prediction DiT & Pixel \\
\bottomrule
\end{tabular}
\end{table*}

\paragraph{Baseline implementation details.}
Table~\ref{tab:app_baseline_configs} summarizes all $13$ baselines across the four model families. NowcastNet~\cite{zhang2023nowcastnet} and NowcastDiT~\cite{xu2026nowcastdit}, whose training implementations are not publicly released, are reimplemented following their published specifications. On SEVIR-1h, Earthformer~\cite{gao2022earthformer}, CasCast~\cite{gong2024cascast} and FREUD~\cite{schusterbauer2026freud} are evaluated using their official released checkpoints, while PreDiff~\cite{gao2023prediff} adapts its released SEVIR-LR ($7\to6$, $128\times128$, $10\,\mathrm{min}$) checkpoint by applying $3\times3$ spatial max-pooling to alternate input frames, followed by bilinear spatial upsampling and linear temporal interpolation anchored at the latest observation. On SEVIR-2h, AlphaPre~\cite{lin2025alphapre} and DuoCast~\cite{wen2026duocast} use their official released weights.
For latent flow and diffusion transformers, FlowCast~\cite{ribeiro2026flowcast} and NowcastDiT are re-trained on SEVIR-2h and Guangdong-3h. Following its published post-training protocol, NowcastDiT optimizes its timestep-aware GRPO reward ($16$ rollouts per group and a $4$-step stochastic window within $10$ sampling steps) and evaluates forecasts with domain guidance (DoG). Finally, JiT~\cite{li2026jit} is implemented directly in spatiotemporal pixel space as a single-scale $x$-prediction transformer, using $32\times32$ pixel patches on SEVIR-1h and $16\times16$ patches on the $128\times128$ benchmarks.

\paragraph{Evaluation metrics.}
Operational event verification is conducted at VIL thresholds $\{16,74,133,160,181,219\}$ on SEVIR and reflectivity thresholds $\{20,30,35,40\}\,\mathrm{dBZ}$ on Guangdong-3h. At each forecast lead time and threshold $\tau$, grid cells with values $\ge\tau$ are binarized as positive events, and hits $H$, misses $M$, false alarms $F$, and correct negatives $C$ are aggregated across all test sequences to compute
\begin{equation}
\begin{gathered}
\mathrm{CSI}=\frac{H}{H+M+F},\\
\mathrm{POD}=\frac{H}{H+M},\\
\mathrm{FAR}=\frac{F}{H+F},\\
\mathrm{HSS}=\frac{2(HC-MF)}{(H+M)(M+C)+(H+F)(F+C)}.
\end{gathered}
\label{eq:app_alignpre_metrics}
\end{equation}
Metrics bearing the suffix \textit{-M} report macro-averages uniformly over all lead times and thresholds, whereas $\mathrm{CSI}_{219}$ and $\mathrm{CSI}_{40}$ average over lead times at the highest threshold of each benchmark. Under standard evaluation conventions, any ratio encountering a zero denominator evaluates to zero (when a smooth deterministic model predicts zero events at an extreme threshold, its nominal zero FAR reflects missed detection rather than false-alarm skill, and must be read jointly with POD and CSI). Perceptual quality is measured via frame-wise SSIM and AlexNet LPIPS~\cite{zhang2018lpips} averaged over all test frames. Deterministic verification and perceptual scores for generative models use single-member forecasts (member $0$) with initial noise seeded deterministically by sequence and member index.
To assess ensemble calibration, we compute the empirical continuous ranked probability score (CRPS) in native physical units (VIL or dBZ) over $K$-member ensembles:
\begin{equation}
\operatorname{CRPS} = \frac{1}{K}\sum_{m=1}^{K}\big|\hat{x}^{(m)} - x\big|
- \frac{1}{2K^2}\sum_{m=1}^{K}\sum_{m'=1}^{K}\big|\hat{x}^{(m)} - \hat{x}^{(m')}\big|,
\label{eq:app_alignpre_crps}
\end{equation}
averaged over all test sequences, lead times, and grid points. Generative models are evaluated with $K=4$ ensemble members under identical checkpoints and sampler settings as their single-member rows. As for deterministic baselines ($K=1$), Equation~\eqref{eq:app_alignpre_crps} reduces to full-test MAE.
\section{Additional Quantitative Results \& Ablation Studies}
\label{sec:app_exp}
\label{sec:app_exp_ablation}

\raggedbottom
\captionsetup{font=footnotesize,skip=4pt}

This appendix expands Section~\ref{sec:exp} of the main text with complete per-threshold benchmark evaluations and detection--false-alarm decompositions (Appendix~\ref{sec:app_exp_main}), extended cross-backbone comparisons of post-training paradigms (Appendix~\ref{sec:app_exp_posttrain}), mathematical formulations and sensitivity analyses for skill-loss calibration (Appendix~\ref{sec:app_kappa_control}), computational accounting across gradient and anchor scopes (Appendix~\ref{sec:app_gradient_anchor}), ensemble calibration diagnostics under varying velocity anchor strengths (Appendix~\ref{sec:app_ablation_anchor}), and inference sensitivity and model scaling studies (Appendix~\ref{sec:app_ablation_grid}). Supplementary qualitative case studies are presented separately in Appendix~\ref{sec:app_exp_qualitative}.


\subsection{Complete Per-Threshold Results \& Detection--False-Alarm Analysis}
\label{sec:app_exp_main}
\label{sec:app_main_analysis}

\subsubsection{Complete benchmark results}
\label{sec:app_exp_full}

Tables~\ref{tab:full_sevir384}--\ref{tab:full_gd} expand Tables~\ref{tab:main_sevir} and~\ref{tab:main_gd} of the main text by reporting frame-wise SSIM and native-grid CSI at every evaluation threshold under identical checkpoints and sampling protocols (Appendix~\ref{sec:app_alignpre_evaluation}). Across all three benchmarks, AlignPre achieves the highest CSI-M, HSS-M, and SSIM, while simultaneously attaining the lowest FAR-M among all probabilistic generators. Specifically, AlignPre surpasses the strongest baseline in CSI-M by $+0.0238$ on SEVIR-1h ($0.4605$ vs.\ $0.4367$ for FlowCast), $+0.0170$ on SEVIR-2h ($0.3169$ vs.\ $0.2999$ for NowcastDiT), and $+0.0274$ on Guangdong-3h ($0.3277$ vs.\ $0.3003$ for NowcastDiT). Relative improvements are most pronounced for severe convection: at the highest evaluation threshold, $\mathrm{CSI}_{219}$ improves by $+15.2\%$ over FlowCast on SEVIR-1h ($0.2041$ vs.\ $0.1772$) and by $+17.6\%$ over DuoCast on SEVIR-2h ($0.0728$ vs.\ $0.0619$), while $\mathrm{CSI}_{40}$ on Guangdong-3h improves by $+25.9\%$ over NowcastDiT ($0.1787$ vs.\ $0.1419$).

\begin{table*}[tp]
\centering
\caption{\textbf{Complete results on SEVIR-1h} ($13\to12$ frames, $60\,\mathrm{min}$; $12{,}159$ test sequences). CRPS uses $K=4$ ensemble members ($K=1$ MAE for deterministic models) in VIL units; all other metrics evaluate single-member forecasts (Appendix~\ref{sec:app_alignpre_evaluation}). Deterministic baselines are shown in gray; among generative models, best and second-best values per column are \textbf{bold} and \underline{underlined}, and deterministic entries strictly exceeding the second-best generative value are also \textbf{bold}.}
\label{tab:full_sevir384}
\footnotesize
\setlength{\tabcolsep}{2.5pt}
\begin{tabular}{@{}l*{13}{c}@{}}
\toprule
 & \multicolumn{7}{c}{Aggregate metrics} & \multicolumn{6}{c}{CSI at each threshold (VIL)} \\
\cmidrule(lr){2-8}\cmidrule(l){9-14}
Method & CSI-M$^{\uparrow}$ & POD-M$^{\uparrow}$ & FAR-M$^{\downarrow}$ & HSS-M$^{\uparrow}$ & SSIM$^{\uparrow}$ & LPIPS$^{\downarrow}$ & CRPS$^{\downarrow}$ & CSI$_{16}^{\uparrow}$ & CSI$_{74}^{\uparrow}$ & CSI$_{133}^{\uparrow}$ & CSI$_{160}^{\uparrow}$ & CSI$_{181}^{\uparrow}$ & CSI$_{219}^{\uparrow}$ \\
\midrule
PhyDNet & \textcolor{gray}{0.3948} & \textcolor{gray}{0.4632} & \textcolor{gray}{\fst{0.3054}} & \textcolor{gray}{0.4915} & \textcolor{gray}{0.7662} & \textcolor{gray}{0.3464} & \textcolor{gray}{6.616} & \textcolor{gray}{0.7578} & \textcolor{gray}{\fst{0.6863}} & \textcolor{gray}{0.3854} & \textcolor{gray}{0.2530} & \textcolor{gray}{0.1962} & \textcolor{gray}{0.0904} \\
SimVP & \textcolor{gray}{0.4027} & \textcolor{gray}{0.4708} & \textcolor{gray}{\fst{0.2698}} & \textcolor{gray}{0.4996} & \textcolor{gray}{0.7833} & \textcolor{gray}{0.3503} & \textcolor{gray}{6.324} & \textcolor{gray}{\fst{0.7651}} & \textcolor{gray}{\fst{0.6890}} & \textcolor{gray}{0.4047} & \textcolor{gray}{0.2659} & \textcolor{gray}{0.2040} & \textcolor{gray}{0.0876} \\
Earthformer & \textcolor{gray}{0.4290} & \textcolor{gray}{0.5113} & \textcolor{gray}{\fst{0.3330}} & \textcolor{gray}{0.5407} & \textcolor{gray}{0.7856} & \textcolor{gray}{0.3564} & \textcolor{gray}{6.373} & \textcolor{gray}{\fst{0.7670}} & \textcolor{gray}{\fst{0.6905}} & \textcolor{gray}{\fst{0.4220}} & \textcolor{gray}{0.3043} & \textcolor{gray}{0.2565} & \textcolor{gray}{0.1338} \\
NowcastNet & \textcolor{gray}{0.4203} & \textcolor{gray}{0.5360} & \textcolor{gray}{0.4148} & \textcolor{gray}{0.5317} & \textcolor{gray}{0.7619} & \textcolor{gray}{\fst{0.1917}} & \textcolor{gray}{7.312} & \textcolor{gray}{0.7473} & \textcolor{gray}{0.6633} & \textcolor{gray}{0.4039} & \textcolor{gray}{0.3057} & \textcolor{gray}{0.2606} & \textcolor{gray}{0.1407} \\
\addlinespace[3pt]
DiffCast & 0.3795 & 0.4903 & 0.4812 & 0.4878 & 0.7512 & \snd{0.2259} & 5.112 & 0.7265 & 0.6217 & 0.3428 & 0.2507 & 0.2157 & 0.1197 \\
CasCast & 0.4168 & \snd{0.5732} & 0.4716 & 0.5340 & 0.7673 & 0.2563 & 5.653 & 0.7400 & 0.6541 & 0.3919 & 0.2978 & 0.2613 & 0.1556 \\
\addlinespace[3pt]
FlowCast & \snd{0.4367} & 0.5403 & \snd{0.3941} & \snd{0.5562} & \snd{0.7889} & 0.2329 & \fst{4.892} & \snd{0.7618} & \snd{0.6740} & \snd{0.4096} & \snd{0.3156} & \snd{0.2819} & \snd{0.1772} \\
NowcastDiT & 0.3948 & 0.5580 & 0.5047 & 0.5077 & 0.7628 & 0.2516 & 5.831 & 0.7363 & 0.6371 & 0.3687 & 0.2630 & 0.2296 & 0.1338 \\
\addlinespace[3pt]
JiT & 0.3473 & 0.4717 & 0.5074 & 0.4485 & 0.6556 & 0.3184 & 6.616 & 0.6066 & 0.6001 & 0.3351 & 0.2380 & 0.1990 & 0.1052 \\
\addlinespace[3pt]
PreDiff & 0.3653 & 0.5071 & 0.5187 & 0.4734 & 0.7139 & 0.2832 & 5.873 & 0.6940 & 0.6036 & 0.3333 & 0.2443 & 0.2063 & 0.1102 \\
FREUD & 0.3646 & 0.4575 & 0.4807 & 0.4684 & 0.7494 & \fst{0.2256} & 5.249 & 0.7153 & 0.6126 & 0.3221 & 0.2321 & 0.1963 & 0.1093 \\
\midrule
AlignPre-Base & 0.4129 & 0.5104 & 0.4140 & 0.5281 & 0.7793 & 0.2411 & \snd{4.906} & 0.7474 & 0.6562 & 0.3822 & 0.2892 & 0.2539 & 0.1482 \\
\rowcolor{oursrow}AlignPre & \fst{0.4605} & \fst{0.5747} & \fst{0.3755} & \fst{0.5836} & \fst{0.7896} & 0.2344 & 5.355 & \fst{0.7625} & \fst{0.6808} & \fst{0.4411} & \fst{0.3531} & \fst{0.3214} & \fst{0.2041} \\
\bottomrule
\end{tabular}
\end{table*}

\begin{table*}[tp]
\centering
\caption{\textbf{Complete results on SEVIR-2h} ($5\to20$ frames, $100\,\mathrm{min}$; $7{,}220$ test sequences). Thresholds and CRPS are in VIL units. Conventions follow Table~\ref{tab:full_sevir384}.}
\label{tab:full_sevir128}
\footnotesize
\setlength{\tabcolsep}{2.5pt}
\begin{tabular}{@{}l*{13}{c}@{}}
\toprule
 & \multicolumn{7}{c}{Aggregate metrics} & \multicolumn{6}{c}{CSI at each threshold (VIL)} \\
\cmidrule(lr){2-8}\cmidrule(l){9-14}
Method & CSI-M$^{\uparrow}$ & POD-M$^{\uparrow}$ & FAR-M$^{\downarrow}$ & HSS-M$^{\uparrow}$ & SSIM$^{\uparrow}$ & LPIPS$^{\downarrow}$ & CRPS$^{\downarrow}$ & CSI$_{16}^{\uparrow}$ & CSI$_{74}^{\uparrow}$ & CSI$_{133}^{\uparrow}$ & CSI$_{160}^{\uparrow}$ & CSI$_{181}^{\uparrow}$ & CSI$_{219}^{\uparrow}$ \\
\midrule
PhyDNet & \textcolor{gray}{0.2552} & \textcolor{gray}{0.3122} & \textcolor{gray}{\fst{0.3574}} & \textcolor{gray}{0.3125} & \textcolor{gray}{0.6427} & \textcolor{gray}{0.3816} & \textcolor{gray}{8.270} & \textcolor{gray}{0.6953} & \textcolor{gray}{\fst{0.5771}} & \textcolor{gray}{0.1696} & \textcolor{gray}{0.0514} & \textcolor{gray}{0.0313} & \textcolor{gray}{0.0063} \\
SimVP & \textcolor{gray}{0.2673} & \textcolor{gray}{0.3270} & \textcolor{gray}{\fst{0.2633}} & \textcolor{gray}{0.3283} & \textcolor{gray}{0.6543} & \textcolor{gray}{0.3830} & \textcolor{gray}{8.138} & \textcolor{gray}{0.6988} & \textcolor{gray}{\fst{0.5861}} & \textcolor{gray}{0.1805} & \textcolor{gray}{0.0667} & \textcolor{gray}{0.0479} & \textcolor{gray}{0.0240} \\
Earthformer & \textcolor{gray}{0.2424} & \textcolor{gray}{0.3031} & \textcolor{gray}{\fst{0.2786}} & \textcolor{gray}{0.2966} & \textcolor{gray}{0.5196} & \textcolor{gray}{0.4138} & \textcolor{gray}{9.417} & \textcolor{gray}{0.6787} & \textcolor{gray}{0.5484} & \textcolor{gray}{0.1351} & \textcolor{gray}{0.0483} & \textcolor{gray}{0.0350} & \textcolor{gray}{0.0089} \\
NowcastNet & \textcolor{gray}{0.2994} & \textcolor{gray}{\fst{0.4370}} & \textcolor{gray}{0.6236} & \textcolor{gray}{0.3816} & \textcolor{gray}{0.6667} & \textcolor{gray}{0.1688} & \textcolor{gray}{9.620} & \textcolor{gray}{0.6850} & \textcolor{gray}{0.5648} & \textcolor{gray}{0.2567} & \textcolor{gray}{\fst{0.1310}} & \textcolor{gray}{\fst{0.0998}} & \textcolor{gray}{0.0592} \\
AlphaPre & \textcolor{gray}{0.2765} & \textcolor{gray}{0.3252} & \textcolor{gray}{\fst{0.3484}} & \textcolor{gray}{0.3414} & \textcolor{gray}{0.6947} & \textcolor{gray}{0.3372} & \textcolor{gray}{7.580} & \textcolor{gray}{\fst{0.7170}} & \textcolor{gray}{\fst{0.5764}} & \textcolor{gray}{0.1963} & \textcolor{gray}{0.0789} & \textcolor{gray}{0.0612} & \textcolor{gray}{0.0293} \\
\addlinespace[3pt]
DiffCast & 0.2713 & 0.3455 & 0.5995 & 0.3490 & 0.6714 & 0.1950 & 6.001 & 0.6704 & 0.5228 & 0.2045 & 0.1015 & 0.0808 & 0.0481 \\
CasCast & 0.2634 & 0.3696 & 0.6464 & 0.3404 & 0.6427 & 0.1858 & 6.704 & 0.6552 & 0.5170 & 0.2084 & 0.0947 & 0.0727 & 0.0326 \\
\addlinespace[3pt]
FlowCast & 0.2902 & 0.3600 & 0.5598 & 0.3734 & 0.6884 & \snd{0.1581} & \fst{5.919} & 0.6869 & 0.5562 & 0.2307 & 0.1137 & 0.0935 & 0.0601 \\
NowcastDiT & \snd{0.2999} & \fst{0.5025} & 0.6465 & \snd{0.3854} & 0.6839 & \fst{0.1549} & 7.102 & \snd{0.6994} & \snd{0.5714} & \snd{0.2576} & \snd{0.1193} & 0.0943 & 0.0572 \\
\addlinespace[3pt]
JiT & 0.2817 & 0.3636 & 0.5852 & 0.3628 & 0.6477 & 0.2002 & 6.523 & 0.6555 & 0.5457 & 0.2258 & 0.1145 & 0.0936 & 0.0553 \\
\addlinespace[3pt]
DuoCast & 0.2818 & 0.3556 & 0.5816 & 0.3594 & 0.6727 & 0.2208 & 6.656 & 0.6725 & 0.5265 & 0.2182 & 0.1156 & \snd{0.0963} & \snd{0.0619} \\
\midrule
AlignPre-Base & 0.2925 & 0.3570 & \snd{0.5407} & 0.3738 & \snd{0.6978} & 0.1888 & \snd{5.935} & 0.6927 & 0.5611 & 0.2337 & 0.1170 & 0.0957 & 0.0549 \\
\rowcolor{oursrow}AlignPre & \fst{0.3169} & \snd{0.3890} & \fst{0.4984} & \fst{0.4076} & \fst{0.6988} & 0.1886 & 6.285 & \fst{0.7003} & \fst{0.5732} & \fst{0.2762} & \fst{0.1518} & \fst{0.1273} & \fst{0.0728} \\
\bottomrule
\end{tabular}
\end{table*}

\begin{table*}[tp]
\centering
\caption{\textbf{Complete results on Guangdong-3h} ($10\to30$ frames, $180\,\mathrm{min}$; $1{,}500$ test sequences). Thresholds and CRPS are in $\mathrm{dBZ}$. Conventions follow Table~\ref{tab:full_sevir384}.}
\label{tab:full_gd}
\footnotesize
\setlength{\tabcolsep}{3.5pt}
\begin{tabular}{@{}l*{11}{c}@{}}
\toprule
 & \multicolumn{7}{c}{Aggregate metrics} & \multicolumn{4}{c}{CSI at each threshold (dBZ)} \\
\cmidrule(lr){2-8}\cmidrule(l){9-12}
Method & CSI-M$^{\uparrow}$ & POD-M$^{\uparrow}$ & FAR-M$^{\downarrow}$ & HSS-M$^{\uparrow}$ & SSIM$^{\uparrow}$ & LPIPS$^{\downarrow}$ & CRPS$^{\downarrow}$ & CSI$_{20}^{\uparrow}$ & CSI$_{30}^{\uparrow}$ & CSI$_{35}^{\uparrow}$ & CSI$_{40}^{\uparrow}$ \\
\midrule
PhyDNet & \textcolor{gray}{0.2297} & \textcolor{gray}{0.2589} & \textcolor{gray}{\fst{0.3443}} & \textcolor{gray}{0.3135} & \textcolor{gray}{0.4510} & \textcolor{gray}{0.5502} & \textcolor{gray}{4.000} & \textcolor{gray}{0.4410} & \textcolor{gray}{0.2698} & \textcolor{gray}{0.1439} & \textcolor{gray}{0.0643} \\
SimVP & \textcolor{gray}{0.2390} & \textcolor{gray}{0.2719} & \textcolor{gray}{\fst{0.3213}} & \textcolor{gray}{0.3250} & \textcolor{gray}{0.3923} & \textcolor{gray}{0.5537} & \textcolor{gray}{4.119} & \textcolor{gray}{0.4490} & \textcolor{gray}{0.2791} & \textcolor{gray}{0.1586} & \textcolor{gray}{0.0691} \\
Earthformer & \textcolor{gray}{0.1622} & \textcolor{gray}{0.1803} & \textcolor{gray}{\fst{0.2899}} & \textcolor{gray}{0.2278} & \textcolor{gray}{0.4640} & \textcolor{gray}{0.6091} & \textcolor{gray}{4.202} & \textcolor{gray}{0.3460} & \textcolor{gray}{0.1852} & \textcolor{gray}{0.0917} & \textcolor{gray}{0.0260} \\
NowcastNet & \textcolor{gray}{0.2586} & \textcolor{gray}{\fst{0.4913}} & \textcolor{gray}{0.6715} & \textcolor{gray}{0.3396} & \textcolor{gray}{0.4386} & \textcolor{gray}{0.3522} & \textcolor{gray}{6.899} & \textcolor{gray}{0.4103} & \textcolor{gray}{0.3036} & \textcolor{gray}{0.2023} & \textcolor{gray}{0.1183} \\
\addlinespace[3pt]
DiffCast & 0.2550 & 0.3579 & 0.5942 & 0.3489 & 0.5289 & 0.2537 & 2.735 & 0.4307 & 0.2917 & 0.1872 & 0.1104 \\
CasCast & 0.2419 & 0.3885 & 0.6545 & 0.3292 & 0.4965 & 0.2846 & 3.279 & 0.4099 & 0.2805 & 0.1817 & 0.0954 \\
\addlinespace[3pt]
FlowCast & 0.2808 & 0.3664 & \snd{0.5191} & 0.3849 & 0.5684 & \snd{0.2380} & \snd{2.667} & 0.4642 & 0.3264 & 0.2133 & 0.1192 \\
NowcastDiT & \snd{0.3003} & \fst{0.4540} & 0.5763 & \snd{0.4078} & \snd{0.5707} & \fst{0.2379} & 2.836 & \snd{0.4751} & \snd{0.3458} & \snd{0.2383} & \snd{0.1419} \\
\addlinespace[3pt]
JiT & 0.2599 & 0.3836 & 0.6084 & 0.3526 & 0.5016 & 0.2652 & 2.835 & 0.4290 & 0.2976 & 0.1965 & 0.1164 \\
\midrule
AlignPre-Base & 0.2778 & 0.3577 & 0.5192 & 0.3792 & 0.5629 & 0.2656 & \fst{2.661} & 0.4551 & 0.3231 & 0.2104 & 0.1225 \\
\rowcolor{oursrow}AlignPre & \fst{0.3277} & \snd{0.4411} & \fst{0.4926} & \fst{0.4438} & \fst{0.5741} & 0.2661 & 2.763 & \fst{0.4792} & \fst{0.3758} & \fst{0.2770} & \fst{0.1787} \\
\bottomrule
\end{tabular}
\end{table*}

\paragraph{Deterministic underprediction and matched-guidance decomposition.}
Regression-trained deterministic baselines (PhyDNet, SimVP, Earthformer, and AlphaPre) exhibit lower nominal FAR-M than generative forecasters because progressive spatial smoothing suppresses high-intensity predictions altogether. On SEVIR-2h at $\mathrm{VIL}\ge219$, for instance, their predicted event area is only $1\text{--}5\%$ of observed storm coverage (compared with $24\%$ for AlignPre), forecasting zero events on $10\text{--}14$ of the $20$ lead times where zero-denominator FAR defaults to zero alongside near-zero skill ($\mathrm{CSI}_{219}\le0.0293$).
Separating classifier-free guidance (CFG) from SAPT further clarifies how AlignPre achieves joint precision--recall gains. Relative to unguided AlignPre-Base, applying CFG raises CSI-M by $+0.011\text{--}0.017$ across the three benchmarks, but simultaneously inflates FAR-M by $+0.016\text{--}0.028$, consistent with guidance expanding predicted precipitation coverage~\cite{schusterbauer2026freud}. Starting from this guided base, SAPT increases CSI-M by an additional $+0.012\text{--}0.037$ while reducing FAR-M by $0.043\text{--}0.067$, with LPIPS changing by at most $0.005$. Concurrently, four-member empirical CRPS increases slightly over the guided base ($5.166\to5.355\,\mathrm{VIL}$ on SEVIR-1h, $6.213\to6.285\,\mathrm{VIL}$ on SEVIR-2h, and $2.727\to2.763\,\mathrm{dBZ}$ on Guangdong-3h), reflecting the expected trade-off between sharpening discrete thresholded event boundaries and continuous distributional spread. Without deterministic branches or external autoencoders, unguided AlignPre-Base already outperforms the single-scale pixel baseline JiT by $+0.011\text{--}0.066$ in CSI-M across benchmarks, and guided AlignPre-Base exceeds all hybrid baselines.

\subsubsection{Detection--false-alarm and lead-time decomposition}
\label{sec:app_detection_analysis}
\label{sec:app_leadtime_analysis}

\begin{figure*}[tp]
\centering
\includegraphics[width=\textwidth]{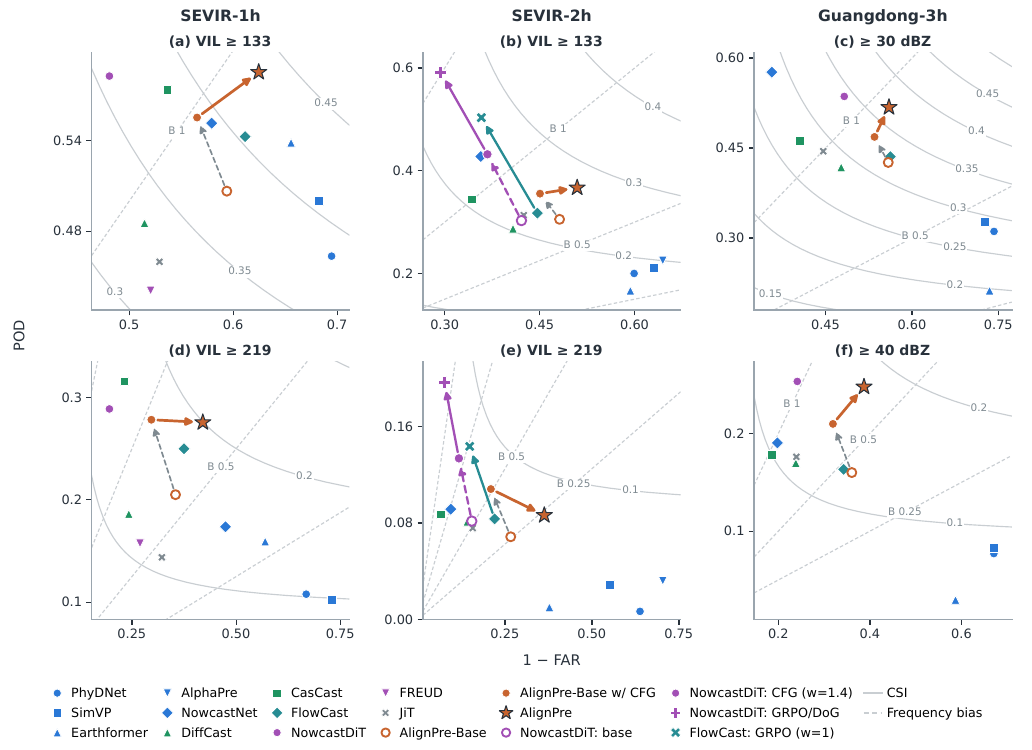}
\caption{\textbf{Roebber performance diagrams across benchmarks and thresholds~\cite{roebber2009performance}.} From left to right, the columns show SEVIR-1h, SEVIR-2h, and Guangdong-3h. The top row uses moderate thresholds ($\mathrm{VIL}\ge133$, $\ge30\,\mathrm{dBZ}$), and the bottom row uses extreme thresholds ($\mathrm{VIL}\ge219$, $\ge40\,\mathrm{dBZ}$). Contingency counts are aggregated over lead times. Solid gray curves denote CSI isolines, and dashed lines denote frequency bias $B=(H+F)/(H+M)$ (upper-right indicates higher skill and precision). Orange trajectories trace $\text{AlignPre-Base} \dashrightarrow \text{AlignPre-Base w/ CFG} \rightarrow \text{AlignPre}$ across all panels, while the SEVIR-2h panels additionally trace $\text{NowcastDiT: base} \dashrightarrow \text{CFG} \rightarrow \text{GRPO/DoG}$ and $\text{FlowCast} \rightarrow \text{GRPO}$.}
\label{fig:perf_diagram}
\label{fig:perf_diagram_full}
\end{figure*}

Figure~\ref{fig:perf_diagram_full} decomposes forecast skill into probability of detection ($\mathrm{POD}$) and success ratio ($1-\mathrm{FAR}$) across moderate and extreme thresholds. In every panel, applying CFG moves AlignPre-Base upward and leftward, increasing recall and frequency bias $B$ at the expense of more false alarms. SAPT then consistently steers the guided base rightward toward higher precision ($1-\mathrm{FAR}$), while simultaneously boosting POD at moderate thresholds ($\mathrm{VIL}\ge133$ and $\ge30\,\mathrm{dBZ}$) and extreme Guangdong-3h reflectivity ($\ge40\,\mathrm{dBZ}$), and curbing high-threshold false alarms at $\mathrm{VIL}\ge219$ to elevate or preserve CSI. By contrast, on SEVIR-2h (Figure~\ref{fig:perf_diagram_full}(b,e)), GRPO post-training drives both NowcastDiT and FlowCast sharply toward the upper-left corner—inflating recall via excessive false alarms ($B>1$).
Resolved across individual lead times under matched guidance in Figure~\ref{fig:leadtime} of the main text, AlignPre reduces FAR below AlignPre-Base at every native forecast lead time—by $0.06\text{--}0.08$ on SEVIR-1h, $0.04\text{--}0.10$ on SEVIR-2h, and $0.04\text{--}0.05$ on Guangdong-3h. Concurrently, CSI increases across all lead times on SEVIR-1h and all lead times after the initial step on SEVIR-2h and Guangdong-3h, reaching peak gains at intermediate horizons ($+0.044$ at $+20\,\mathrm{min}$, $+0.011$ at $+25\,\mathrm{min}$, and $+0.042$ at $+54\,\mathrm{min}$) and maintaining clear positive margins at the end of each horizon ($+0.035$ at $+60\,\mathrm{min}$, $+0.005$ at $+100\,\mathrm{min}$, and $+0.028$ at $+180\,\mathrm{min}$).
\subsection{Extended Comparison of Post-Training Paradigms}
\label{sec:app_exp_posttrain}

\subsubsection{Protocols and cross-backbone test results}
\label{sec:app_posttrain_protocols}
\label{sec:app_posttrain_analysis}

We compare SAPT against group-relative policy optimization (GRPO)~\cite{xu2026nowcastdit,li2026mixgrpo} and two-stage sequential preference optimization (SPO)~\cite{xu2026syncast} on SEVIR-2h across pixel-space AlignPre-Base and the latent-space transformers NowcastDiT and FlowCast. Starting from the same pretrained checkpoint on each backbone, every backbone--method pair is run for three independent post-training repeats. To ensure a uniform protocol across pixel-space and frozen-VAE latent backbones, SAPT fixes the calibrated false-alarm weight $\kappa'$, samples one trajectory per condition, and applies full-chain gradient backpropagation with all-step velocity anchoring for $1{,}365$ updates. GRPO follows the timestep-aware formulation and composite reward of NowcastDiT, sampling $16$ trajectories per group ($4$-step stochastic exploration window) for $1{,}000$ updates (global batch $8$). Following SynCast~\cite{xu2026syncast}, SPO runs $1{,}000$ Diffusion-DPO updates for FAR preference in Stage~1, and resumes from the Stage-1 checkpoint attaining the lowest validation FAR-M under LPIPS and SSIM fidelity constraints for $360$ Diffusion-SPO updates targeting CSI preference in Stage~2 (global batch $8$), placing the Stage-1-to-Stage-2 transition at $1000/1360\approx73.5\%$ of the update budget in Figure~\ref{fig:posttrain_curves}. On AlignPre-Base and NowcastDiT, SAPT uses classifier-free guidance (CFG) while GRPO and SPO use domain guidance following their published protocols; all three methods run unguided on FlowCast. Validation curves in Figure~\ref{fig:posttrain_curves} evaluate EMA checkpoints on the fixed $512$-sequence validation subset, plotting stage-wise means and one-standard-deviation bands estimated via $1{,}000$ bootstrap resamples ($512$ sequences each) across the three repeats; re-evaluating $37$ saved checkpoints across three inference seeds confirms negligible sampling variance ($\mathrm{SD}\le0.0005$ for CSI-M and LPIPS, and $\le0.0026$ for FAR-M). For the cross-paradigm test comparisons in Figure~\ref{fig:posttrain_bars} and Tables~\ref{tab:app_posttrain}--\ref{tab:posttrain_grpo}, checkpoints are selected using the $\kappa_1$ skill criterion under this shared full-rollout protocol, whereas main benchmark Tables~\ref{tab:main_sevir} and~\ref{tab:full_sevir128} report the full-epoch endpoint of default final-step SAPT.

\begin{table*}[tp]
\centering
\caption{\textbf{Cross-backbone comparison of post-training methods on SEVIR-2h.} Unshaded base rows report pretrained performance without and with classifier-free guidance (CFG), matching Figure~\ref{fig:posttrain_bars}. Best post-trained values within each backbone are \textbf{bold}.}
\label{tab:app_posttrain}
\footnotesize
\setlength{\tabcolsep}{8pt}
\begin{tabular}{@{}l*{5}{c}@{}}
\toprule
Backbone \& Method & CSI-M$^{\uparrow}$ & CSI$_{219}^{\uparrow}$ & FAR-M$^{\downarrow}$ & POD-M$^{\uparrow}$ & LPIPS$^{\downarrow}$ \\
\midrule
\multicolumn{6}{@{}l}{\textit{AlignPre-Base (end-to-end pixel space)}}\\
\hspace{0.5em}Pretrained Base & 0.2925 & 0.0549 & 0.5407 & 0.3570 & 0.1888 \\
\hspace{0.5em}Pretrained Base w/ CFG & 0.3045 & 0.0728 & 0.5635 & 0.3904 & 0.1836 \\
\hspace{0.5em}+ Two-stage SPO & 0.3039 & 0.0645 & 0.5305 & 0.3771 & 0.1955 \\
\hspace{0.5em}+ GRPO & 0.3160 & 0.0787 & 0.6189 & \fst{0.4939} & \fst{0.1691} \\
\rowcolor{oursrow}\hspace{0.5em}+ SAPT & \fst{0.3272} & \fst{0.0958} & \fst{0.5292} & 0.4179 & 0.1913 \\
\addlinespace[4pt]
\multicolumn{6}{@{}l}{\textit{NowcastDiT (latent space, frozen VAE decoder)}}\\
\hspace{0.5em}Pretrained Base & 0.2799 & 0.0531 & 0.5836 & 0.3520 & 0.1625 \\
\hspace{0.5em}Pretrained Base w/ CFG & 0.3013 & 0.0632 & 0.6185 & 0.4379 & 0.1580 \\
\hspace{0.5em}+ Two-stage SPO & 0.3196 & 0.0673 & \fst{0.5628} & 0.4266 & 0.2588 \\
\hspace{0.5em}+ GRPO & 0.2999 & 0.0572 & 0.6465 & \fst{0.5025} & \fst{0.1549} \\
\rowcolor{oursrow}\hspace{0.5em}+ SAPT & \fst{0.3223} & \fst{0.0847} & 0.5753 & 0.4434 & 0.1718 \\
\addlinespace[4pt]
\multicolumn{6}{@{}l}{\textit{FlowCast (latent space, frozen VAE decoder)}}\\
\hspace{0.5em}Pretrained Base & 0.2902 & 0.0601 & 0.5598 & 0.3600 & 0.1581 \\
\hspace{0.5em}+ Two-stage SPO & 0.2976 & 0.0562 & 0.5089 & 0.3549 & 0.2152 \\
\hspace{0.5em}+ GRPO & 0.3160 & 0.0788 & 0.5974 & \fst{0.4526} & \fst{0.1502} \\
\rowcolor{oursrow}\hspace{0.5em}+ SAPT & \fst{0.3193} & \fst{0.0800} & \fst{0.5027} & 0.3881 & 0.1868 \\
\bottomrule
\end{tabular}
\end{table*}

\paragraph{Precision--recall trade-offs and perceptual fidelity.}
Table~\ref{tab:app_posttrain} presents the complete test metrics corresponding to Figure~\ref{fig:posttrain_bars}. Optimizing NowcastDiT's composite reward (which combines low-threshold pooled CSI with perceptual SSIM and LPIPS terms~\cite{xu2026nowcastdit}), GRPO lowers LPIPS across all backbones but systematically over-expands predicted precipitation areas: POD-M surges to $0.4939$, $0.5025$, and $0.4526$ at the expense of severe false alarms, driving FAR-M up to $0.6189$ on AlignPre-Base, $0.6465$ on NowcastDiT, and $0.5974$ on FlowCast. On NowcastDiT, where guidance already expands storm coverage, further indiscriminate expansion fails to improve CSI-M ($0.2999$ vs.\ $0.3013$ for the guided base) and degrades extreme-precipitation $\mathrm{CSI}_{219}$ ($0.0572$ vs.\ $0.0632$). Conversely, two-stage SPO reduces FAR-M across backbones by contracting predicted rain areas, which lowers POD-M relative to the corresponding base on AlignPre-Base ($0.3904\to0.3771$) and FlowCast ($0.3600\to0.3549$) and causes substantial perceptual blurring on the latent models (LPIPS worsens from $0.1580$ to $0.2588$ on NowcastDiT and from $0.1581$ to $0.2152$ on FlowCast), as preference pairs frequently favor smooth, ensemble-mean-like forecasts~\cite{xu2026syncast}. Running Stage-1 FAR-DPO without Stage-2 CSI recovery collapses forecast coverage even further, driving validation FAR-M down to $0.1185\pm0.0252$ while crashing CSI-M to $0.1854\pm0.0055$ and LPIPS to $0.3426\pm0.0050$ across the three repeats. In contrast, SAPT calibrates event boundaries via reference contingency statistics to simultaneously raise POD-M and lower FAR-M relative to the corresponding base across all three generators, achieving the highest CSI-M ($0.3272$, $0.3223$, and $0.3193$) and $\mathrm{CSI}_{219}$ ($0.0958$, $0.0847$, and $0.0800$). Moreover, SAPT attains its strongest overall and extreme-event skill on AlignPre-Base, confirming the benefit of end-to-end pixel-space intensity calibration without a frozen latent decoder bottleneck.

\begin{table*}[tp]
\centering
\caption{\textbf{Decoupling reward objective from optimization paradigm on AlignPre-Base} on SEVIR-2h. GRPO optimizes either the NowcastDiT reward (\textit{NCD}) or our negative weighted error count (\textit{our cost}, Equation~\eqref{eq:cost}) using group size $G=16$ and stochastic window $K=4$ ($^{\ast}G=4, K=2$). Best values per column are \textbf{bold}.}
\label{tab:posttrain_grpo}
\footnotesize
\setlength{\tabcolsep}{6pt}
\begin{tabular}{@{}l*{7}{c}@{}}
\toprule
Method & CSI-M$^{\uparrow}$ & CSI$_{219}^{\uparrow}$ & FAR-M$^{\downarrow}$ & HSS-M$^{\uparrow}$ & LPIPS$^{\downarrow}$ & SSIM$^{\uparrow}$ & Cond. \\
\midrule
Pretrained Base w/ CFG & 0.3045 & 0.0728 & 0.5635 & 0.3911 & 0.1836 & 0.6952 & -- \\
GRPO (NCD) & 0.3160 & 0.0787 & 0.6189 & 0.4079 & \fst{0.1691} & 0.6141 & 4,000 \\
GRPO (our cost) & 0.3268 & 0.0871 & \fst{0.5008} & 0.4203 & 0.2558 & 0.6372 & 8,000 \\
GRPO (our cost)$^{\ast}$ & 0.3254 & 0.0863 & 0.5100 & 0.4179 & 0.2324 & 0.6435 & 4,000 \\
\rowcolor{oursrow}SAPT & \fst{0.3272} & \fst{0.0958} & 0.5292 & \fst{0.4229} & 0.1913 & \fst{0.7041} & \fst{1,536} \\
\bottomrule
\end{tabular}
\end{table*}

\subsubsection{Decoupling objective design and convergence trajectories}
\label{sec:app_posttrain_progress}

\begin{figure}[!htbp]
\centering
\includegraphics[width=.38\textwidth]{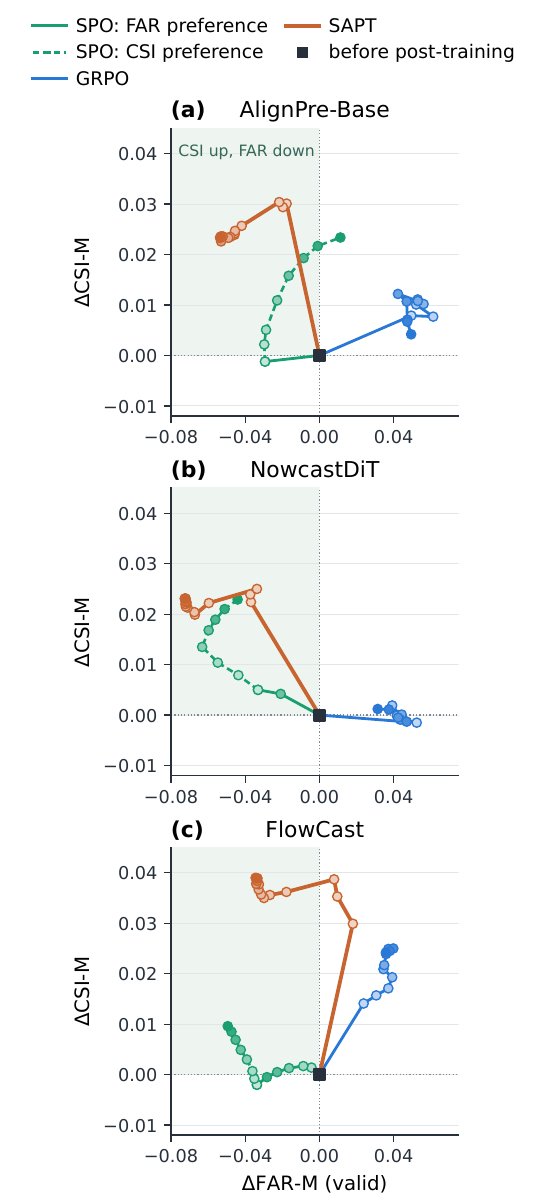}
\caption{\textbf{Validation trajectories in the $(\Delta\mathrm{FAR\text{-}M}, \Delta\mathrm{CSI\text{-}M})$ plane} on SEVIR-2h ($512$-sequence validation subset). Coordinates measure changes relative to each backbone's corresponding pretrained model (black square); the shaded upper-left quadrant indicates simultaneous CSI improvement and FAR reduction. Marker color deepens with post-training updates; dashed green curves trace SPO Stage~2 initialized from the selected Stage-1 FAR checkpoint.}
\label{fig:posttrain_compare}
\end{figure}

Table~\ref{tab:posttrain_grpo} isolates the effect of our reference-calibrated cost objective from the gradient-based optimization mechanism on AlignPre-Base. Replacing the NowcastDiT reward in GRPO with the negative weighted error count $-J_{\kappa'}$ from Equation~\eqref{eq:cost} eliminates false-alarm inflation, lowering FAR-M from $0.6189$ to $0.5008$ while nearly matching SAPT in CSI-M ($0.3268$ vs.\ $0.3272$). This confirms that our analytic linear cost formulation governs the desired precision--recall direction even under black-box policy gradients. Nevertheless, trajectory-level policy optimization requires $2.6\text{--}5.2\times$ more input conditions ($4{,}000\text{--}8{,}000$ vs.\ $1{,}536$, before accounting for $G$-fold group rollouts) and severely degrades perceptual fidelity—raising LPIPS by $+0.0722$ ($0.1836\to0.2558$) and dropping SSIM by $0.0580$ ($0.6952\to0.6372$)—whereas SAPT preserves fine-scale realism ($\mathrm{LPIPS}=0.1913$, $\mathrm{SSIM}=0.7041$) and achieves superior extreme-event skill ($\mathrm{CSI}_{219}=0.0958$) via grid-level gradients and velocity anchoring.
Figure~\ref{fig:posttrain_compare} further illustrates sample efficiency in the $(\Delta\mathrm{FAR\text{-}M}, \Delta\mathrm{CSI\text{-}M})$ plane: within the first $516$ training conditions, SAPT already lifts validation CSI-M by $+0.030$, $+0.022$, and $+0.030$ across AlignPre-Base, NowcastDiT, and FlowCast—entering the joint-improvement quadrant immediately on the guided bases and turning into it right after recovering initial recall on unguided FlowCast—whereas GRPO drifts rightward into high-FAR territory and SPO requires two-stage switching at substantial perceptual cost.

\subsection{Skill-Loss Formulations \& Cost Calibration}
\label{sec:app_kappa_control}
\label{sec:app_matched_cost_controls}

\paragraph{Formulations of ablated cost rules and skill losses.}
To isolate objective design from gradient truncation, Figures~\ref{fig:kappa_curves} and~\ref{fig:skill_loss_curves} and Table~\ref{tab:kappa_control} evaluate SAPT on AlignPre-Base on SEVIR-2h across $1{,}365$ updates ($16{,}380$ conditions, $\eta=1.1$) under full-chain gradient backpropagation and all-step velocity anchoring, matching the validation protocol in Appendix~\ref{sec:app_exp_posttrain}. Using the offline reference counts $H_g^{\mathrm{ref}}$, $F_g^{\mathrm{ref}}$, $P_g$, and reference denominator $D_g = P_g + F_g^{\mathrm{ref}}$ for each active lead-time--threshold group $g=(l,k)\in G$, the compared variants are defined as follows:
\begin{itemize}
  \item \textbf{Endpoint, peak-scale, and curriculum rules (Figure~\ref{fig:kappa_curves}, Table~\ref{tab:kappa_control}):} \textit{Fixed $\kappa_1$} sets $\kappa_g = \kappa_{1,g} = H_g^{\mathrm{ref}}/D_g$ (targeting CSI alone), and \textit{Fixed $\kappa_2$} sets $\kappa_g = \kappa_{2,g} = H_g^{\mathrm{ref}}/F_g^{\mathrm{ref}}$ (targeting FAR alone). Writing $r_{l,k} = \kappa_{1,l,k} / \max_{l'} \kappa_{1,l',k}$, \textit{Peak $\rho=0.8$} and \textit{Peak $\rho=1.2$} evaluate clipped peak-scaled weights $\operatorname{clip}_{[\kappa_{1,l,k},\,\kappa_{2,l,k}]}(\rho\,r_{l,k})$ around the default $\rho=0.99$. \textit{Two-stage $\kappa_1\to\kappa'$} initializes weights at $\kappa_{1,g}$ and switches to $\kappa'_g$ after $1{,}600$ training conditions (marked by the vertical dotted line in Figure~\ref{fig:kappa_curves}). Table~\ref{tab:kappa_control} further evaluates \textit{Constant over lead times}, which sets $\kappa_{l,k}=c_k$ per threshold $k$ to match the threshold's normalized reference false-alarm sum $\sum_l \kappa'_{l,k} F_{l,k}^{\mathrm{ref}}/D_{l,k}$, and \textit{Shared multiplier}, which sets $\kappa_{l,k} = m\,\kappa_{1,l,k}$ with a single scalar $m$ matching the global reference sum $\sum_{l,k}\kappa'_{l,k} F_{l,k}^{\mathrm{ref}}/D_{l,k}$.
  \item \textbf{Mean and midpoint scaling controls (Figure~\ref{fig:skill_loss_curves}, Table~\ref{tab:kappa_control}):} \textit{Mean normalization} replaces lead-time peak scaling with lead-time mean scaling $\kappa_{l,k} = s_{\mathrm{mean}}\,\kappa_{1,l,k} / (\frac{1}{L}\sum_{l'} \kappa_{1,l',k})$, calibrating $s_{\mathrm{mean}}$ to match the global sum $\sum_{l,k}\kappa'_{l,k} F_{l,k}^{\mathrm{ref}}/D_{l,k}$. \textit{Midpoint} sets $\kappa_g = (\kappa_{1,g}+\kappa_{2,g})/2$, which disproportionately amplifies false-alarm gradients in high-threshold groups where $F_g^{\mathrm{ref}}$ is small ($\kappa_{2,g}\gg \kappa_{1,g}$).
  \item \textbf{Alternative skill objectives and denominator normalization (Figure~\ref{fig:skill_loss_curves}):} \textit{$\kappa\equiv1$} fixes unit false-alarm weight across all groups while retaining groupwise reference-denominator normalization $D_g^{-1}(\tilde{M}_g + \tilde{F}_g)$, where $\tilde{M}_g+\tilde{F}_g = \sum_i (s_i - y_i)^2$ for $s_i = \sigma((\hat{x}_i-\tau_k)/T)$ and $y_i = \mathbf{1}[x_i\ge\tau_k]$. \textit{$\mathrm{AT}^2$ loss}~\cite{choi2026atloss} adapts squared event penalties to multi-threshold post-training by keeping $\kappa_g=1$ and replacing group-specific $D_g$ with their shared harmonic mean $\bar{D}_H = |G| \big/ \sum_{g\in G} D_g^{-1}$:
  \begin{equation}
  \mathcal{L}_{\mathrm{AT}^2}(\theta) = \frac{1}{|G|\,\bar{D}_H}\sum_{g\in G}\sum_{i}\big(s_{g,i}(\theta)-y_{g,i}\big)^2.
  \label{eq:app_at2}
  \end{equation}
  Finally, \textit{MSE skill loss} replaces $\mathcal{L}_{\mathrm{skill}}(\theta)$ with $c_{\mathrm{mse}}\|\hat{x}-x\|^2$, where $c_{\mathrm{mse}}$ is calibrated at initialization to match the parameter-gradient norm of $\mathcal{L}_{\mathrm{skill}}$.
\end{itemize}

\begin{table*}[t]
\centering
\caption{\textbf{Controlled comparison of fixed false-alarm weight rules on SEVIR-2h} ($512$-sequence validation endpoints at update $1{,}365$, shared calibration subset and training seed). Shading marks the default SAPT rule.}
\label{tab:kappa_control}
\footnotesize
\setlength{\tabcolsep}{6pt}
\begin{tabular}{@{}l*{4}{c}@{}}
\toprule
False-alarm weight rule & CSI-M$^{\uparrow}$ & FAR-M$^{\downarrow}$ & LPIPS$^{\downarrow}$ & POD-M$^{\uparrow}$ \\
\midrule
\rowcolor{oursrow}Fixed $\kappa'$ ($\rho=0.99$, default) & 0.3646 & 0.4662 & 0.2080 & 0.4642 \\
Constant over lead times & 0.3635 & 0.4712 & 0.2079 & 0.4691 \\
Shared multiplier across thresholds & 0.3680 & 0.4731 & 0.2085 & 0.4728 \\
Mean normalization over lead times & 0.3598 & 0.4583 & 0.2076 & 0.4560 \\
Peak scale $\rho=0.8$ & 0.3683 & 0.4783 & 0.2076 & 0.4787 \\
Peak scale $\rho=1.2$ & 0.3600 & 0.4541 & 0.2087 & 0.4500 \\
Fixed $\kappa_2$ endpoint & 0.3445 & 0.4184 & 0.2192 & 0.4051 \\
\bottomrule
\end{tabular}
\end{table*}

\paragraph{Calibration trade-offs and normalization effects.}
Table~\ref{tab:kappa_control} complements Figure~\ref{fig:kappa_curves} under strictly matched calibration statistics and random seeds. Optimizing the FAR endpoint $\kappa_2$ yields the lowest FAR-M ($0.4184$) but overly suppresses storm detection ($\mathrm{POD\text{-}M}=0.4051$, $\mathrm{CSI\text{-}M}=0.3445$) and degrades LPIPS ($0.2192$). Varying $\rho$ from $0.8$ to $1.2$ smoothly trades recall and CSI-M ($0.3683\to0.3600$) for stricter false-alarm suppression ($0.4783\to0.4541$) at stable LPIPS, while $\rho=0.99$ achieves a balanced operating point that outperforms lead-time-constant and mean-normalized weights in CSI-M and matches the two-stage $\kappa_1\to\kappa'$ curriculum without scheduling overhead. In Figure~\ref{fig:skill_loss_curves}, replacing groupwise $D_g$ scaling in $\kappa\equiv1$ with the shared denominator $\bar{D}_H$ in $\mathrm{AT}^2$ allows widespread low-threshold rain to dominate gradients and severely inflates LPIPS to $0.3044$ ($\mathrm{CSI\text{-}M}=0.3442$), whereas continuous MSE blurs fine structure ($\mathrm{LPIPS}=0.2427$, $\mathrm{CSI\text{-}M}=0.3448$) and uncalibrated $\kappa\equiv1$ underperforms in CSI-M, confirming that both calibrated $\kappa'_g$ and groupwise reference-denominator normalization are essential. Finally, varying the offline calibration subset $\mathcal{D}_{\mathrm{cal}}$ across evenly spaced $512$, randomly sampled $512$, and randomly sampled $1{,}024$ unaugmented training sequences yields tightly clustered validation endpoints ($\mathrm{CSI\text{-}M}=0.3648\pm0.0005$, $\mathrm{FAR\text{-}M}=0.4652\pm0.0026$, $\mathrm{LPIPS}=0.2077\pm0.0003$), confirming that $512$ offline sequences suffice for reliable calibration.

\subsection{Gradient Depth, Anchor Scope \& Computational Cost Accounting}
\label{sec:app_gradient_anchor}

\begin{table*}[t]
\centering
\caption{\textbf{Gradient depth and velocity-anchor scope on SEVIR-2h} ($512$-sequence validation endpoints at update $1{,}365$, $N=10$, with CFG). \textit{NFEs} list gradient-enabled / no-grad network batch evaluations per condition (counting both CFG branches), and \textit{TFLOPs/sample} count denoiser matrix and attention operations.}
\label{tab:gradient_anchor}
\footnotesize
\setlength{\tabcolsep}{5pt}
\begin{tabular}{@{}llccccc@{}}
\toprule
Grad. & Anchor & CSI-M$^{\uparrow}$ & FAR-M$^{\downarrow}$ & LPIPS$^{\downarrow}$ & NFEs (grad / no-grad) & TFLOPs/sample$^{\downarrow}$ \\
\midrule
Full & All & 0.3643 & 0.4630 & \snd{0.2074} & $20\;/\;40$ & 121.67 \\
Full & Last & \fst{0.3785} & \fst{0.4452} & 0.2595 & $20\;/\;22$ & \snd{102.08} \\
Last & All & 0.3611 & 0.4645 & \fst{0.2072} & $20\;/\;40$ & 121.67 \\
\rowcolor{oursrow}Last & Last & \snd{0.3731} & \snd{0.4604} & 0.2147 & $\fst{2\;/\;22}$ & \fst{32.15} \\
\bottomrule
\end{tabular}
\end{table*}

\paragraph{Scope interactions and per-sample FLOP accounting.}
Table~\ref{tab:gradient_anchor} reports the common-budget validation endpoints and computational costs corresponding to Figure~\ref{fig:gradient_anchor_curves} of the main text. Under $N=10$ guided Euler steps, truncating skill gradients while retaining an all-step velocity anchor (\textit{Last / All}, evaluating detached intermediate states) still requires parameter-gradient backpropagation at all $10$ steps, yielding zero FLOP reduction ($121.67\,\mathrm{TFLOPs/sample}$) while over-constraining CSI-M ($0.3611$). Conversely, backpropagating skill gradients through the full rollout while anchoring only the final step (\textit{Full / Last}) leaves intermediate transitions unregularized, causing continual perceptual drift ($\mathrm{LPIPS}$ rises to $0.2595$). Restricting both skill gradients and velocity anchoring to the final step (\textit{Last / Last}) reduces denoiser compute by $73.6\%$ ($32.15\,\mathrm{TFLOPs/sample}$) and cuts steady-state median wall-clock update time by $63.5\%$ ($16.91\,\mathrm{s}\to6.17\,\mathrm{s}$ per update under identical multi-GPU batch layouts), while surpassing \textit{Full / All} in both validation CSI-M ($0.3731$ vs.\ $0.3643$) and FAR-M ($0.4604$ vs.\ $0.4630$).

To compute hardware-independent denoiser TFLOPs per training sample in Table~\ref{tab:gradient_anchor}, we count all matrix multiplications and attention contractions ($1\text{ multiply-add}=2\,\mathrm{FLOPs}$) across both CFG branches on the SEVIR-2h \textrm{B} model. To avoid storing multi-step intermediate activations in memory, the training pipeline encodes the historical K/V cache once per model in no-grad mode, executes a $10$-step no-grad rollout to record trajectory states, evaluates the frozen reference model without gradients on the anchored steps, and recomputes forward--backward passes with per-step activation checkpointing at the trainable steps. Let $H=0.21638\,\mathrm{TFLOPs}$ denote one historical K/V cache encoding, $F=2.17693\,\mathrm{TFLOPs}$ denote the sum of one conditional (cached) and one unconditional no-grad future forward pass, and $T_c=7.77020\,\mathrm{TFLOPs}$ ($T_d=7.76970\,\mathrm{TFLOPs}$) denote a complete guided trainable forward--backward pass with (without) state-input differentiation, including historical re-encoding. Across $N=10$ steps, the four configurations evaluate to:
\begin{equation}
\begin{aligned}
\text{Full / All:} &\quad 2H + 20F + 9T_c + T_d = 121.67\,\mathrm{TFLOPs},\\
\text{Full / Last:} &\quad 2H + 11F + 9T_c + T_d = 102.08\,\mathrm{TFLOPs},\\
\text{Last / All:} &\quad 2H + 20F + 10T_d = 121.67\,\mathrm{TFLOPs},\\
\text{Last / Last:} &\quad 2H + 11F + T_d = 32.15\,\mathrm{TFLOPs}.
\end{aligned}
\label{eq:app_tflops_accounting}
\end{equation}
In terms of network batch evaluations (NFEs) per condition across both CFG branches, no-grad rollout and reference passes total $2(10+10)=40$ for all-step anchoring versus $2(10+1)=22$ for last-step anchoring, while trainable forward--backward passes drop from $20$ to $2$ under \textit{Last / Last}.
\subsection{Velocity Anchor Strength \& Ensemble Calibration Diagnostics}
\label{sec:app_ablation_anchor}

\begin{figure}[!htbp]
\centering
\includegraphics[width=\linewidth]{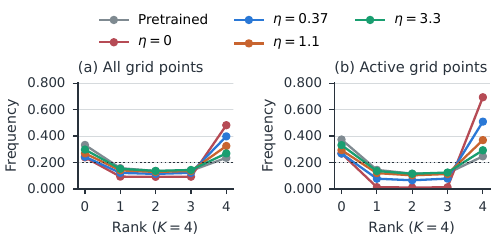}
\caption{\textbf{Observation rank histograms across anchor strengths $\eta$ on SEVIR-2h.} Rank frequencies of ground-truth observations among $K=4$ ensemble members across (a) all grid points and (b) active grid points ($\mathrm{VIL}>0$ in observation or any member). Dotted lines mark the uniform reference $1/(K+1)=0.2$ (ties broken uniformly).}
\label{fig:ablation_anchor_detail}
\end{figure}

\paragraph{Finite-ensemble calibration metrics.}
To evaluate ensemble dispersion and probabilistic calibration without finite-sample bias over $K=4$ members in Figure~\ref{fig:ablation_anchor} and Table~\ref{tab:ablation_anchor}, we compute the unbiased fair CRPS~\cite{schusterbauer2026freud}, ensemble spread, and finite-ensemble corrected spread--skill ratio ($\mathrm{SSR}$):
\begin{equation}
\begin{aligned}
\operatorname{CRPS}_{\mathrm{fair}}
&= \frac{1}{K}\sum_{m=1}^{K}\big|\hat{x}^{(m)} - x\big| \\
&\quad - \frac{1}{2K(K-1)}\sum_{m\neq m'}\big|\hat{x}^{(m)} - \hat{x}^{(m')}\big|,
\end{aligned}
\label{eq:app_fair_crps}
\end{equation}
\begin{equation}
\begin{aligned}
\operatorname{Spread}
&= \sqrt{\mathbb{E}\!\left[
    \frac{1}{K-1}\sum_{m=1}^{K}
    \big(\hat{x}^{(m)} - \bar{\hat{x}}\big)^2
  \right]}, \\
\operatorname{SSR}
&= \sqrt{\frac{K+1}{K}}\,
   \frac{\operatorname{Spread}}
        {\sqrt{\mathbb{E}\big[(\bar{\hat{x}} - x)^2\big]}},
\end{aligned}
\label{eq:app_ssr}
\end{equation}
where $\bar{\hat{x}} = \frac{1}{K}\sum_{m=1}^K \hat{x}^{(m)}$ is the ensemble mean, expectations pool squared quantities over all $7{,}220$ SEVIR-2h test sequences, $20$ lead times, and $128\times128$ grid points prior to taking square roots, and $\sqrt{(K+1)/K}=\sqrt{5/4}$ accounts for finite-member mean estimation ($\mathrm{SSR}=1$ denotes ideal second-moment dispersion calibration).

\begin{table*}[t]
\centering
\caption{\textbf{Velocity anchor strength ablation on SEVIR-2h} (at the $1{,}365$-update budget endpoint, with CFG). Deterministic verification and perceptual scores evaluate member $0$; $\mathrm{CRPS}_{\mathrm{fair}}$ ($\mathrm{VIL}$), $\mathrm{Spread}$ ($\mathrm{VIL}$), and dimensionless $\mathrm{SSR}$ evaluate $K=4$ ensemble members. Shading marks the default anchor weight $\eta=1.1$.}
\label{tab:anchor_strength}
\label{tab:ablation_anchor}
\footnotesize
\setlength{\tabcolsep}{3.8pt}
\begin{tabular}{@{}l*{10}{c}@{}}
\toprule
Anchor weight & CSI-M$^{\uparrow}$ & CSI$_{219}^{\uparrow}$ & POD-M$^{\uparrow}$ & FAR-M$^{\downarrow}$ & HSS-M$^{\uparrow}$ & LPIPS$^{\downarrow}$ & SSIM$^{\uparrow}$ & Fair CRPS$^{\downarrow}$ & Spread & SSR \\
\midrule
Pretrained Base w/ CFG & 0.3046 & 0.0730 & 0.3904 & 0.5634 & 0.3912 & 0.1837 & 0.6951 & 5.537 & 9.140 & 0.596 \\
$\eta=0$ (unanchored) & 0.3330 & 0.0751 & 0.4038 & 0.4829 & 0.4260 & 0.2993 & 0.6235 & 7.524 & 1.122 & 0.073 \\
$\eta=0.37$ & 0.3244 & 0.0747 & 0.3951 & 0.4875 & 0.4169 & 0.2024 & 0.6741 & 6.163 & 6.153 & 0.406 \\
\rowcolor{oursrow}$\eta=1.1$ (default) & 0.3170 & 0.0730 & 0.3891 & 0.4981 & 0.4077 & 0.1887 & 0.6988 & 5.727 & 7.809 & 0.513 \\
$\eta=3.3$ & 0.3104 & 0.0703 & 0.3846 & 0.5152 & 0.3991 & 0.1853 & 0.6987 & 5.581 & 8.605 & 0.564 \\
\bottomrule
\end{tabular}
\end{table*}

\begin{table*}[t]
\centering
\caption{\textbf{Zero-shot generalization across $10$-step inference grids at fixed weights on SEVIR-2h} (with CFG). $t_9$ denotes the flow time entering the final Euler step. Shading marks the default training/inference grid, and best metric values within each model block are \textbf{bold}.}
\label{tab:ablation_grid_scores}
\label{tab:ablation_grid}
\footnotesize
\setlength{\tabcolsep}{6pt}
\begin{tabular}{@{}l*{6}{c}@{}}
\toprule
Model \& Inference Grid & $t_9$ & CSI-M$^{\uparrow}$ & CSI$_{219}^{\uparrow}$ & POD-M$^{\uparrow}$ & FAR-M$^{\downarrow}$ & LPIPS$^{\downarrow}$ \\
\midrule
Base, Uniform ($s=1.0$) & 0.900 & 0.3031 & 0.0720 & \fst{0.3949} & 0.5736 & \fst{0.1669} \\
Base, Constant $s=1.5$ & 0.857 & 0.3037 & \fst{0.0730} & 0.3932 & 0.5702 & 0.1738 \\
\rowcolor{oursrow}Base, Constant $s=2.5$ (default) & 0.783 & 0.3046 & \fst{0.0730} & 0.3904 & 0.5634 & 0.1837 \\
Base, Constant $s=4.0$ & 0.692 & \fst{0.3050} & 0.0721 & 0.3873 & \fst{0.5552} & 0.1946 \\
Base, Decaying $s=2.5$ & 0.891 & 0.3032 & 0.0722 & 0.3928 & 0.5708 & 0.1696 \\
Base, Decaying $s=7.5$ & 0.880 & 0.3041 & 0.0723 & 0.3912 & 0.5656 & 0.1724 \\
\midrule
SAPT, Uniform ($s=1.0$) & 0.900 & \fst{0.3200} & 0.0762 & \fst{0.4001} & 0.5089 & \fst{0.1727} \\
SAPT, Constant $s=1.5$ & 0.857 & 0.3191 & 0.0781 & 0.3975 & 0.5090 & 0.1779 \\
\rowcolor{oursrow}SAPT, Constant $s=2.5$ (default) & 0.783 & 0.3170 & 0.0730 & 0.3891 & \fst{0.4981} & 0.1887 \\
SAPT, Constant $s=4.0$ & 0.692 & 0.3160 & \fst{0.0789} & 0.3903 & 0.5075 & 0.1947 \\
SAPT, Decaying $s=2.5$ & 0.891 & 0.3197 & 0.0764 & 0.3983 & 0.5071 & 0.1752 \\
SAPT, Decaying $s=7.5$ & 0.880 & 0.3192 & 0.0769 & 0.3965 & 0.5062 & 0.1773 \\
\bottomrule
\end{tabular}
\end{table*}

\begin{table*}[t]
\centering
\caption{\textbf{Model scaling results across backbone capacities on SEVIR-2h.} Architectural specifications for \textrm{S}, \textrm{B}, \textrm{L}, and \textrm{XL} are given in Table~\ref{tab:app_alignpre_sizes}; values correspond directly to Figure~\ref{fig:ablation_scaling} of the main text.}
\label{tab:ablation_scaling}
\footnotesize
\setlength{\tabcolsep}{6pt}
\begin{tabular}{@{}lc*{6}{c}@{}}
\toprule
Model Scale \& Stage & Params (M) & CSI-M$^{\uparrow}$ & CSI$_{219}^{\uparrow}$ & POD-M$^{\uparrow}$ & FAR-M$^{\downarrow}$ & HSS-M$^{\uparrow}$ & LPIPS$^{\downarrow}$ \\
\midrule
\textrm{S}: AlignPre-Base & 51.2 & 0.2880 & 0.0510 & 0.3546 & 0.5564 & 0.3679 & 0.1966 \\
\textrm{S}: AlignPre-Base w/ CFG & 51.2 & 0.3039 & 0.0690 & 0.3965 & 0.5767 & 0.3904 & 0.1897 \\
\rowcolor{oursrow}\textrm{S}: AlignPre (SAPT) & 51.2 & 0.3187 & 0.0758 & 0.3961 & 0.5015 & 0.4110 & 0.1945 \\
\midrule
\textrm{B}: AlignPre-Base & 115.0 & 0.2925 & 0.0549 & 0.3570 & 0.5407 & 0.3738 & 0.1888 \\
\textrm{B}: AlignPre-Base w/ CFG & 115.0 & 0.3045 & 0.0728 & 0.3904 & 0.5635 & 0.3911 & 0.1836 \\
\rowcolor{oursrow}\textrm{B}: AlignPre (SAPT) & 115.0 & 0.3169 & 0.0728 & 0.3890 & 0.4984 & 0.4076 & 0.1886 \\
\midrule
\textrm{L}: AlignPre-Base & 363.7 & 0.2964 & 0.0596 & 0.3570 & 0.5196 & 0.3787 & 0.1827 \\
\textrm{L}: AlignPre-Base w/ CFG & 363.7 & 0.3033 & 0.0736 & 0.3799 & 0.5473 & 0.3888 & 0.1780 \\
\rowcolor{oursrow}\textrm{L}: AlignPre (SAPT) & 363.7 & 0.3093 & 0.0710 & 0.3723 & 0.4859 & 0.3962 & 0.1852 \\
\midrule
\textrm{XL}: AlignPre-Base & 692.6 & 0.2972 & 0.0585 & 0.3542 & 0.5051 & 0.3784 & 0.1790 \\
\textrm{XL}: AlignPre-Base w/ CFG & 692.6 & 0.3007 & 0.0685 & 0.3713 & 0.5378 & 0.3840 & 0.1734 \\
\rowcolor{oursrow}\textrm{XL}: AlignPre (SAPT) & 692.6 & 0.3018 & 0.0648 & 0.3596 & 0.4886 & 0.3842 & 0.1829 \\
\bottomrule
\end{tabular}
\end{table*}

\paragraph{Preventing mode collapse and spectral loss.}
Table~\ref{tab:ablation_anchor} and Figure~\ref{fig:ablation_anchor_detail} empirically validate the model collapse. Optimizing the soft skill loss without velocity anchoring ($\eta=0$) induces severe reward hacking via ensemble collapse: although single-member CSI-M rises to $0.3330$ and FAR-M drops to $0.4829$, ensemble spread plummets by $87.7\%$ ($9.140\to1.122\,\mathrm{VIL}$, $\mathrm{SSR}=0.073$), fair CRPS worsens from $5.537$ to $7.524\,\mathrm{VIL}$, and forecasts blur into oversmoothed deterministic-like fields ($\mathrm{LPIPS}=0.2993$, $\mathrm{SSIM}=0.6235$), as visualized at $+100\,\mathrm{min}$ in Figure~\ref{fig:anchor_case} (illustrated on test sequence $6172$, selected near the active-storm median $\mathrm{VIL}\ge133$ coverage across $256$ evenly spaced test sequences). Consistently, Figure~\ref{fig:ablation_anchor_detail} shows that at $\eta=0$, active-point observation ranks concentrate heavily at the extreme boundary ranks ($0$ and $4$) with almost zero interior mass, accompanying the collapse of sub-$96\,\mathrm{km}$ spectral power in Figure~\ref{fig:ablation_anchor}(b).
Increasing $\eta$ monotonically restores ensemble spread (at nearly invariant ensemble-mean RMSE of $16.94\text{--}17.20\,\mathrm{VIL}$), recovers interior rank frequencies toward the pretrained distribution, and preserves short-wavelength convective power. Our default $\eta=1.1$ retains $85.4\%$ of pretrained spread ($\mathrm{SSR}=0.513$), improves SSIM above the pretrained model ($0.6988$ vs.\ $0.6951$), keeps LPIPS within $0.0050$ ($0.1887$), and secures substantial gains in CSI-M ($+0.0124$) and FAR-M ($-0.0653$).
\subsection{Inference Sensitivity \& Model Scaling Analyses}
\label{sec:app_ablation_grid}

\subsubsection{Zero-shot generalization across inference timestep grids}
\label{sec:app_ablation_grid_sub}

\begin{figure}[!htbp]
\centering
\includegraphics[width=0.45\textwidth]{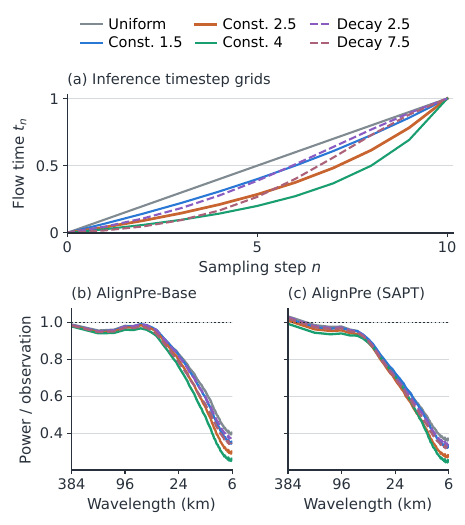}
\caption{\textbf{Inference timestep grids and spectral profiles on SEVIR-2h.} (a) Six $10$-step Euler schedules mapping step $n$ to flow time $t_n$. (b,c) Radially averaged Fourier power relative to ground truth at $N=10$ for AlignPre-Base and AlignPre (SAPT) across all six grids.}
\label{fig:ablation_grid}
\label{fig:ablation_grid_detail}
\end{figure}

Holding both pretrained and SAPT checkpoints fixed (trained under the constant-shift $s=2.5$ schedule), we test zero-shot inference across six $10$-step discretization schedules defined over $u_n=n/N$: uniform ($t_n=u_n$), constant shift ($t_n=u_n/[u_n+s(1-u_n)]$ for $s\in\{1.5,2.5,4.0\}$), and decaying shift ($t_n=r_n u_n/[1-u_n+r_n u_n]$ with $r_n=s^{u_n-1}$ for $s\in\{2.5,7.5\}$) (Figure~\ref{fig:ablation_grid}(a)). Although SAPT differentiates only through the final training step $t_9=0.783$, Table~\ref{tab:ablation_grid} demonstrates that its alignment transfers seamlessly across all six inference grids ($t_9\in[0.692,0.900]$), improving CSI-M by $+0.0110\text{--}0.0169$ and reducing FAR-M by $0.0477\text{--}0.0653$ over AlignPre-Base under every matched schedule. Uniform and decaying grids yield sharper LPIPS ($0.1727\text{--}0.1773$) and higher CSI-M ($0.3192\text{--}0.3200$) alongside stronger high-frequency power retention (Figure~\ref{fig:ablation_grid}(b,c)), whereas the matched constant-$2.5$ grid achieves the lowest FAR-M ($0.4981$). Because final-step backpropagation updates the shared clean-field predictor $f_\theta$, intensity calibration generalizes across trajectories rather than overfitting to a single integration node.

\subsubsection{Guidance scale and sampling-step sweeps}
\label{sec:app_ablation_inference}

\begin{figure}[!htbp]
\centering
\includegraphics[width=.89\linewidth]{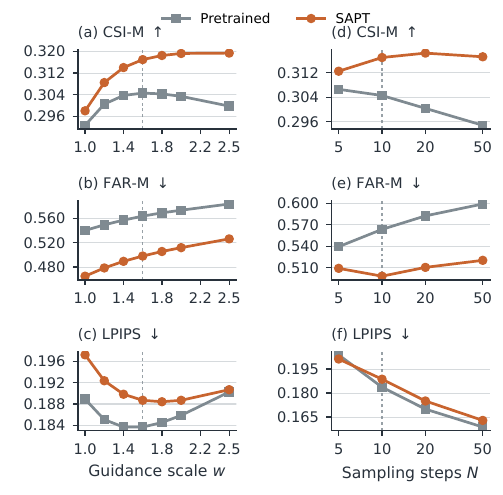}
\caption{\textbf{Inference sensitivity to classifier-free guidance scale $w$ and Euler step count $N$ on SEVIR-2h} (fixed checkpoints). Left: varying $w\in[1.0,2.5]$ at $N=10$. Right: varying $N\in\{5,10,20,50\}$ at $w=1.6$. Vertical dotted lines mark the SAPT rollout settings ($w=1.6, N=10$).}
\label{fig:ablation_inference}
\end{figure}

Figure~\ref{fig:ablation_inference} evaluates fixed-checkpoint robustness when varying guidance scale $w\in[1.0,2.5]$ and Euler integration steps $N\in\{5,10,20,50\}$ away from the SAPT rollout configuration ($w=1.6, N=10$). Across all tested settings, SAPT consistently achieves higher CSI-M and substantially lower FAR-M than AlignPre-Base. On the pretrained model, increasing $w$ peaks at $w=1.6$ ($\mathrm{CSI\text{-}M}=0.305$) before over-forecasting degrades CSI-M to $0.300$ at $w=2.5$ ($\mathrm{FAR\text{-}M}=0.583$); after SAPT, CSI-M scales monotonically up to $0.319$ at $w=2.5$ while keeping FAR-M ($0.526$) well below even the unguided base ($0.540$). Similarly, increasing integration steps from $N=5$ to $50$ monotonically improves LPIPS for both models, yet causes pretrained CSI-M to drop ($0.307\to0.295$) and FAR-M to rise ($0.539\to0.599$), as finer multi-step synthesis generates small-scale convective cells whose slight spatial displacements incur grid-point double penalties. SAPT eliminates this multi-step skill degradation, sustaining CSI-M at $0.317\text{--}0.318$ across $N\in\{10,20,50\}$ while improving LPIPS down to $0.163$ at $N=50$. Thus, neither tuning CFG nor increasing ODE solver steps can substitute for skill-aligned post-training.

\subsubsection{Model scaling analysis}
\label{sec:app_ablation_scaling}

Table~\ref{tab:ablation_scaling} provides the complete test metrics across our four hierarchical pixel-space transformer scales (\textrm{S}: $51.2\mathrm{M}$, \textrm{B}: $115.0\mathrm{M}$, \textrm{L}: $363.7\mathrm{M}$, and \textrm{XL}: $692.6\mathrm{M}$; Table~\ref{tab:app_alignpre_sizes}) plotted in Figure~\ref{fig:ablation_scaling} of the main text. Without guidance, scaling model capacity from \textrm{S} to \textrm{XL} monotonically improves CSI-M ($0.2880\to0.2972$), FAR-M ($0.5564\to0.5051$), and LPIPS ($0.1966\to0.1790$). Applying fixed classifier-free guidance yields diminishing CSI-M gains as capacity grows ($+0.0159$ on \textrm{S} vs.\ $+0.0035$ on \textrm{XL}, peaking on \textrm{B}) while inflating FAR-M by $+0.020\text{--}0.033$. Under a shared unretuned post-training recipe ($\rho=0.99$, $\eta=1.1$ calibrated for \textrm{B}), SAPT improves CSI-M and HSS-M while driving FAR-M down by $0.0492\text{--}0.0752$ over the guided base across every scale. Because larger pretrained models (\textrm{L} and \textrm{XL}) already exhibit more conservative storm coverage and lower reference false alarms $F_g^{\mathrm{ref}}$ (guided POD-M decreases from $0.3965$ on \textrm{S} to $0.3713$ on \textrm{XL}, with higher $\kappa_{2,g}=H_g^{\mathrm{ref}}/F_g^{\mathrm{ref}}$), directly reusing the fixed peak scale $\rho=0.99$ imposes stricter false-alarm suppression at high capacities—pushing FAR-M down to $0.4859\text{--}0.4886$ while contracting POD-M and narrowing net CSI-M gains, analogous to the larger-$\rho$ regime in Table~\ref{tab:kappa_control} (which can be relaxed via a smaller $\rho$ when prioritizing recall on large models). Notably, SAPT enables the compact $51.2\mathrm{M}$ \textrm{S} model ($\mathrm{CSI\text{-}M}=0.3187$, $\mathrm{FAR\text{-}M}=0.5015$) and $115.0\mathrm{M}$ \textrm{B} model ($\mathrm{CSI\text{-}M}=0.3169$, $\mathrm{FAR\text{-}M}=0.4984$) to substantially outperform the $13.5\times$ larger guided \textrm{XL} base ($\mathrm{CSI\text{-}M}=0.3007$, $\mathrm{FAR\text{-}M}=0.5378$), demonstrating high parameter efficiency for operational deployment.

\clearpage
\onecolumn
\setlength{\intextsep}{8pt plus 2pt minus 2pt}
\section{Supplementary qualitative case studies}
\label{sec:app_exp_qualitative}
\label{sec:app_exp_cases}

\begin{figure}[!htbp]
\centering
\includegraphics[width=\textwidth,height=.67\textheight,keepaspectratio]{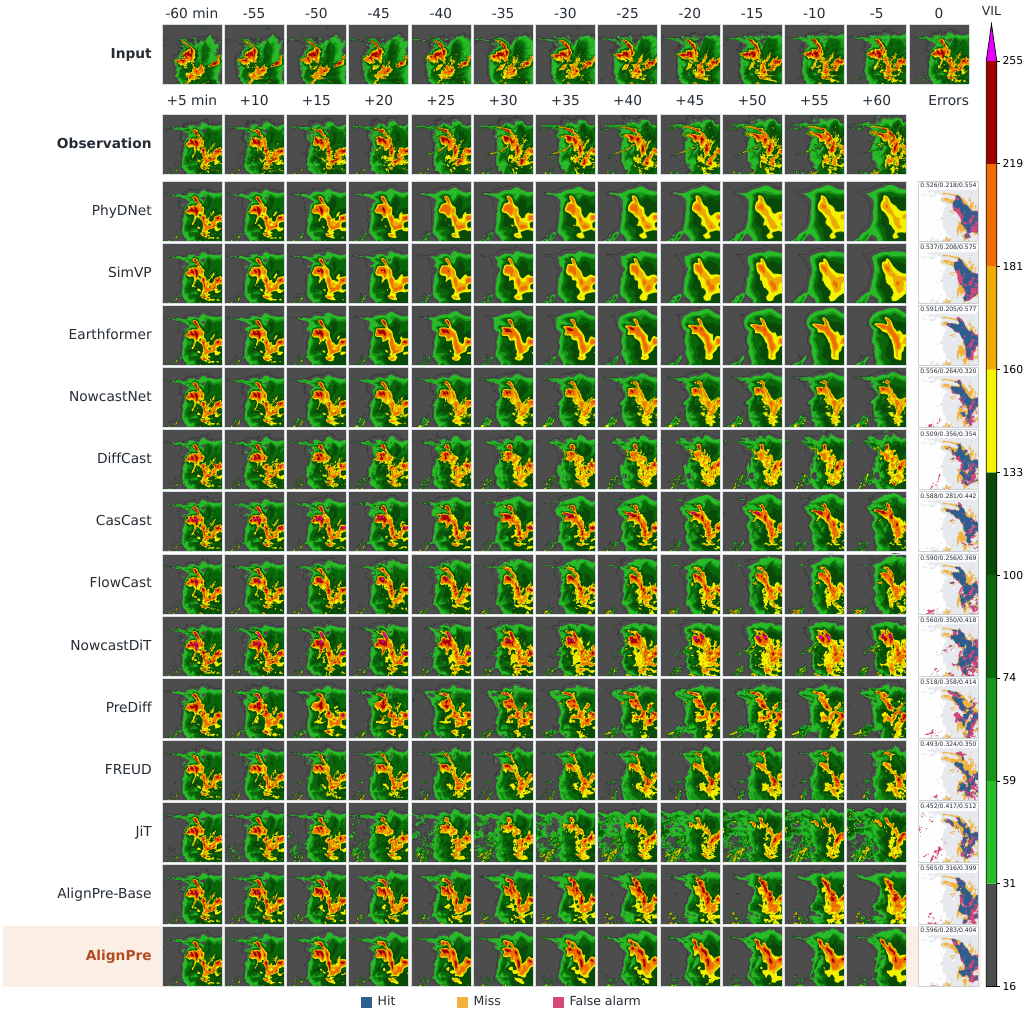}
\caption{\textbf{SEVIR-1h, case 1 (test sequence 10589).} The top row shows all 13 radar input frames, ending at 0\,min. Forecasts are shown every 5\,min from +5 to +60\,min. The Errors column marks hits, misses, and false alarms at +60\,min for VIL $\geq 133$. The numbers report case-level CSI-M/FAR-M/LPIPS over all forecast leads.}
\label{fig:supp_case_sevir_1h_1_1}
\end{figure}

\begin{figure}[!htbp]
\centering
\includegraphics[width=\textwidth]{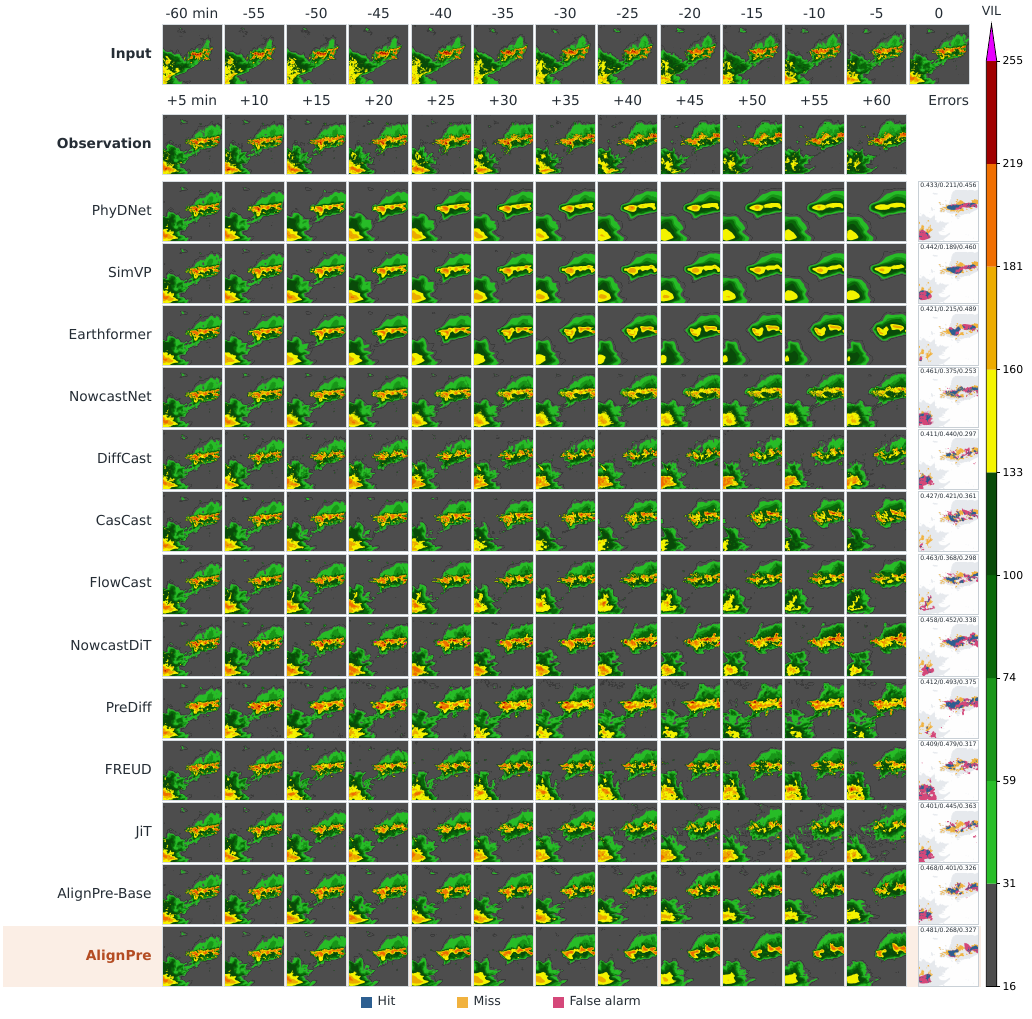}
\caption{\textbf{SEVIR-1h, case 2 (test sequence 784).} The top row shows all 13 radar input frames, ending at 0\,min. Forecasts are shown every 5\,min from +5 to +60\,min. The Errors column marks hits, misses, and false alarms at +60\,min for VIL $\geq 133$. The numbers report case-level CSI-M/FAR-M/LPIPS over all forecast leads.}
\label{fig:supp_case_sevir_1h_3_1}
\end{figure}

\begin{figure}[!htbp]
\centering
\includegraphics[width=\textwidth,height=.88\textheight,keepaspectratio]{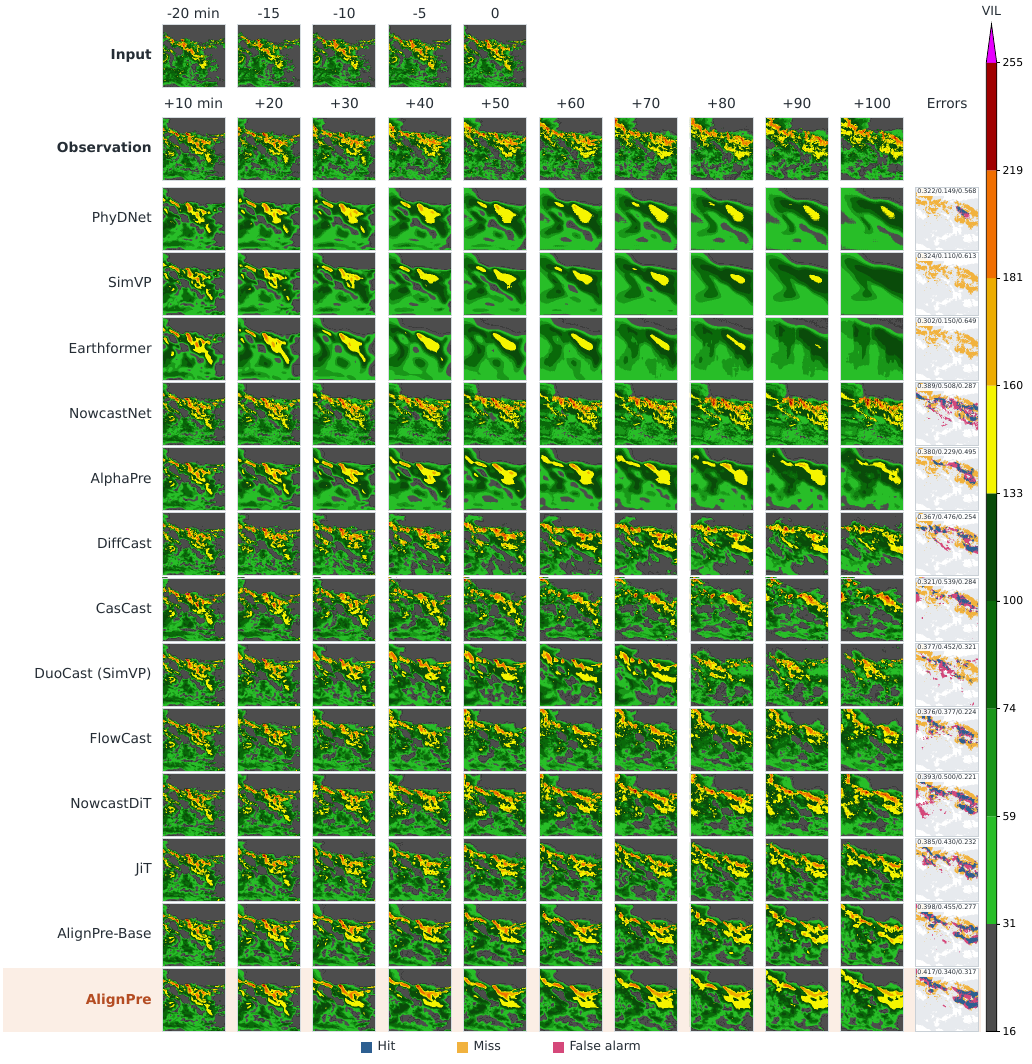}
\caption{\textbf{SEVIR-2h, case 1 (test sequence 154).} The top row shows all 5 radar input frames, ending at 0\,min. Forecasts are shown every 10\,min from +10 to +100\,min. The Errors column marks hits, misses, and false alarms at +100\,min for VIL $\geq 133$. The numbers report case-level CSI-M/FAR-M/LPIPS over all forecast leads.}
\label{fig:supp_case_sevir_2h_1_1}
\end{figure}

\begin{figure}[!htbp]
\centering
\includegraphics[width=\textwidth]{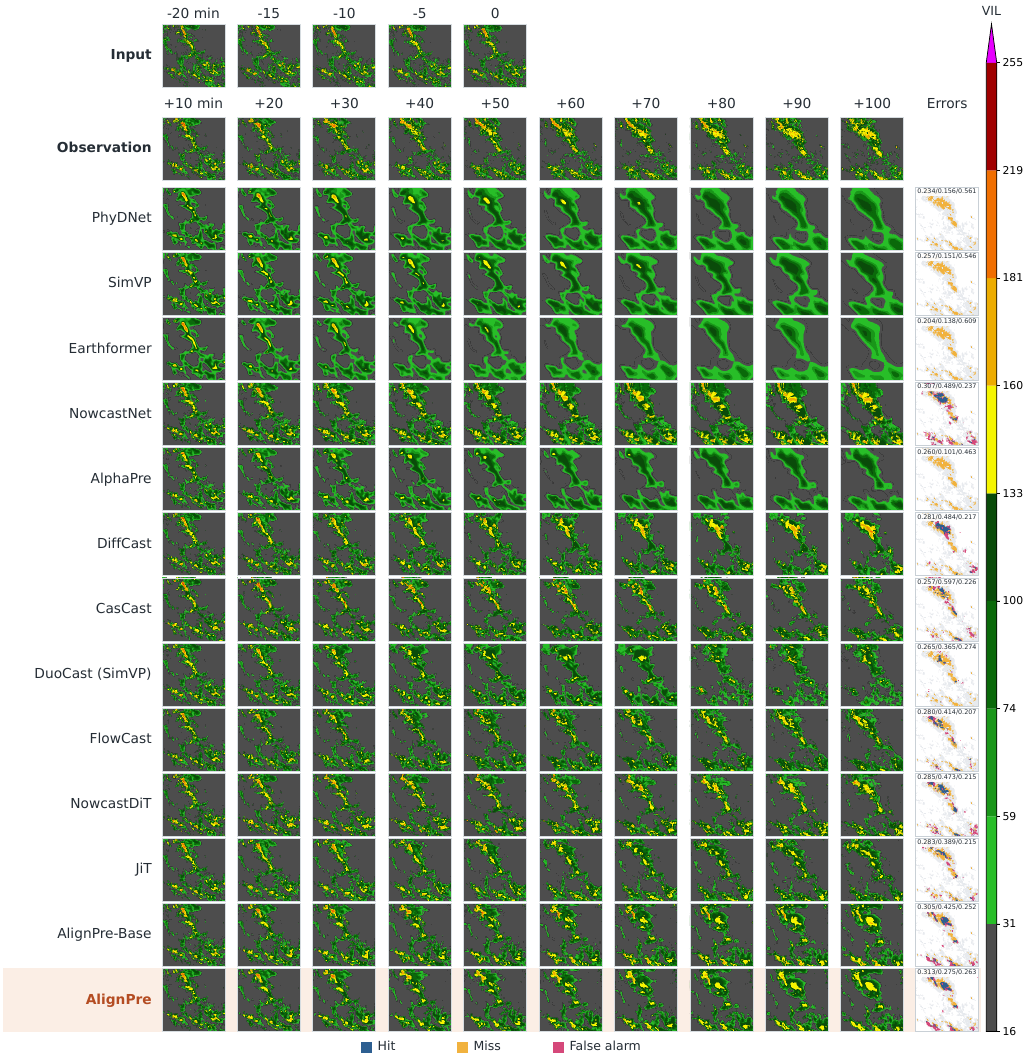}
\caption{\textbf{SEVIR-2h, case 2 (test sequence 3072).} The top row shows all 5 radar input frames, ending at 0\,min. Forecasts are shown every 10\,min from +10 to +100\,min. The Errors column marks hits, misses, and false alarms at +100\,min for VIL $\geq 133$. The numbers report case-level CSI-M/FAR-M/LPIPS over all forecast leads.}
\label{fig:supp_case_sevir_2h_2_1}
\end{figure}

\begin{figure}[!htbp]
\centering
\includegraphics[width=\textwidth]{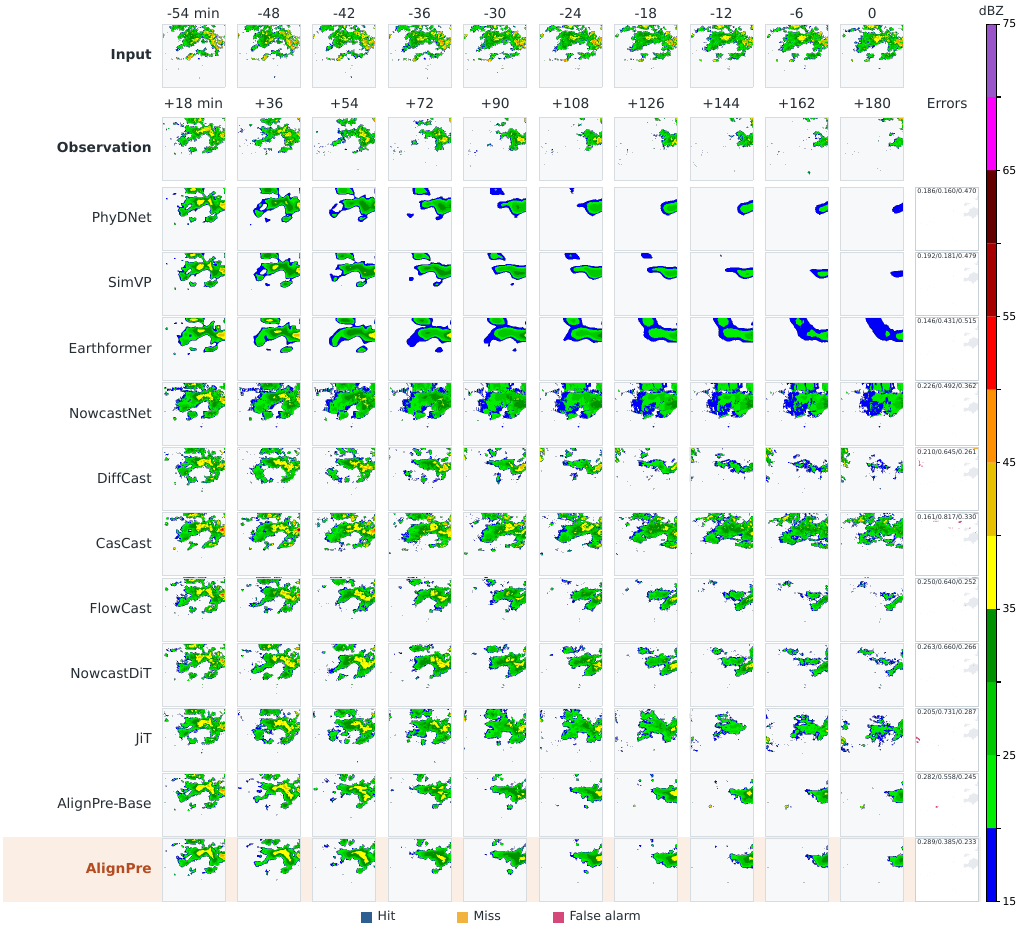}
\caption{\textbf{Guangdong-3h, case 1.} The top row shows all 10 radar input frames, ending at 0\,min. Forecasts are shown every 18\,min from +18 to +180\,min. The Errors column marks hits, misses, and false alarms at +180\,min for $\geq 35$\,dBZ. The numbers report case-level CSI-M/FAR-M/LPIPS over all forecast leads.}
\label{fig:supp_case_guangdong_3h_2_1}
\end{figure}

\begin{figure}[!htbp]
\centering
\includegraphics[width=\textwidth]{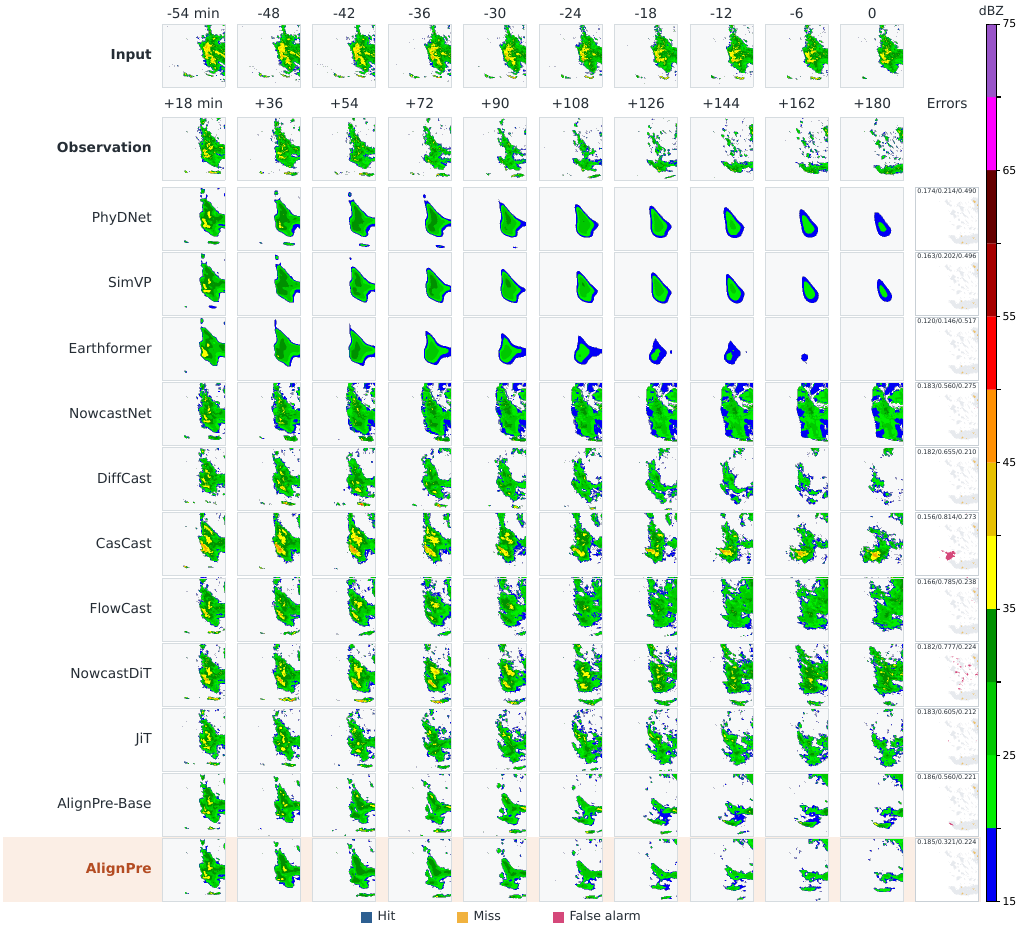}
\caption{\textbf{Guangdong-3h, case 2.} The top row shows all 10 radar input frames, ending at 0\,min. Forecasts are shown every 18\,min from +18 to +180\,min. The Errors column marks hits, misses, and false alarms at +180\,min for $\geq 35$\,dBZ. The numbers report case-level CSI-M/FAR-M/LPIPS over all forecast leads.}
\label{fig:supp_case_guangdong_3h_3_1}
\end{figure}

\paragraph{Qualitative comparisons.}
To provide a broader and uncurated view of spatiotemporal evolution across diverse storm regimes, Appendices~\ref{sec:app_cases_sevir384}--\ref{sec:app_cases_guangdong} present additional full-sequence case studies per benchmark (Figures~\ref{fig:supp_case_sevir_1h_1_1}--\ref{fig:supp_case_guangdong_3h_3_1}), comparing AlignPre against deterministic, hybrid, latent-space, and single-scale pixel-space baselines.
Across these extended sequences, regression-trained predictors such as SimVP progressively smooth out convective cores and frequently forecast zero echo above high thresholds at long lead times (e.g., $+100\,\mathrm{min}$ on SEVIR-2h and $+180\,\mathrm{min}$ on Guangdong-3h), whereas unaligned or RL-tuned generative baselines tend to produce scattered false-alarm cells around storm perimeters.
AlignPre consistently preserves organized convective rainbands and sharper hit--false-alarm boundaries throughout the forecast horizon, although its finest sub-grid textures remain moderately smoother than radar observations at extended lead times.

\subsection{SEVIR-1h case studies}
\label{sec:app_cases_sevir384}


\clearpage
\subsection{SEVIR-2h case studies}
\label{sec:app_cases_sevir128}


\FloatBarrier
\subsection{Guangdong-3h case studies}
\label{sec:app_cases_guangdong}



\end{document}